\documentclass[11pt]{article} 

\newif\ifplots
\plotstrue \usepackage[letterpaper,margin=1in]{geometry}

\newif\ifred
\redtrue 

\usepackage[T1]{fontenc}
\usepackage{microtype}

\usepackage{titlesec}
\titleformat*{\paragraph}{\bfseries}

\usepackage{amsthm,amsmath,amssymb,amsfonts,amssymb,mathtools}
\usepackage{amsmath,amssymb,amsfonts,amssymb,mathtools}
\usepackage{xcolor}
\usepackage{empheq}
\usepackage{dsfont}
\usepackage{mathrsfs}
\usepackage{graphicx}
\usepackage{enumitem}
\usepackage{aliascnt}

\ifred

\else

\fi

\usepackage{caption}
\usepackage{tikz}
\usetikzlibrary{patterns}
\usetikzlibrary{arrows,shapes,automata,backgrounds,petri,positioning}
\usetikzlibrary{shadows}
\usetikzlibrary{calc}
\usetikzlibrary{spy}
\usetikzlibrary{angles, quotes}
\ifplots
\usetikzlibrary{matrix}
\usepackage{pgf,pgfplots}
\usepackage{pgfmath,pgffor}
\pgfplotsset{compat=1.17}
\fi
\usepackage{framed}
\usepackage{algorithm2e}
\usepackage[noend]{algpseudocode}
\SetAlCapFnt{\normalfont\small}\SetAlCapNameFnt{\unskip\itshape\small}

\definecolor[named]{ACMBlue}{cmyk}{1,0.1,0,0.1}
\definecolor[named]{ACMYellow}{cmyk}{0,0.16,1,0}
\definecolor[named]{ACMOrange}{cmyk}{0,0.42,1,0.01}
\definecolor[named]{ACMRed}{cmyk}{0,0.90,0.86,0}
\definecolor[named]{ACMLightBlue}{cmyk}{0.49,0.01,0,0}
\definecolor[named]{ACMGreen}{cmyk}{0.20,0,1,0.19}
\definecolor[named]{ACMPurple}{cmyk}{0.55,1,0,0.15}
\definecolor[named]{ACMDarkBlue}{cmyk}{1,0.58,0,0.21}

\usepackage[colorlinks,citecolor=blue,linkcolor=magenta,bookmarks=true]{hyperref}
\usepackage[nameinlink]{cleveref}
\crefname{ineq}{Inequality}{Inequality}
\creflabelformat{ineq}{#2{\upshape(#1)}#3}
\crefname{sub}{Subsection}{Subsection}
\creflabelformat{Subsection}{#2{\upshape(#1)}#3}
\crefname{sdp}{SDP}{SDP}
\creflabelformat{sdp}{#2{\upshape(#1)}#3}
\crefname{lp}{LP}{LP}
\creflabelformat{lp}{#2{\upshape(#1)}#3}
\usepackage{aliascnt}
\usepackage{cleveref}
\crefname{ineq}{Inequality}{Inequality}
\creflabelformat{ineq}{#2{\upshape(#1)}#3}
\crefname{sub}{Subsection}{Subsection}
\creflabelformat{Subsection}{#2{\upshape(#1)}#3}
\crefname{sdp}{SDP}{SDP}
\creflabelformat{sdp}{#2{\upshape(#1)}#3}
\crefname{lp}{LP}{LP}
\creflabelformat{lp}{#2{\upshape(#1)}#3}

\makeatletter
\newenvironment{Ualgorithm}[1][htpb]{\def\@algocf@post@ruled{\kern\interspacealgoruled\hrule  height\algoheightrule\kern3pt\relax}\def\@algocf@capt@ruled{under}\setlength\algotitleheightrule{0pt}\SetAlgoCaptionLayout{centerline}\begin{algorithm}[#1]}
{\end{algorithm}}
\makeatother

\newcommand{\CS}{Cauchy-Schwarz}

\newtheorem{theorem}{Theorem}[section]

\newtheorem{lemma}[theorem]{Lemma}
\newtheorem{informal theorem}[theorem]{Theorem (informal statement)}

\newtheorem{proposition}[theorem]{Proposition}
\newtheorem{corollary}[theorem]{Corollary}
\newtheorem{claim}[theorem]{Claim}
\newtheorem{fact}[theorem]{Fact}

\newtheorem{remark}[theorem]{Remark}

\newtheorem{definition}[theorem]{Definition}
\newcommand{\eqdef}{\coloneqq}

\newcommand\snorm[2]{\left\| #2 \right\|_{#1}}
\renewcommand\vec[1]{\mathbf{#1}}
\DeclareMathOperator*{\pr}{\mathbf{Pr}}
\DeclareMathOperator*{\E}{\mathbf{E}}
\newcommand{\proj}{\mathrm{proj}}

\def\d{\mathrm{d}}

\DeclareMathOperator*{\argmin}{argmin}

\newcommand{\bx}{\mathbf{x}}
\newcommand{\by}{\mathbf{y}}

\newcommand{\bw}{\mathbf{w}}

\newcommand{\R}{\mathbb{R}}

\newcommand{\Z}{\mathbb{Z}}

\newcommand{\eps}{\epsilon}

\newcommand{\poly}{\mathrm{poly}}
\newcommand{\polylog}{\mathrm{polylog}}
\newcommand{\var}{\mathbf{Var}}

\newcommand{\sgn}{\mathrm{sign}}
\newcommand{\sign}{\mathrm{sign}}

\newcommand{\Ex}{\mathop{{\bf E}\/}}
\newcommand{\opt}{\mathrm{opt}}
\newcommand{\D}{D}

\newcommand{\Ind}{\mathds{1}}
\newcommand{\1}{\Ind}

\newcommand{\littlesum}{\mathop{\textstyle \sum}}

\newcommand{\wt}{\widetilde}
\newcommand{\wh}{\widehat}

\newcommand{\wstar}{\bw^{\ast}}
\newcommand{\x}{\vec x}

\newcommand{\ith}{^{(i)}}
\newcommand{\tth}{^{(t)}}
\newcommand{\Exx}{\E_{\x\sim \D_\x}}
\newcommand{\Ey}{\E_{(\x,y)\sim \D}}
\newcommand{\vpar}[2]{\vec #1^{\|_{\vec #2}}}
\newcommand{\vperp}[2]{\vec #1^{\perp{\vec #2}}}

\newcommand{\boundc}{\mu}
\newcommand{\boundd}{\xi}
\newcommand{\booundd}{\xi}

\newcommand{\citet}{\cite}
\newcommand{\citep}{\cite}

\title{Learning a Single Neuron with Adversarial Label Noise \\ via Gradient Descent}
\author{
Ilias Diakonikolas\thanks{Supported by NSF Medium Award CCF-2107079,
NSF Award CCF-1652862 (CAREER), a Sloan Research Fellowship, and
a DARPA Learning with Less Labels (LwLL) grant.}\\
UW Madison\\
{\tt ilias@cs.wisc.edu}\\
\and
Vasilis Kontonis\thanks{Supported in part by NSF Award CCF-2144298 (CAREER).}\\
UW Madison\\
{\tt kontonis@wisc.edu }\\
\and
Christos Tzamos\thanks{Supported by NSF Award CCF-2144298 (CAREER).}\\
UW Madison\\
{\tt tzamos@wisc.edu}
\and
Nikos Zarifis\thanks{Supported in part by NSF Award CCF-1652862 (CAREER) 
and a DARPA Learning with Less Labels (LwLL) grant.}\\
UW Madison\\
{\tt zarifis@wisc.edu}\\
}

\begin{document}

\maketitle

\begin{abstract}We study the fundamental problem of learning a single neuron, i.e., a 
function of the form $\x \mapsto \sigma(\vec w \cdot \x)$ for monotone activations 
$\sigma:\R \mapsto \R$, with respect to the $L_2^2$-loss in the presence of adversarial label noise. 
Specifically, we are given labeled examples from a distribution $D$ on $(\x, y) \in \R^d \times \R$
such that there exists $\vec w^\ast \in \R^d$ achieving $F(\vec w^\ast)  = \eps$, where
$F(\vec w) = \E_{(\x,y) \sim D}[(\sigma(\vec w\cdot \x) - y)^2]$. The goal of the learner
is to output a hypothesis vector $\wt{\vec w}$ such that $F(\wt{\vec w}) = C \, \eps$ with 
high probability, where $C>1$ is a universal constant. As our main contribution, we give
efficient constant-factor approximate learners 
for a broad class of distributions (including log-concave distributions)
and activation functions (including ReLUs and sigmoids).
Concretely, for the class of isotropic log-concave distributions, we obtain
the following important corollaries:
\begin{itemize}[leftmargin=2pc, rightmargin = 1.5pc]
\item For the logistic activation, i.e., $\sigma(t) = 1/(1+e^{-t})$, we obtain the first
polynomial-time constant factor approximation (even under the Gaussian distribution).
Our algorithm has sample complexity $\wt{O}(d/\eps)$, which is tight within 
polylogarithmic factors.  

\item For the ReLU activation, i.e., $\sigma(t) =  \max(0,t)$, we give an efficient algorithm with 
sample complexity $\wt{O}(d \, \polylog(1/\eps))$. Prior to our work, the best known 
constant-factor approximate learner had sample complexity $\tilde{\Omega}(d/\eps)$.
\end{itemize} 
In both of these settings, our algorithms are simple, performing gradient-descent on the (regularized) $L_2^2$-loss. 
The correctness of our algorithms relies on novel structural results that we establish, 
showing that (essentially all) stationary points of the underlying non-convex loss
are approximately optimal.
\end{abstract}

\setcounter{page}{0}
\thispagestyle{empty}
\newpage

\section{Introduction}
\subsection{Background and Motivation}
The recent success of deep learning has served as a practical motivation for the development
of provable efficient learning algorithms for various natural classes of neural networks.
Despite extensive investigation, our theoretical understanding of the assumptions 
under which neural networks are provably efficiently learnable remains somewhat limited.
Here we focus on arguably the simplest possible setting of learning a {\em single} neuron,
i.e., a real-valued function of the form $\mathbf{x} \mapsto \sigma(\mathbf{w} \cdot \mathbf{x})$, 
where $\mathbf{w}$ is the weight vector of parameters and $\sigma: \R \to \R$ is a fixed 
non-linear and monotone activation function. Concretely, the learning problem is the following:
Given i.i.d.\ samples from a distribution
$D$ on $(\mathbf{x}, y)$, where $\mathbf{x} \in \R^d$ is the feature vector and $y \in \R$ 
is the corresponding label, our goal is to learn the underlying function in $L_2^2$-loss.
That is, the learner's objective is to output a hypothesis
$h: \R^d \to \R$ such that $\E_{(\mathbf{x}, y) \sim D} [(h(\mathbf{x}) - y)^2]$ 
is as small as possible, compared to the minimum possible loss 
$\opt: = \min_{\mathbf{w} \in \R^d} \E_{(\mathbf{x}, y) \sim D} [(\sigma(\mathbf{w} \cdot \mathbf{x}) - y)^2]$. 
Settings of particular interest for the activation $\sigma$ 
include the ReLU and sigmoid functions, corresponding to
$\sigma(u) = \mathrm{ReLU}(u) \eqdef \max\{0, u\}$ and $\sigma(u) \eqdef 1/(1+\exp(-u))$ respectively.
Recall that a learning algorithm is called {\em proper} if the hypothesis $h$
is restricted to be of the form 
$h_{\widehat{\mathbf{w}}}(\mathbf{x}) = \sigma(\widehat{\mathbf{w}} \cdot \mathbf{x})$.
Throughout this paper, we focus on developing efficient proper learners.

In the realizable case, i.e., when the labels $y$ are consistent with a function in the class,
the above learning problem is known to be solvable in polynomial time 
for a range of activation functions. A line of work, see, e.g.,~\citet{kalai2009isotron, Sol17, YS20} 
and references therein, has shown that simple algorithms like gradient-descent 
efficiently converge to an optimal solution under additional assumptions 
on the marginal distribution $D_{\bx}$ on examples.

In this work, we focus on the {\em agnostic} learning model, where no realizability assumptions 
are made on the distribution $D$. Roughly speaking, the agnostic model corresponds 
to learning in the presence of adversarial label noise. 

\begin{definition}[Learning Single Neurons with Adversarial Noise] \label{def:agnostic-learning}
Fix $\eps, W > 0$, $\delta \in(0,1)$, and a class of distributions $\mathcal{G}$ on $\R^d$.
Let $\sigma : \R \mapsto \R$ be an activation function and $\D$ a distribution on labeled examples
$(\x,y) \in \R^d \times \R$ such that its $\x$-marginal belongs in $\mathcal{G}$. 
We define the population $L_2^2$-loss as 
$F^{D, \sigma}(\vec w) \triangleq (1/2)\E_{(\x, y) \sim \D} [(\sigma(\vec w \cdot \vec x) - y)^2]$.
We say that $D$ is $(\eps, W)$-corrupted if \( \inf_{\|\vec w\|_2 \leq W} F^{D,\sigma}(\vec w) \leq
\eps\,.  \) For some $C \geq 1$, a $C$-approximate learner is given $\eps, \delta, W$ and
i.i.d.\ labeled examples from $\D$ and outputs a function $h(\x): \R^d \mapsto \R$
such that, with probability at least $1-\delta$, it holds 
\( (1/2)\E_{(\x, y) \sim \D}[(h(\x) - y)^2] \leq C  \eps\,.  \)
\end{definition}

\noindent Some comments are in order. First, we note that the parameter $\eps$ 
quantifies the degree of contamination --- in the sense that the closest function in the class
has $L_2^2$-loss $\eps$. (Sometimes, this is denoted by $\opt$ in the relevant literature.)
The parameter $W$ quantifies the radius of the ball that contains an optimal vector $\vec w^{\ast}$.
For uniformly bounded activations (e.g., sigmoids), one can remove the 
restriction on the norm of the weight vector $\vec w$, i.e., take $W = +\infty$. 
In this case, we call $D$ $\eps$-corrupted. 
When the distribution and activation are clear from the context,  
we will write $F(\vec w)$ instead of $F^{D,\sigma}(\vec w)$.

\paragraph{Related Prior Work}
In this paper, we focus on developing efficient {\em constant-factor} approximate proper learners
for a range of activation functions. For this to be possible in polynomial time, one needs to make
some qualitative assumptions on the underlying marginal distribution on examples. Indeed, it is
known~\citep{DKMR22} that in the distribution-free setting no constant-factor approximation is
possible (even for improper learning), under cryptographic assumptions, for a range of activations
including ReLUs and sigmoids.  On the other hand, even under Gaussian marginals, achieving error
$\opt+\eps$ (corresponding to $C=1$ in \Cref{def:agnostic-learning}) requires time
$d^{F(1/\eps)}$, for some function $F$ with $\lim_{u \to \infty}F(u) = \infty$~\citep{GoelKK19,
DKZ20,GGK20, DKPZ21}.  
These hardness results motivate the design of constant-factor approximate learners under ``well-behaved'' distributions.

On the algorithmic side,~\citet{GoelKKT17} gave an algorithm with error $\opt+\eps$ and runtime
$\poly(d) 2^{\poly(1/\eps)}$ that succeeds as long as the distribution on examples is supported on
the unit sphere. \citet{FCG20} study ReLU activations under structured distributions and show that
gradient-descent on the $L_2$ loss converges to a weight vector with error $O(d ~ \eps )$. The most
relevant prior work is by~\citet{DGKKS20} who gave a $\poly(d/\eps)$-time constant-factor
approximate proper learner for ReLU activations under isotropic log-concave distributions. Their
algorithm makes essential use of the ReLU activation.  In fact,~\citet{DGKKS20} asked whether
efficient constant-factor approximations exist for other activations, including sigmoids. In this
work, we answer this open question in the affirmative.

The aforementioned discussion motivates the following broad question:
\begin{center}
\emph{Is there an efficient constant-factor approximate learner\\ 
for single neurons under well-behaved distributions?}
\end{center}
In this work, we answer this question in the affirmative for a range of activations including ReLUs
and sigmoids and a variety of well-behaved distributions.  In fact, we show that a simple
gradient-based method on the $L_2^2$-loss suffices.

\subsection{Our Results}

\paragraph{Distributional Assumptions}
We develop algorithms that are able to learn single neurons under a large class of structured
distributions.  We make mild distributional assumptions requiring only concentration,
anti-concentration, and anti-anti-concentration on the $\x$-marginal of the examples.  In
particular, we consider the following class of well-behaved distributions.
\begin{definition}[Well-behaved Distributions] \label{def:bounds}
Let $L, R >0$.
An isotropic (i.e., zero mean and identity covariance) distribution $\D_{\bx}$ on $\R^d$ is called
$(L,R)$-well-behaved if for any projection $(\D_{\bx})_V$ of $\D_{\bx}$ onto a subspace $V$ of
dimension at most two, the corresponding pdf $\gamma_V$ on $\R^2$ satisfies the following:
\begin{itemize}
\item   For all $\vec x \in V$ such that $\snorm{\infty}{\vec x} \leq R$ it holds $\gamma_V(\vec x)
\geq L$  (anti-anti-concentration). 
\item  For all $\x \in V$ it holds that  $\gamma_V(\x) \leq (1/L)(e^{-L \| \x \|_2 })$
(anti-concentration and concentration).
\end{itemize}
When the parameters $L,R$ are bounded above by universal constants (independent of the dimension),
we will simply say that the distribution $\D_\x$ is well-behaved.
\end{definition}

The class of well-behaved distributions is fairly broad.  Specifically, isotropic log-concave
distributions are well-behaved, i.e., they are $(L, R)$-well-behaved for some $L, R = O(1)$, see,
e.g., \cite{LV07,KlivansLT09}.  Similar assumptions were introduced in~\cite{DKTZ20} 
and have been used in various classification and regression settings~\citep{DKTZ20c, DKTZ20b, DKKTZ20, DKKTZ21b, FCG20,Frei21,ZL21,ZFG21}.  

\paragraph{Learning Sigmoidal Activations}
Our first main result holds for a natural class of activations that roughly have
``sigmoidal'' shape. 
\begin{definition}[Sigmoidal Activations]
\label{def:well-behaved-bounded-intro}
Let $\sigma:\R\mapsto\R$ be a non-decreasing activation function and $\tau,\boundc,\boundd>0$. 
We say that $\sigma$ is $(\tau, \boundc, \boundd)$-sigmoidal if it satisfies 
(a) $\sigma'(t) \geq \tau$, for all $t \in [-1,1]$, and (b) $\sigma'(t) \leq \boundd e^{- \boundc |t| }$, for all $t \in \R$.  
We will simply say that an activation is ``sigmoidal'' when $\tau, \boundc, \boundd$ are universal constants.
\end{definition}

Arguably the most popular sigmoidal activation is the logistic activation or sigmoid, corresponding
to $\sigma(t) = 1/(1+e^{-t})$.  Other well-studied sigmoidal activations include the hyperbolic
tangent, the Gaussian error function, and the ramp activation (see \Cref{fig:tanh}).  We note that all these activations
satisfy the requirement of \Cref{def:well-behaved-bounded-intro} for some universal constants
$\tau,\boundc,\booundd$.  In what follows, we will simply refer to them as sigmoidal.

The most commonly used method to solve our learning problem in practice 
is to directly attempt to minimize the $L_2^2$-loss via (stochastic) gradient descent.
Due to the non-convexity of the objective, this method is of a heuristic nature in general
and comes with no theoretical guarantees, even in noiseless settings.
In our setting, the situation is even more challenging due to the adversarial noise
in the labels. Indeed, we show that the ``vanilla'' $L_2^2$-objective may contain bad local optima, 
even under Gaussian marginals. Specifically, 
even with an arbitrarily small amount of adversarial noise, 
the vanilla $L_2^2$ objective will have local-minima whose $L_2^2$-error is larger than $1/2$ 
(which is essentially trivial, since sigmoidal activations take values in $[-1,1]$); 
see \Cref{theorem:intro_bad_stationary} and \Cref{main-fig:non-convex-landscape}.  

Our main structural result for sigmoidal activations is that we can ``correct'' the optimization
landscape of the $L_2^2$-loss by introducing a standard $\ell_2$-regularization term.  We prove the
following theorem showing that any stationary point of the regularized $L_2^2$-loss is approximately
optimal. 
\begin{theorem}[Informal: Landscape of Sigmoidals]
\label{main-thm:stationary-points-ramp-restate}
For sigmoidal activations and $\eps$-corrupted well-behaved distributions,  any (approximate)
stationary point $\bar{\vec w}$ of the $\ell_2$ regularized objective $F_\rho(\vec w) = F(\vec w) +
(\rho/2) \|\vec w\|_2^2$ with $\rho= \Theta(\eps^3)$, satisfies $F(\bar{\vec w})= O(\eps)$.
\end{theorem}

Standard gradient methods (such as SGD) are known to efficiently converge to stationary points of
non-convex objectives under certain assumptions. By running any such method on our regularized
objective, we readily obtain an efficient algorithm that outputs a weight vector with $L_2^2$-loss
$O(\eps)$. This is already a new result in this context.
Yet, black-box application of optimization results for finding stationary
points of non-convex functions would resulting 
in a sample complexity with suboptimal dependence on $\eps$, up to polynomial factors.

Aiming towards an algorithmic result with near-optimal sample complexity, we perform a
``white-box'' analysis of gradient descent, leveraging the optimization landscape of sigmoidals.
Specifically, we show that ``vanilla'' gradient descent, with a fixed step size, finds 
an approximately optimal solution when run on the empirical (regularized) $L_2^2$-loss with 
a near-optimal number of samples.

\begin{theorem}[Informal: Learning Sigmoidals via Gradient Descent] \label{thm:informal-bounded}
For sigmoidal activations and $\eps$-corrupted well-behaved distributions, gradient descent on the
empirical regularized loss $\wh{F}_{\rho}(\cdot)$ with $N = \wt{\Theta}(d/\eps)$ samples, converges, in
$\poly(1/\eps)$ iterations, to a vector $\wt{\vec w}$ satisfying \( F(\wt{\vec  w} )
\leq O(\eps) \) with high probability.
\end{theorem}

\Cref{thm:informal-bounded} gives the first efficient constant-factor approximate learner for
sigmoid activations in the presence of adversarial label noise, answering an open problem
of~\citet{DGKKS20}. As an additional bonus, our algorithm is simple and potentially practical
(relying on gradient-descent), has near-optimal sample complexity (see~\Cref{lem:sample-lower-bound-sigmoidal}), 
and succeeds for a broad family of bounded activation
functions (i.e., the ones satisfying \Cref{def:well-behaved-bounded-intro}).

For simplicity of the exposition, we have restricted our attention to sigmoidal activations 
(corresponding to $ \tau, \mu,\xi$ being universal constants in \Cref{def:well-behaved-bounded-intro}) 
and well-behaved distributions (corresponding to $L, R = O(1)$ in \Cref{def:bounds}).
In the general case, we note that the complexity of our algorithm is
polynomial in the parameters $L, R, \tau, \mu,\xi$, see \Cref{thm:bounded-main-theorem}.

\paragraph{Learning Unbounded Activations}
We next turn our attention to activation functions that are not uniformly bounded.  
The most popular such activation is the ReLU function $\sigma(t) = \mathrm{Relu}(t) = \max(0,t)$. 
Our algorithmic results apply to the following class of unbounded activations.

\begin{definition}[Unbounded Activations]
\label{def:well-behaved-unbounded-intro}
Let $\sigma:\R\mapsto\R$ be a non-decreasing activation function and $\alpha,\lambda>0$. 
We say that $\sigma$ is $(\alpha, \lambda)$-unbounded if it satisfies (a) $\sigma$ is $\lambda$-Lipschitz
and (b) $\sigma'(t) \geq \alpha$, for all $t \in [0, +\infty)$.  
We will simply say that an activation is unbounded
when the parameters $\alpha, \lambda$ are universal constants.
\end{definition}

We use the term ``unbounded'' for these activations, as they tend to $\infty$ as $t \to +\infty$.
Most well-known unbounded activation functions such as the ReLU, Leaky-ReLU, ELU are 
$(\alpha,\lambda)$-unbounded for some absolute constants $\alpha, \lambda > 0$ (see \Cref{fig:unbounded-activation-functions}).  
For example, the ReLU activation is $(1,1)$-unbounded.  

Our main structural result for unbounded activations is that all stationary points $\vec w$ 
of the $L_2^2$-loss that lie in the halfspace $\vec w^\ast \cdot \vec w \geq 0$, 
where $\vec w^\ast$ is the optimal weight vector, are approximately optimal.
In more detail, we establish the following result.

\begin{theorem}[Informal: Landscape of Unbounded Activations] 
\label{main-prop:stationary-points-unbounded}
For unbounded activations and $\eps$-corrupted well-behaved distributions, 
any stationary point $\bar{\vec w}$ of $F(\vec w)$ with $\vec w \cdot \wstar \geq 0$ 
satisfies $F(\bar{\vec w})=O(\eps)$, and any $\vec w$ with $\vec w \cdot \wstar \geq 0$ 
and $F(\vec w) = \Omega(\eps)$, satisfies $\nabla F(\vec w) \cdot (\vec w - \vec w^\ast) > 0$.
\end{theorem}

We remark that the constant in the error guarantee of the above theorem only depends on the
(universal) constants of the distribution and the activation and not on the radius $W$ of
\Cref{def:agnostic-learning}.  Interestingly, \Cref{main-prop:stationary-points-unbounded} 
does not preclude the existence of suboptimal stationary points.  On the other hand, similarly to
the case of sigmoidal activations, our structural result can be readily combined with ``black-box''
optimization to efficiently find a constant-factor approximately optimal weight vector $\vec w$ (see
\Cref{main-sec:bounded-plain-landscape}). 
To obtain near-optimal sample complexity, we again perform an ``white-box'' analysis of
(approximate) gradient descent, showing that simply by initializing at $\vec 0$ 
we can avoid bad stationary points.

\begin{theorem}[Informal: Learning Unbounded Activations via Gradient Descent]
\label{thm:informal-unbounded-main-theorem}
For unbounded activations and $(\eps,W)$-corrupted well-behaved distributions, (approximate) gradient
descent on the $L_2^2$-loss $F(\cdot)$ with sample size $N = \wt{\Theta}(d W^2$
$\max(\polylog(W/\eps), 1))$ and $\polylog(1/\eps)$ iterations,  converges to a vector $\wt{\vec w}
\in \R^d$, satisfying \( F(\wt{\vec  w} ) \leq O(\eps) \) with high probability.
\end{theorem}

\Cref{thm:informal-unbounded-main-theorem} is a broad
generalization of the main result of~\citet{DGKKS20}, which gave 
a constant-factor approximation (using a different approach)
for the special case of ReLU activations under isotropic log-concave distributions. 
While the result
of~\citet{DGKKS20} was tailored to the ReLU activation, \Cref{thm:informal-unbounded-main-theorem}
works for a fairly broad class of unbounded activations (and a more general class of distributions). 
A key conceptual difference between the two results
lies in the overall approach: The prior work \citet{DGKKS20} used a {\em convex} surrogate for optimization.
In contrast, we leverage our structural result and directly optimize the natural non-convex objective.

Additionally, \Cref{thm:informal-unbounded-main-theorem} achieves significantly better
(and near-optimal) sample complexity as a function of both $d$ and $1/\eps$. In more detail,
the algorithm of \citet{DGKKS20} had sample complexity $\Omega(d/\eps)$ 
(their results are phrased for the special case that $W=1$), while our algorithm has sample complexity
$\wt{O}(d) \polylog(1/\eps)$ --- i.e., near-linear in $d$ and polylogarithmic in $1/\eps$.

\section{Preliminaries}\label{sec:prelims}
\paragraph{Basic Notation}
For $n \in \Z_+$, let $[n] \eqdef \{1, \ldots, n\}$.  We use small boldface characters for vectors
and capital bold characters for matrices.  For $\bx \in \R^d$ and $i \in [d]$, $\bx_i$ denotes the
$i$-th coordinate of $\bx$, and $\|\bx\|_2 \eqdef (\littlesum_{i=1}^d \bx_i^2)^{1/2}$ denotes the
$\ell_2$-norm of $\bx$.  We will use $\bx \cdot \by $ for the inner product of $\bx, \by \in \R^d$
and $ \theta(\bx, \by)$ for the angle between $\bx, \by$.  We slightly abuse notation and denote
$\vec e_i$ the $i$-th standard basis vector in $\R^d$.  We will use $\1_A$ to denote the
characteristic function of the set $A$, i.e., $\1_A(\x)= 1$ if $\x\in A$ and $\1_A(\x)= 0$ if
$\x\notin A$.  For vectors $\vec v,\vec u\in \R^d$, we denote $\vec v^{\perp_{\vec u}}$ the
projection of $\vec v$ into the subspace orthogonal to $\vec u$, furthermore, we denote $\vec
v^{\|_{\vec u}}$ the projection of $\vec v$ into the direction $\vec u$, i.e., $\vec v^{\|_{\vec
u}}\eqdef ((\vec v\cdot\vec u)\vec u)/\|\vec u\|_2^2$.

\paragraph{Asymptotic Notation}
We use the standard $O(\cdot), \Theta(\cdot), \Omega(\cdot)$ asymptotic notation. We also use
$\wt{O}(\cdot)$ to omit poly-logarithmic factors.  We write $E \gtrsim F$, two non-negative
expressions $E$ and $F$ to denote that \emph{there exists} some positive universal constant $c > 0$
(independent of the variables or parameters on which $E$ and $F$ depend) such that $E \geq c \, F$.
In other words, $E = \Omega(F)$.
For non-negative expressions $E, F$ we write $E \gg F$ to denote that $E \geq C \, F$, where
$C>0$ is a \emph{sufficiently large} universal constant (again independent of the parameters of $E$
and $F$).  The notations $\lesssim, \ll$ are defined similarly.

\paragraph{Probability Notation}
We use $\E_{x\sim \D}[x]$ for the expectation of the random variable $x$ according to the
distribution $\D$ and $\pr[\mathcal{E}]$ for the probability of event $\mathcal{E}$. For simplicity
of notation, we may omit the distribution when it is clear from the context.  For $(\x,y)$
distributed according to $\D$, we denote $\D_\x$ to be the distribution of $\x$ and $\D_y$ to be the
distribution of $y$. For unit vector $\vec v\in \R^d$, we denote $\D_{\vec v}$ the distribution of
$\x$ on the direction $\vec v$, i.e., the distribution of $\x_{\vec v}$. For a set $B$ and a
distribution $\D$, we denote $\D_B$ to be the distribution $D$ conditional on $B$.

\section{The Landscape of the $L_2^2$ Loss}
\label{main-sec:bounded-plain-landscape}
In this section, we present our results on the landscape of the $L_2^2$ loss for the sigmoidal activation functions 
of \Cref{def:well-behaved-bounded-intro} and the unbounded activation functions of \Cref{def:well-behaved-unbounded-intro}.
Before we proceed, we remark again that \Cref{def:well-behaved-bounded-intro} models a general class
of bounded activation functions, including, for example, the logistic activation, $\sigma(t) =
1/(1+e^{-t})$, the hyperbolic tangent, $\sigma(t) = \tanh(t)$, the Gaussian error function, and the
ramp activation, see \Cref{fig:tanh}.

\pgfplotsset{every axis title/.append style={at={(0.5,-0.25)}}} 
\usepgfplotslibrary{groupplots}
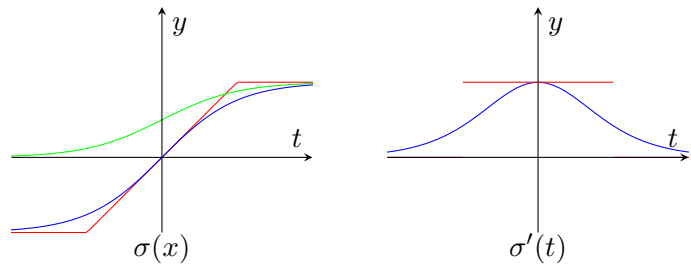
\begin{figure}[h]
\centering
\begin{tikzpicture}
 \begin{groupplot}[group style={group size=3 by 2},height=3cm,width=3cm,
 restrict y to domain=-4:4,xlabel=$t$,ylabel=$y$, legend pos=north west,axis y line =middle,
 axis x line =middle,
 				axis on top=true,
 				xmin=-2,
 				xmax=2,
 				ymin=-1,
 				ymax=2,
 				y=1cm,
                x=1cm,
                   yticklabels={,,},
 				    xticklabels={,,},
 				      ytick=\empty,
 				        xtick=\empty,]
				  \nextgroupplot[title=${\sigma(x)}$]
				\addplot[color=red, samples=100, domain=-4:0]
				{max(x,-1)};
				\addplot[color=red, samples=100, domain=0:4]
				{min(x, 1)};
				\addplot[color=blue, samples=100, domain=-4:4]
				{tanh(x)};
				\addplot[color=green, samples=100, domain=-4:4]
				{1/(1+e^(-2*x)};
\nextgroupplot[title=${\sigma'(t)}$]
				\addplot[color=blue, samples=100, domain=-4:4]
				{(1/cosh(x))^2};
				\addplot[color=red, samples=100, domain=-1:1]
				{1};
				\addplot[color=red, samples=100, domain=-4:-1]
				{0};
				\addplot[color=red, samples=100, domain=1:4]
				{0};
\end{groupplot}

\end{tikzpicture}
    \caption{
    The $\tanh(\cdot)$ activation is $(0.4, 1, 1.4)$-sigmoidal.  The ramp activation is $(1, 1,
    3)$-sigmoidal. The logistic activation is $(0.19, 1, 1)$-sigmoidal.  In the right figure, we plot
    the derivative of the ramp activation that is simply a rectangular function $\1\{|t| \leq 1\}$
    (drawn in red) and the derivative of $\tanh(t)$ (drawn in blue). Notice that both decay at least
    exponentially fast: the derivative of the ramp activation is non-zero only in the interval
    $[-1,1]$ and the derivative of $\tanh(t)$ decays exponentially fast, i.e., it is always smaller
    than $(4/3) e^{-|x|}$.
    } 
    \label{fig:tanh}
\end{figure}

\begin{figure}
\centering
\includegraphics[width=0.43\textwidth]{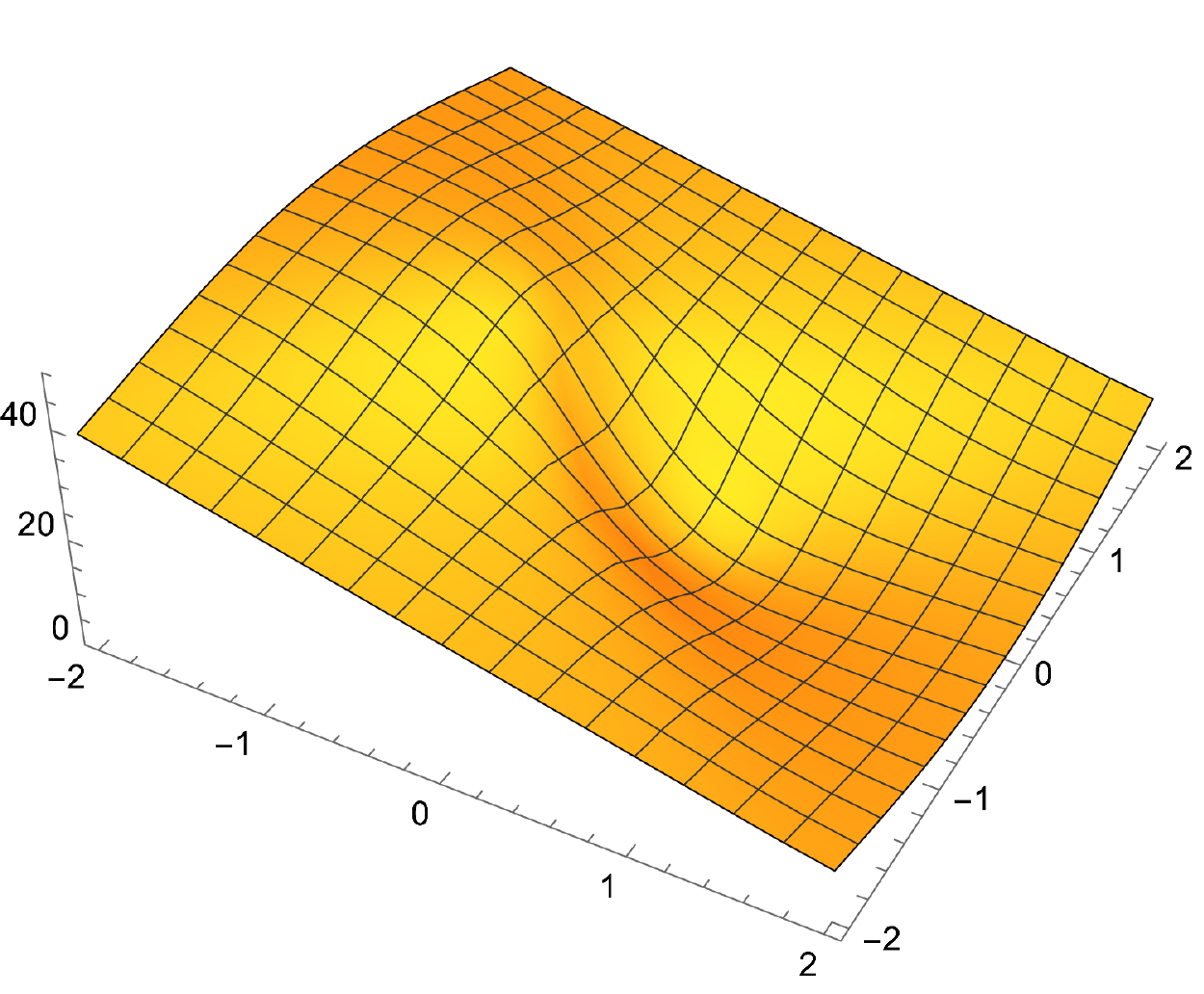}
~~
\includegraphics[width=0.43\textwidth]{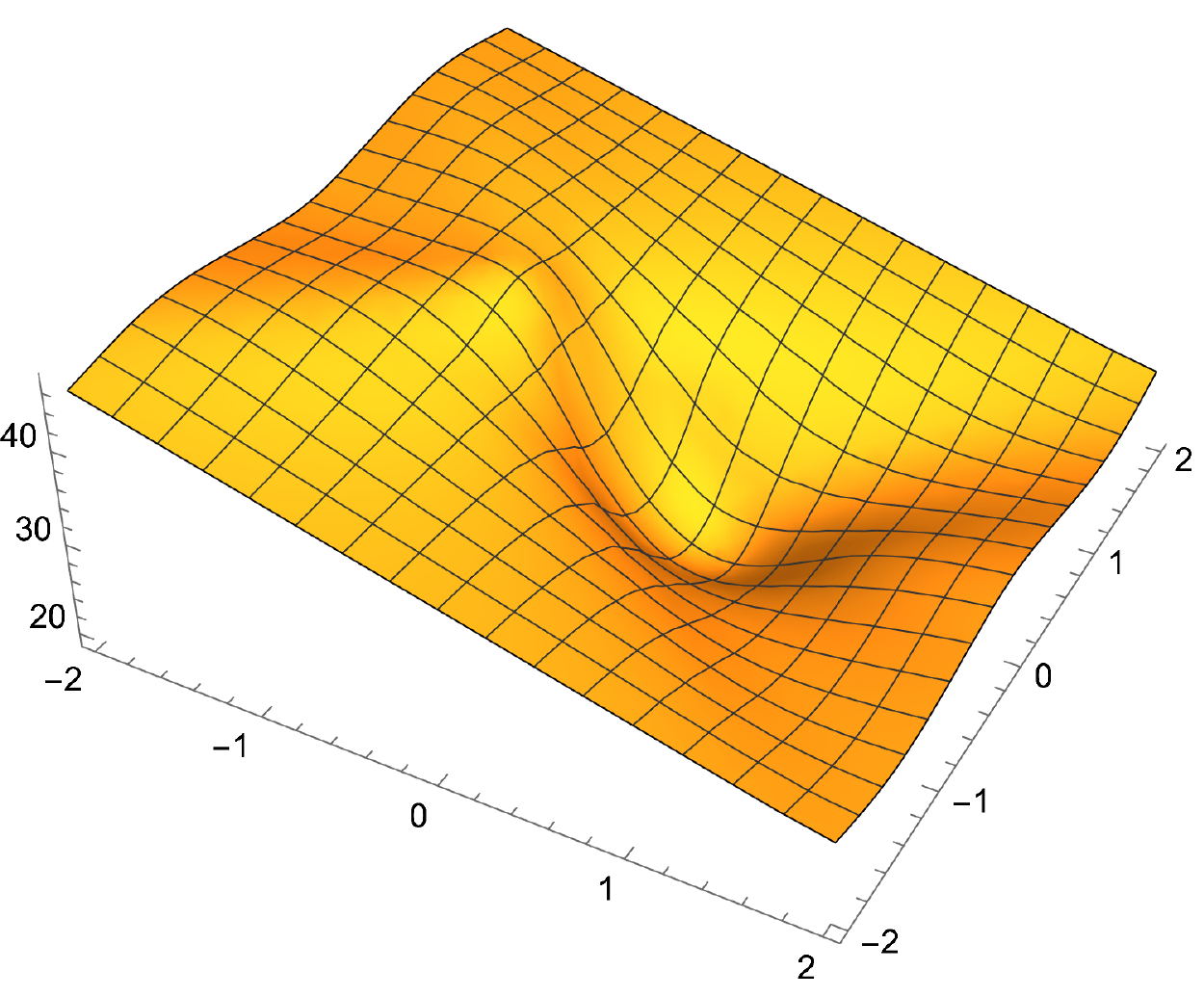}
\caption{The non-convex optimization landscape of the ``vanilla'' $L_2^2$ loss.
The activation is a ramp function $r(t)$, defined in \Cref{eq:ramp-definition}.  The $\x$-marginal
is the uniform distribution on the square $[-2,2] \times [-2,2]$.  The ``true'' underlying weight
vector is $\vec w^\ast = (1,0)$.  In the left figure, we plot the noiseless population objective
$F(\vec w)$ as a function of $\vec w \in[-2,2] \times [-2,2]$, where $y = r(\vec w^\ast \cdot \x)$.
We observe that even though the objective is non-convex, in this case, the only stationary point is
the true weight vector $\vec w^\ast$.  In the right figure, we introduce noise and observe that 
the objective has another stationary point, which in fact is $-\vec w^\ast$.  Finally,
we notice that the landscape becomes more ``flat'' as we move away from the origin and the ``bad'' stationary
point is a local-minimum that the ``noise'' was able to create in a particularly flat region.
}
\label{main-fig:non-convex-landscape}
\end{figure}

 We first show that we can construct noisy instances that have ``bad'' stationary points, i.e.,
 local-minima whose $L_2^2$ loss is $\omega(\eps)$.  Perhaps surprisingly, this is the case even
 when the underlying $\x$-marginal is the standard normal and the level of corruption $\eps$ of the
 corresponding labeled instance $\D$ is arbitrarily small.  Moreover, the sigmoidal activation used
 in the construction of the counterexample is very simple (in particular, we use the ramp
 activation).  We show that, even though the constructed instance is only $\eps$-corrupted, there
 exists a local minimum whose $L_2^2$ loss is at least $\omega(\eps)$ (and in fact larger than some
 universal constant).  The proof of the following result can be found in
 \Cref{app:bounded-plain-landscape}; an example of the noisy $L_2^2$ landscape is shown in
 \Cref{main-fig:non-convex-landscape}.
\begin{proposition}[Vanilla $L_2^2$ has ``Bad'' Local-Minima] 
  \label{theorem:intro_bad_stationary}
  For any $\eps \in (0,1]$, there exists a well-behaved sigmoidal activation $\sigma(\cdot)$
  and an $\eps$-corrupted distribution $\D$ on $\R^d \times \{\pm 1 \}$ with standard Gaussian $\x$-marginal 
  such that $F^{\D,\sigma}(\cdot)$ has a local minimum $\vec u$ with  $F^{D,\sigma}(\vec u) \geq 1/2$.
  \end{proposition}

Our positive result shows that by using a regularized version of the $L_2^2$ loss, we can guarantee that
all stationary points have error within a constant multiple of $\eps$.  Before stating our formal
result, we first define the norm that we will use frequently together with its corresponding dual norm.
In fact, we show that this norm characterizes the landscape of the (regularized) $L_2^2$ loss, in the
sense that minimizing the gradient with respect to its dual norm 
will give a point with small error.  

\begin{definition}[$\vec w$-weighted Euclidean Norm] 
  \label{main-def:weighted-euclidean}
  Given some vector $\vec u\in \R^d$, we define its weighted Euclidean 
norm with respect to a non-zero vector $\vec w \in \R^d$ to be 
\[
\| \vec u\|_{\vec w} 
 = 
 \frac{\|\proj_{\vec w} \vec u \|_2}{\|\vec w\|_2^{3/2}}+
 \frac{\| \proj_{\vec w^\perp} \vec u \|_2}{\|\vec w\|_2^{1/2}}\;.
\]
We also define the dual norm of $\|\cdot \|_{\vec w}$ 
as follows:
\[
\| \vec v \|_{*, \vec w} =
\max ( \|\proj_{\vec w} \vec v\|_2 \|\vec w\|_2^{3/2},
\|\proj_{\vec w^\perp} \vec v\|_2 \|\vec w\|_2^{1/2})
\,.
\]
\end{definition}

The main intuition behind the norm of \Cref{main-def:weighted-euclidean} is the following: 
for well-behaved distributions and sigmoidal activations, the ``noiseless'' $L_2^2$ loss $\E_{\x \sim
\D}[(\sigma(\vec w \cdot \x) - \sigma(\vec w^\ast \cdot \x))^2] $ behaves similarly to the non-convex
function $\vec w \mapsto \|\vec w - \vec w^\ast\|_{\vec w}^2$.  For more details and intuition on
why this norm naturally appears in our results, we refer to (the proofs of)
\Cref{main-lem:noise-contribution-new} and \Cref{lem:bound-difference-bounded}.

\begin{remark}[Distribution/Activation parameters]\label{rem:parameters}
  \emph{
Before we present our main structural result, we would like to revisit the parameters of
\Cref{def:bounds} and \Cref{def:well-behaved-bounded-intro}.  We observe that an 
$(L, R)$-well-behaved distribution is also $(L', R')$-well-behaved for any $L' \leq L,  R' \leq R$. 
Therefore, without loss of generality (and to simplify the presentation), we shall assume that
$L, R \in (0, 1]$.  For the same reason, for sigmoidal activations, we will assume that $\xi\in[1,\infty)$
and $\tau, \mu \in (0, 1]$. Similarly, for unbounded activations we assume $\lambda \in[1,\infty), \alpha
\in (0,1]$.  }
\end{remark}

We now state our main structural result for sigmoidal activations, 
namely that all stationary points of the $\ell_2$-regularized
$L_2^2$ objective are approximately optimal solutions.  

\begin{theorem}[Stationary Points of Sigmoidal Activations]
\label{thm:stationary-points-regular} 
Let $\D$ be an $\eps$-corrupted, $(L, R)$-well-behaved distribution on $\R^d \times \R$ and $\sigma$
be a $(\tau, \boundc, \boundd)$-sigmoidal activation.  Set $\kappa=L^6 R^6\boundc^3\tau^4/\boundd^2$ and $\rho=C\eps^3/\kappa^5$, where $C>0$ is a
universal constant, and define the
$\ell_2$-regularized objective as
\( F_\rho(\vec w) =  (1/2)\E_{(\x,y)\sim \D}[(\sigma(\vec w\cdot \x)-y)^2] + (\rho/2) \|\vec w\|_2^2 \,.  \) 
For some sufficiently small universal constant $c>0$, we have the following
\begin{itemize}
    \item If $\|\vec w\|_2\leq 2/R$ and $\|\nabla F_\rho(\vec w)\|_2 \leq  c \sqrt{\eps}$, then $F(\vec w) = \eps \cdot
    \poly(1/\kappa) $.
    \item  If $\|\vec w\|_2\geq 2/R$ and $\|\nabla F_\rho(\vec w)\|_{*,\vec w} \leq c
\sqrt{\eps}$, then $F(\vec w) = \eps\cdot  \poly(1/\kappa)$.
\end{itemize}
\end{theorem}

\Cref{thm:stationary-points-regular} shows that the $L_2^2$ objective behaves differently when 
$\vec w$ lies close vs far from the origin. The landscape becomes more ``flat'' as we move further away 
from the origin, and in order to achieve loss $\eps$ we have to make the dual of the weighted $\vec w$-norm of
the gradient small; for example, the component orthogonal to $\vec w$ has to be smaller than
$\sqrt{\eps}/\|\vec w\|_2^{1/2}$.

At a high level, we prove \Cref{thm:stationary-points-regular} in two main steps. First, we analyze the
population $L_2^2$ loss without a regularizer, and show that all ``bad'' stationary points are in
fact at distance $\Omega(1/\eps)$ from the origin.   Since the non-regularized (vanilla) gradient
field becomes very flat far from the origin, adding a small amount of noise potentially creates bad
local minima (see also \Cref{main-fig:non-convex-landscape}).  We show that by adding an appropriate
$\ell_2$ regularizer, we essentially introduce radial gradients pulling towards the origin whose
contribution is large enough to remove bad stationary points, while ensuring that the ``good''
stationary points do not change by a lot.

Before proceeding to the proof of \Cref{thm:stationary-points-regular}, we first highlight an
intricacy of the landscape of the $L_2^2$ loss for sigmoidal activations: 
the $L_2^2$ loss may be ``minimized'' for some direction of infinite length. 
Consider, for example, the case where the
activation is the logistic loss $\sigma(t) = 1/(1 + e^{-t})$, the label $y$ is equal to $\sgn(\vec
w^\ast \cdot \x)$ for some unit vector $\vec w^\ast$, and the $\x$-marginal is the standard normal
distribution.  Then, we have that for any fixed vector $\vec w$ it holds that 
$\E_{(\x,y) \sim \D}[ (\sigma(\vec w \cdot \x) - y)^2] >0$
and 
\[
  \lim_{t \to +\infty } \E_{(\x,y) \sim \D}[ (\sigma(t ~ \vec w^\ast \cdot \x) - y)^2]  = 0 \,.
  \]
We prove the following lemma showing that even though the true ``minimizer'' may be a vector of
infinite length, there exist almost optimal solutions inside a ball of radius $O(1/\eps)$. 

\begin{lemma}[Radius of Approximate Optimality of Sigmoidal Activations]
  \label{lem:radius-approx}
Let $\D$ be an $(L,R)$-well-behaved distribution in $\R^d$ and let $\sigma(t)$ be a $(\tau, \boundc,
\boundd)$-sigmoidal activation function.  There exists a vector $\vec v$ with $\|\vec v\|_2 \leq
1/\eps $ and $F(\vec v) \leq (1+O(\frac{\xi}{ \mu L }))\eps $.
\end{lemma}

To prove \Cref{thm:stationary-points-regular} we will show that for most vectors $\vec w$ the
gradients of vectors that have large error, i.e., $\vec w$ with $F(\vec w) \geq \Omega(\eps)$, will
have large contribution towards some optimal vector $\vec w^\ast$, i.e., $\nabla F_{\rho}(\vec w)
\cdot (\vec w - \vec w^\ast) > 0$. This is a more useful property than simply proving that the norm
of the gradient is large, and we will later use it to obtain an efficient algorithm.  However, as we
already discussed, this is not always the case: we can only show that the gradient field ``pulls''
towards some optimal solution only inside a ball of radius roughly $1/\eps$ around the origin (see
Cases 1,2 of \Cref{prop:gradient-bound-regular}).  Outside of this ball, we show that the
regularizer will take effect and pull us back into the ball where the gradient field helps to
improve the guess; this corresponds to Case 3 of \Cref{prop:gradient-bound-regular}.
In particular, we show that the projection of the gradient of the $L_2^2$ objective on the direction
$\vec w - \wstar$ is proportional to the standard $\ell_2$ distance of $\vec w - \wstar$, when $\vec
w$ is close to the origin, and proportional to the $\vec w$-weighted Euclidean distance of
\Cref{main-def:weighted-euclidean}, when $\vec w$ is far from the origin.

\begin{proposition}[Gradient of the Regularized $L_2^2$ Loss]
\label{prop:gradient-bound-regular}
 Let $\D$ be an $(L,R)$-well-behaved distribution and  define $F_\rho(\vec w)=(1/2)\Ey
 [(\sigma(\vec w\cdot \x)-y)^2]+(1/2)\rho \|\vec w\|_2^2$, where $\sigma$ is a $(\tau, \boundc,
 \boundd)$-sigmoidal activation and $\rho>0$. Let $\eps\in(0,1)$ and let $\wstar \in \R^d$ such that
 $F(\wstar)\leq \eps$ and $\|\wstar\|_2\leq U/\eps$, for some $U\geq 0$. 
 Furthermore, set $\kappa=L^6 R^6\boundc^3\tau^4/\boundd^2$ and $\rho=C\eps^3/\kappa^5$, 
 where $C\geq 0$ is a sufficiently large universal constant.  
 There exists a universal constant
 $c' > 0$, such that for any $\vec w\in \R^d$, we have:
\begin{enumerate}
   \item 
When $\|\vec w\|_2\leq 2/R$ and $\|\wstar-\vec w\|_2\geq \sqrt{\eps}/(c'\kappa^5)$, then
 \(
\nabla F_\rho(\vec w)\cdot(\vec w-\wstar) \geq c'\sqrt{\eps}\|\wstar-\vec w\|_2 \,. \)
 \item When $ 2/R\leq\|\vec w\|_2 \leq  c'\kappa/\eps$ and either 
 $\|\vec w-\wstar\|_{\vec w}\geq \sqrt{\eps} (U/(c'\kappa^5)) $ or $\|\vec w\|_2\geq 2\|\wstar\|_2$, then
\(
   \nabla F_\rho(\vec w)\cdot(\vec w-\wstar)\geq c'\sqrt{\eps}\|\wstar-\vec w\|_{\vec w}
   \,.
\)
    \item When $\|\vec w\|_2\geq  c'\kappa/(2 \eps)$, then
 \(
\nabla F_\rho(\vec w)\cdot\vec w \geq  c'\sqrt{\eps}\|\vec w\|_{\vec w}\;.
 \)
\end{enumerate}
\end{proposition}

Before we prove \Cref{prop:gradient-bound-regular}, we show that without adding the regularizer the
gradient field of the ``vanilla'' $L_2^2$ loss points towards some optimal solution $\vec w^\ast$, as
long as the guess $\|\vec w\|_2$ is not very far from the origin.

\begin{proposition}[Gradient of the ``Vanilla'' $L_2^2$ Loss (Inside a Ball)]
\label{prop:gradient-bound-ramp}
 Let $\D$ be an $(L,R)$-well-behaved distribution and let $\sigma$ be a $(\tau, \boundc,
 \boundd)$-sigmoidal activation.  Let $\eps\in(0,1)$ be smaller than a sufficiently small multiple
 of $L^2R^6\tau^4/\boundd^2$ and let $\wstar \in \R^d$ with $F(\wstar)\leq \eps$.   Set $\kappa=L^6
 R^6\boundc^3\tau^4/\boundd^2$.  There exists a universal constant $c' > 0$, such that for any $\vec
 w\in \R^d$ with $\| \vec w\|_2\leq c'\kappa/\eps$, we have:
\begin{enumerate}
\item 
 When $\|\vec w\|_2\leq 2/R$ and $\|\wstar-\vec w\|_2\geq \sqrt{\eps}\boundd/(c'L R^4\tau^2)$, then
 \[
\nabla F(\vec w)\cdot(\vec w-\wstar) \geq c'\sqrt{\eps}\|\wstar-\vec w\|_2 \,. \]
\item  When $\|\vec w\|_2 \geq 2/R$ and either $\|\wstar-\vec w\|_{\vec w}\geq \sqrt{\eps/\kappa}$ or $\|\vec w\|_2\geq 2\|\wstar\|_2$, then
\[
\nabla F(\vec w)\cdot(\vec w-\wstar) \geq c' \sqrt{\eps}\|\wstar-\vec w\|_{\vec w}\,. \]
  \end{enumerate}
\end{proposition}

\paragraph{Proof Sketch of \Cref{prop:gradient-bound-ramp}}
    Let $\wstar\in\R^d$ be a target weight vector with $F(\wstar)\leq \eps$.  As we discussed, we
    want to show that when $\vec w$ is far from $\wstar$ (in $\vec w$-weighted Euclidean norm), then
    the inner product $\nabla F(\vec w)\cdot(\vec w -\wstar)$ is strictly positive. We can decompose
    this inner product in two parts: $I_1$ that depends on how ``noisy'' the labels are and $I_2$
    that corresponds to the contribution that we would have if all labels were ``clean'', i.e., $y =
    \sigma(\vec w^\ast \cdot \x)$,
\begin{align}
    \nabla F(\vec w)\cdot(\vec w-\wstar)&=\underbrace{\E[(\sigma(\wstar\cdot \x)-y)\sigma'(\vec w\cdot \x)(\vec w\cdot \x-\wstar \cdot \x)]}_{I_1} 
    \nonumber
    \\ &+\underbrace{\E[(\sigma(\vec w\cdot \x)-\sigma(\wstar\cdot \x))\sigma'(\vec w\cdot \x)(\vec w\cdot \x-\wstar \cdot \x)]}_{I_2}\;.
    \label{main-eq:gradient-decomposition-bounded}
\end{align}
We crucially use the monotonicity of the activation $\sigma(\cdot)$:  since $\sigma(\cdot)$ is
non-decreasing, we have that $\sigma'(t) \geq 0$ and $(\sigma(\vec w \cdot \x) - \sigma(\wstar \cdot
\x)) (\vec w \cdot \x - \wstar\cdot \x) \geq 0$.  These facts immediately imply that $I_2 \geq 0$.
In what follows, we will show that $I_2$ is strictly positive.

In the worst case, the noisy term $I_1$ will be negative, i.e., the noise will try to make the
gradient point in the wrong direction (move $\vec w$ ``away'' from the target $\wstar$).  In the
first part of the proof, we show that $|I_1|$ cannot be too large: the positive contribution $I_2$
is much larger than the contribution of the noisy term $|I_1|$.  Apart from the monotonicity of the
activation $\sigma(\cdot)$, we will also rely on the fact that if we view its derivative as a
distribution, it satisfies anti-concentration and anti-anti-concentration properties, i.e.,
$\sigma'(t) \geq \tau$ for all $t \in [-1,1]$ and $\sigma'(t) \leq \xi e^{-\mu |t|}$ for all $t \in
\R$, see \Cref{def:well-behaved-bounded-intro}. A simple sigmoidal activation is the ramp
activation, defined as follows:
\begin{align}
  \label{eq:ramp-definition}
  r(t) \triangleq (-1)  ~ \1\{ t < -1\}
  + t ~ \1\{ |t| \leq 1\} + (+1) ~ \1\{ t > 1\} \,.
\end{align}
We observe that the ramp activation is $(1,e,1)$-sigmoidal since its derivative is $r'(t) = 1\{ |t|
\leq 1\}$ and vanishes exactly outside the interval $[-1,1]$.  In the formal proof, we show
how to reduce the analysis of general sigmoidal activations to the ramp activation.  To keep
this sketch simple, we will focus on the ramp activation.

\paragraph{Estimating the Contribution of the Noise}
We start by showing that the noise cannot affect the gradient by a lot, i.e., we bound the
contribution of $I_1$. We prove the following lemma.
\begin{lemma}\label{main-lem:noise-coontribution-ramp}
 Let $\D$ be a well-behaved distribution. For any vector $\vec w\in \R^d$ it holds
 that
 \(
 |I_1| \lesssim \sqrt{\eps} ~ \min(\|\wstar-\vec w\|_{\vec w},\|\wstar-\vec w\|_2 )\;.
 \)
\end{lemma}
\noindent \emph{Proof.} (Sketch)
Using the \CS\ inequality, we obtain:
\begin{align}
\label{main-eq:I_1-bound}
   |I_1|&\leq
   \E[|(r(\wstar\cdot \x)-y)r'(\vec w\cdot \x)(\vec w\cdot \x-\wstar \cdot \x)|]
   \nonumber
   \\
    &\leq   (\E[(r(\wstar\cdot \x)-y)^2])^{
    1/2}
    (\E[(\vec w\cdot \x-\wstar \cdot \x)^2r'(\vec w\cdot \x)^2])^{1/2}
    \nonumber
    \\
    &= \sqrt{F(\wstar)}  
    ~
    (\E[(\vec w\cdot \x-\wstar \cdot \x)^2r'(\vec w\cdot \x)^2])^{1/2}\nonumber
    \\   &\leq \sqrt{\eps}  
    ~
    (\E[(\vec w\cdot \x-\wstar \cdot \x)^2r'(\vec w\cdot \x)^2])^{1/2} \,.
\end{align}
We proceed to bound the term  $\E[(\vec w\cdot \x-\wstar \cdot \x)^2r'(\vec w\cdot \x)^2]$.  Note
that we can use the global upper bound on the derivative of the activation function, i.e., that
$r'(t) \leq e$ for all $t \in \R$.  However, this would only result in the bound $\E[(\vec w\cdot
\x-\wstar \cdot \x)^2r'(\vec w\cdot \x)^2] \leq O(\|\vec w - \wstar\|_2^2)$.  For sigmoidal
activation functions, we need a tighter estimate that takes into account the fact that the functions
have exponential tails outside of the interval $[-1, 1]$.  In particular, for the ramp function, we
have that as $\|\vec w\|_2$ becomes larger the derivative $r'(\vec w \cdot \x ) = \1\{ |\vec w \cdot
\x | \leq 1 \}$ becomes a very thin band around the origin.  Therefore, by the anti-concentration
and anti-anti-concentration of well-behaved distributions (see \Cref{def:bounds}), the
aforementioned integral decays to $0$ as $\|\vec w\|_2 \to \infty$ at a rate of $1/\|\vec w\|_2$.  
\Cref{main-lem:noise-coontribution-ramp} follows from combining \Cref{main-eq:I_1-bound} along with 
the following lemma.  
\begin{lemma}\label{main-lem:noise-contribution-new}
 Let $\D$ be a well-behaved distribution. For any vectors $\wstar,\vec w\in \R^d$, it holds
 that
\(
\Exx[(\vec w \cdot \x - \wstar \cdot \x)^2 (r'(\vec w\cdot \x))^2]
\lesssim \min \left(\|\wstar-\vec w\|_{\vec w}^2 , \|\wstar-\vec w\|_2^2 \right)\;.
\)
\end{lemma}
We will provide a proof of the above lemma because it highlights why the weighted Euclidean norm
of \Cref{main-def:weighted-euclidean} captures the non-convex geometry of the optimization landscape.
\begin{proof}(Sketch)
Here we assume that $\D_\x$ is the standard $d$-dimensional Gaussian; see
\Cref{lem:noise-contribution-new} for the formal version.  First note that 
$ \Exx[ (\vec w \cdot \vec x- \wstar \cdot \vec x)^2  r'(\vec w \cdot \vec x)^2 ] 
\leq  \Exx[ ((\vec w-\wstar)\cdot \x)^2]= \|\vec w-\wstar\|_2^2$, 
where we used the fact that since the distribution $\D_\x$ is isotropic,
i.e., for any vector $\vec u\in \R^d$ it holds that $\Exx[(\vec u \cdot \x)^2] = \|\vec u\|_2^2$.
This bound is tight when $\|\vec w\|_2$ is small (see Case 1 of \Cref{prop:gradient-bound-ramp}).
When $\|\vec w\|_2\geq 1$, we see that the upper bound decays with $\| \vec w\|_2$.  
Let $\vec q=\wstar-\vec w$ and denote by $G$ the one-dimensional standard Gaussian. 
We can decompose the difference $\vec q$ to its component parallel to $\vec w$: $\vec q^{\|_\vec w}$ 
and its component orthogonal to $\vec w$: $\vec q^{\vec w^\perp}$. 
From the Pythagorean theorem, we have that 
$(\vec q \cdot \vec x)^2 = (\vpar{q}{w} \cdot \x)^2 + (\vperp{q}{w}\cdot\x)^2$. 
Using this we get:
\begin{align*}
&\E[ (\vec q\cdot \x)^2\1\{|\vec w\cdot \x|\leq 1\}]
     = \E[ (\vpar{q}{w} \cdot \x)^2\1\{|\vec w\cdot \x|\leq 1\} ]+ \E[(\vperp{q}{w}\cdot\x)^2) \1\{|\vec w\cdot \x|\leq 1\} ] \\
&= \|\vpar{q}{w}\|_2^2\E_{z\sim G}[z^2\1\{|z|\leq \|\vec w\|_2^{-1}\} ]+  
          \|\vperp{q}{w}\|_2^2 \pr_{z\sim G}[|z|\leq \|\vec w\|_2^{-1} ]
     \lesssim    
     \frac{ \|\vec q^{\|_{\vec w}}\|_2^2}{\|\vec w\|_2^3} + 
      \frac{\|\vec q^{\perp_{\vec w}}\|_2^2}{\|\vec w\|_2}
      \lesssim \|\vec q\|_{\vec w}^2\;,
\end{align*}
where in the second equality we used that under the Gaussian distribution any two orthogonal
directions are independent, and in the last inequality we used that $a^2+b^2\leq (a+b)^2$ for
$a,b\geq 0$. Observe that the orthogonal direction is only scaled by roughly the probability of the
slice, which is $1/\|\vec w\|_2$, while the parallel component (which is restricted in the interval
$|z| \leq 1/\|\vec w\|_2$) is scaled by $1/\|\vec w\|_2^3$ (similarly to the fact that $\int_{-a}^a
t^2 d t = O(a^3))$.
\end{proof}

\paragraph{Estimating the Contribution of the ``Noiseless'' Gradient}
We now bound from below the contribution of the ``noiseless'' gradient, i.e., the term $I_2$ of
\Cref{main-eq:gradient-decomposition-bounded}.  
To bound $I_2$ from below, we consider several different cases depending on how far the vector $\vec
w$ is from the target vector $\wstar$; the full proof is rather technical.  In this sketch, we
shall only consider the case where the angle between the weight vectors $\vec w$ and $\wstar$ is
in $(0, \pi/2)$ and also assume that the norm of the guess $\vec w$ is larger than $2/R$, since this
is the regime where the norm of the gradient behaves similarly to the weighted Euclidean norm of
\Cref{main-def:weighted-euclidean}. Notice that when $\vec w$ is close to the target $\wstar$, then
its projection on the orthogonal complement of $\vec w$, i.e., $\|\proj_{\vec w^\perp}\wstar\|_2 $, 
will be small and also its projection on the direction of $\vec w$, i.e., $\|\proj_{\vec w}\vec
v\|_2$, will be close to $\|\vec w\|_2$.  In this sketch, we will show how to handle the case where
$\theta(\vec w,\wstar)\in(0,\pi/2)$, $\|\proj_{\vec w^\perp}\wstar\|_2 \leq 2/R$ and 
$\|\proj_{\vec w}\wstar\|_2\leq 2\|\vec w\|_2$, 
i.e., the case where $\vec w$ and $\wstar$ are not extremely far apart.  
\begin{lemma}\label{main-lem:lower-bound-square}
Let $\D$ be a well-behaved distribution.  For any vectors $\vec w,\wstar\in \R^d$ with 
$\|\vec w\|_2 > 2/R$, $\|\proj_{\vec w^\perp}\wstar\|_2 \leq 2/R$, $\|\proj_{\vec w}\wstar\|_2\leq 2\|\vec w\|_2$ and
$\theta(\vec w,\wstar) \in (0,\pi/2)$,
it holds that
\(
 I_2 \gtrsim  \|\vec w-\wstar\|_{\vec w}^2
\,.
    \)
\end{lemma}
We can now finish the proof of (one case of) \Cref{prop:gradient-bound-ramp}. 
From \Cref{main-lem:noise-coontribution-ramp}, if $\|\vec w\|_2\geq 2/R$, 
it holds that the noisy gradient term 
$|I_1| \lesssim \sqrt{\eps}\|\wstar-\vec w\|_{\vec w}$. 
Using \Cref{main-lem:lower-bound-square}, there exists
a universal constant $c>0$,  such that
$
    I_1+I_2 \geq c\|\wstar-\vec w\|_{\vec w}( 
    \|\wstar-\vec w\|_{\vec w}-\sqrt{\eps}/c)\gtrsim \sqrt{\eps}\|\wstar-\vec w\|_{\vec w}\;,
    $
where in the last inequality we used that $\|\wstar-\vec w\|_{\vec w}\gg \sqrt{\eps}$.
This completes the proof sketch of \Cref{prop:gradient-bound-ramp}. For the full
proof, we refer to \Cref{app:bounded-plain-landscape}.

\subsection{Proof of \Cref{prop:gradient-bound-regular}}
  Similarly to the statement of \Cref{prop:gradient-bound-regular}, in our proof we shall 
  consider several cases depending on how large $\|\vec w\|_2$ is.
For the rest of the proof, let $c'$ be the absolute constant of \Cref{prop:gradient-bound-ramp} and
denote by $\kappa=L^3 R^6\boundc^3\tau^4/\boundd^2$, $\Lambda_1=16$, $\Lambda_2=1/\kappa \geq \xi^{6/5}/(c'
\kappa L^2 \mu^6)^{1/5}$ and $K= \Lambda_2^{2.5}U/(\sqrt{c'\kappa}LR)$.  

First, we show Case 3 of \Cref{prop:gradient-bound-regular} and bound from below
the contribution of the gradient on the direction of $\vec w$, when the norm of $\vec w$ is large.
In other words, we show that in this case the gradient contribution of the regularizer is large
enough so that the gradient field of the regularized $L_2^2$-objective  ``pulls'' $\vec w$ towards
the origin.
\begin{claim}\label{clm:paral-contribution}
 If $\|\vec w\|_2\geq c'\kappa/(\eps\Lambda_1)$, then 
 \(
\nabla F_\rho(\vec w)\cdot\vec w \geq  \sqrt{\eps} \|\vec w\|_{\vec w} \,. 
\)
\end{claim}
\begin{proof}
We have that 
\[
\nabla F_\rho(\vec w)\cdot\vec w = \Ey [(\sigma(\vec w\cdot \x)-y)\sigma'(\vec w\cdot \x)\vec w\cdot\x]+\rho \|\vec w\|_2^2\;.
\]
First we bound from below the quantity $\Ey [(\sigma(\vec w\cdot \x)-y)\sigma'(\vec w\cdot \x)\vec w\cdot\x]$.
We need to bound the maximum value that $\sigma$ can obtain.
\begin{fact}\label{clm:upper-bound-value-bounded}
Let $\sigma$ be a $(\tau,\boundc, \boundd)$-sigmoidal activation, then  $|\sigma(t_1)-\sigma(t_2)|\leq 2\boundd/\boundc$ for any $t_1,t_2\in \R$.
\end{fact}
\begin{proof}
Using the fundamental theorem of calculus, for any $t_1,t_2\in \R$, it holds
\[
|\sigma(t_1)-\sigma(t_2)|=\left|\int_{t_1}^{t_2}\sigma'(t)\d t\right|\leq \int_{-\infty}^{\infty}\sigma'(t)\d t\leq 2\int_{0}^{\infty}\boundd e^{-t \boundc}\d t\leq 2\boundd/\boundc\;,
\]
where we used that $\sigma$ is non-decreasing and that $\sigma'(t)\leq \boundd e^{-t \boundc}$.
\end{proof}
\noindent Using \Cref{clm:upper-bound-value-bounded}, we have that
\begin{align*}
    \Ey [(\sigma(\vec w\cdot \x)-y)\sigma'(\vec w\cdot \x)\vec w\cdot\x]
    &\geq -\Ey [|(\sigma(\vec w\cdot \x)-y)||\sigma'(\vec w\cdot \x)\vec w\cdot\x|]
    \\&\geq - 2 \frac{\boundd}{\boundc}\Exx [\sigma'(\vec w\cdot \x)^2|\vec w\cdot\x|]
    \\&\geq -2 \frac{\boundd^3}{\boundc}\Exx [\exp(-2\boundc|\vec w\cdot \x|\leq 1)|\vec w\cdot\x|]
    \\&\geq -\frac{\boundd^3}{L\boundc^3\|\vec w\|_2}\;,
\end{align*}
where in the second inequality we used that the maximal difference of $(\sigma(\vec w\cdot \x)-y)$ is less than $2\boundd/\boundc$, and in the third the upper bound in the derivative of $\sigma$. Therefore, we have that
\begin{align*}
    \nabla F_\rho(\vec w)\cdot\vec w &\geq (1/2)\rho \|\vec w\|_2^2 + (1/2)\|\vec w\|_2^2 \left(\rho -\frac{2\boundd^3}{L\boundc^3\|\vec w\|_2^3}\right)
    \\ &\geq (1/2)\rho \|\vec w\|_2^2\;,
\end{align*}
where in the last inequality we used that $\rho\geq \frac{2\eps^3\Lambda_1^3 \boundd^3}{c'^3\kappa^3\boundc^3L}\geq \frac{2\boundd^3}{L\boundc^3\|\vec w\|_2^3}$.

Therefore, we have that $ \nabla F_\rho(\vec w)\cdot\vec w \geq (1/2)\rho \|\vec w\|_2^{2.5}\|\vec w\|_{\vec w}$. By using that $\|\vec w\|_2\geq c'\kappa/(\eps\Lambda_1)$ and $\rho\geq \frac{2\eps^3\Lambda_1^3 \boundd^3}{c'^3\kappa^3\boundc^3L}$, we get that
\[
 \nabla F_\rho(\vec w)\cdot\vec w \geq \sqrt{\eps \Lambda_1}\frac{\boundd^3}{\boundc^3 \sqrt{c'\kappa}L} \|\vec w\|_{\vec w}\geq 
 \sqrt{\eps }\|\vec w\|_{\vec w}\;,
\]
where we used that $\boundd,\Lambda_1\geq 1, \boundc,\kappa,L\leq 1$ (see \Cref{rem:parameters}) and
this completes the proof of \Cref{clm:paral-contribution}.

\end{proof}
Next we bound from below the contribution of the gradient in the direction $\vec w- \vec w^\ast$ when
$\vec w$ is close to the origin, i.e., when $\|\vec w\|_2\leq 2/R$ (first case of
\Cref{prop:gradient-bound-regular}).  In this case, we show that our choice of regularization $\rho$,
ensures that when $\vec w$ is not very large, then the gradient behaves qualitatively similarly to
that of the vanilla $L_2^2$ objective, i.e., as in \Cref{prop:gradient-bound-ramp}.
\begin{claim}\label{clm:small-norm-2} 
 If $ \|\vec w\|_2\leq 2/R$ and $\|\wstar-\vec w\|_2\geq \sqrt{\eps}\boundd/(c'L R^4\tau^2)$, then
 \[
\nabla F_\rho(\vec w)\cdot(\vec w-\wstar) \geq (c'/2)\boundd\sqrt{\eps}\|\wstar-\vec w\|_2 \,. \]
\end{claim}
\begin{proof}
From \Cref{prop:gradient-bound-ramp}, we have that if $ \|\vec w\|_2\leq 2/R$ and $\|\wstar-\vec w\|_2\geq \sqrt{\eps}\boundd/(c'L R^4\tau^2)$, then
$ \nabla F(\vec w)\cdot(\vec w-\wstar) \geq c'\sqrt{\eps}\|\wstar-\vec w\|_2$. Therefore, we have that 
  \begin{align*}
 \nabla F_{\rho}(\vec w)\cdot (\vec w-\wstar)&\geq  c'\sqrt{\eps}\|\wstar-\vec w\|_2+\rho \|\vec w\|_2(\|\vec w\|_2-\|\wstar\|_2\cos\theta)
 \\&\geq  c'\sqrt{\eps}\|\wstar-\vec w\|_2-\rho \|\vec w\|_2\|\wstar-\vec w\|_2
    \\&\geq  (c'/2)\sqrt{\eps}\|\wstar-\vec w\|_2
  \,,
  \end{align*}
  where in the third inequality we used that $\|\vec w\|_2\leq 2/R$ and that $\rho =O(F^3(\wstar))$.
\end{proof}

Next we handle the case where  $2/R \leq\|\vec w\|_2 \leq  c'\kappa/(\eps\Lambda_2)$  (see Case 2 of
\Cref{prop:gradient-bound-regular}).  Again, the fact that we choose a small regularization
parameter allows us to show that  the gradient field in this case behaves similarly to that of the
unregularized $L_2^2$ objective (see Case 2 of \Cref{prop:gradient-bound-ramp}).
\begin{claim}\label{clm:constant-contri-reg-2}
If $2/ R \leq\|\vec w\|_2 \leq  c'\kappa/(\eps\Lambda_2) $, then 
    \begin{align*}
   \nabla F_\rho(\vec w)\cdot(\vec w-\wstar)\geq (\sqrt{c'}/2)\sqrt{\eps} \|\wstar-\vec w\|_{\vec w}
   \,.
\end{align*}

\end{claim}
\begin{proof}
 We compute the contribution of the regularizer in the direction $\vec w-\wstar$. This is equal to
 $\rho \|\vec w\|_2(\|\vec w\|_2-\|\wstar\|_2\cos\theta)$. Note that this is positive when $\|\vec
 w\|_2-\|\wstar\|_2\cos\theta\geq 0 $ and negative otherwise. Hence, if $\theta\in(\pi/2,\pi)$ the
 contribution of the regularizer is positive, and therefore it is bounded from below by the
 contribution of the gradient without the regularizer. We need to choose the value of $\rho$ so that
 the regularizer cancels out the contribution of the noise when $\|\vec w\|_2$ is large (i.e., when
 the regularizer has positive contribution).

From \Cref{prop:gradient-bound-ramp}, we have that if
$\|\wstar-\vec w\|_{\vec w}\geq \sqrt{\eps/\kappa}$, then $\nabla F(\vec w)\cdot(\vec w-\wstar)\geq c'\sqrt{\eps}\|\vec w-\wstar\|_{\vec w}$, hence, it holds
\begin{align*}
   \nabla F_\rho(\vec w)\cdot(\vec w-\wstar) &\geq c'\sqrt{\eps}\|\vec w-\wstar\|_{\vec w} + \rho \|\vec w\|_2(\|\vec w\|_2-\|\wstar\|_2\cos\theta)
   \\&\geq  c'\sqrt{\eps}\|\vec w-\wstar\|_{\vec w} - \rho \|\vec w\|_2|\|\vec w\|_2-\|\wstar\|_2\cos\theta|
   \\&\geq \frac{\|\vec w\|_2-\|\wstar\|_2\cos\theta|}{\|\vec w\|_2^{3/2}}( c'\sqrt{\eps}-\rho \|\vec w\|_2^{2.5}) + c'\sqrt{\eps}\frac{\|\wstar\|_2\sin\theta}{\|\vec w\|_2}
   \\&\geq (c'/2)\sqrt{\eps}\|\vec w-\wstar\|_{\vec w}
   \,,
\end{align*}
where for the last inequality we used that $ \|\vec w\|_2 \leq c'\kappa/(\eps\Lambda_2)$ and that
$\rho\leq \frac{\Lambda_2^{2.5} \eps^3}{4 c'^{1.5} \kappa^{2.5} }$.

\end{proof}

Finally, we consider the case where $c'\kappa/(\eps\Lambda_2)\leq\|\vec w\|_2 \leq c'\kappa/\eps $
and $\|\proj_{\vec w^\perp} \wstar\|_2\geq 2K/R$.  We show that the contribution in the direction
$\vec w-\wstar$ of the unregularized gradient of $L_2^2$ is greater than the contribution of the 
gradiend corresponding to the regularizer.  The proof of the following claim can be found in
\Cref{app:bounded-plain-landscape}.
\begin{claim}\label{clm:constant-contri-reg-1}
If
$c'\kappa/(\eps\Lambda_2)\leq\|\vec w\|_2 \leq c'\kappa/\eps$ and $\| \proj_{\vec w^\perp} \wstar \|_2\geq 2K/R$ then 
    \begin{align*}
   \nabla F_\rho(\vec w)\cdot(\vec w-\wstar)\geq (c'/2) \sqrt{\eps} \|\wstar-\vec w\|_{\vec w}
   \,.
\end{align*}

\end{claim}

To prove the \Cref{prop:gradient-bound-regular}, it remains to show that $\|\vec w-\wstar\|_{\vec w}$ is small 
when $\| \proj_{\vec w^\perp} \wstar \|_2\leq 2K/R$ and $c'\kappa/(\eps\Lambda_2) \leq \|\vec
w\|_2 \leq  c'\kappa/\eps$. 
Similarly to the proof of \Cref{clm:constant-contri-reg-1}, we need to 
consider only the case where $\theta\in(0,\pi/2)$ and $\|\vec w\|_2\leq 2\|\wstar\|_2$.
 In fact, we show that in this case the vector $\vec w$ is
close to the target vector $\wstar$.
 We have:
\begin{align*}
    \|\vec w-\wstar\|_{\vec w}&\leq \frac{2\|\vec w\|_2+\|\wstar\|_2}{\|\vec w\|_2^{3/2}}+ \frac{\| \proj_{\vec w^\perp} \wstar \|_2}{\|\vec w\|_2^{1/2}}
    \\&\leq \frac{K/R +1}{\|\vec w\|_2^{1/2}} +\frac{\|\wstar\|_2}{\|\vec w\|_2^{3/2}}\;,
\end{align*}
where in the first inequality we used the triangle inequality and in the second the fact 
that $\| \proj_{\vec w^\perp} \wstar \|_2\leq 2K/R$. Moreover, recall that $c'\kappa/(\eps\Lambda_2)  \leq \|\vec w\|_2$,
$\|\wstar\|_2\leq U/\eps$ and $K= \Lambda_2^{2.5}U/(\sqrt{c'\kappa}R L)$. 
Therefore, we have that
 $\|\vec w-\wstar\|_{\vec w}\leq U\sqrt{\eps}\Lambda_2^{3}/(c'\kappa)^{3/2}\leq (c U/\kappa^5) \sqrt{\eps}
 $, for some absolute constant $c>0$.

\subsection{Proof of \Cref{thm:stationary-points-regular}}
Before proving \Cref{thm:stationary-points-regular}, we need to show that when the two vectors $\vec
w,\vec v$ have small distance with respect to an appropriate norm $\|\cdot\|$ (either the $\ell_2$
or the Weighted-Euclidean norm of \Cref{main-def:weighted-euclidean}), then the $L_2^2$
distance of the corresponding sigmoidals, i.e., 
$\Exx[(\sigma(\vec w\cdot \x)-\sigma(\vec v\cdot \x))^2] $, is small.
The proof of the following result can be found in \Cref{app:bounded-plain-landscape}.
\begin{lemma}[Parameter vs $L_2^2$ Distance]\label{lem:bound-difference-bounded}
Let $\D_\x$ be an $(L,R)$ well-behaved distribution.  Let $\sigma$ be a $(\tau,\xi,\mu)$-sigmoidal
activation.  For any vectors $\vec w,\vec v\in \R^d$, we have $
\Exx[(\sigma(\vec w\cdot
\x)-\sigma(\vec v\cdot \x))^2] \leq \xi^2 \|\vec w-\vec v\|_2^2$.
Moreover, if $\theta=\theta(\vec w,\vec v) < \pi/4$, 
$\|\vec w\|_2\leq \delta\|\vec v\|_2$ and $\delta\geq 1$, and $\|\vec w\|_2 > 2/R$, 
 it holds
\[
 \Exx[(\sigma(\vec w\cdot \x)-\sigma(\vec v\cdot \x))^2] \lesssim \frac{\boundd^2\delta^3}{L^4\boundc^3} \|\vec w-\vec v\|_{\vec w}^2  \,.
 \]
\end{lemma}
\noindent We now can now prove \Cref{thm:stationary-points-regular}.
\begin{proof}[Proof of \Cref{thm:stationary-points-regular}]
We note that from \Cref{lem:radius-approx}, there exists a $\wstar\in \R^d$, such that 
$F(\wstar)\|\wstar\|_2=1+O(\boundd/(\boundc L))=U$ and $F(\wstar)\lesssim \eps$.
 We can assume that $\eps\leq \poly(\xi/(L R\mu))$, since otherwise any vector $\vec w$, gets error $F(\vec w)\leq 2\eps$. 
First we consider the case where $\|\vec w\|_2\leq 2/R$. From \Cref{prop:gradient-bound-regular} for $c'>0$ a sufficiently small absolute constant, we have that if $\|\wstar-\vec w\|_2\geq \sqrt{\eps}/(c'\kappa^5)$, then
\begin{align*}
\|\nabla &F(\vec w)\|_2\geq\nabla F(\vec w)\cdot\frac{\vec w-\wstar}{\|\vec w-\wstar\|_2}\geq c'\sqrt{\eps}\;.
\end{align*}
In order to reach a contradiction, assume that $\|\nabla F(\vec w)\|_2< c'\sqrt{\eps}$ and $\E_{\x\sim \D_\x}[(\sigma(\vec w\cdot \x)-\sigma(\wstar\cdot\x))^2]> \xi^2\eps/(c'^2\kappa^{10})$. From \Cref{prop:gradient-bound-regular}, we have that 
$\|\wstar-\vec w\|_2\leq \sqrt{\eps}/(c'\kappa^5)$.
Therefore, from \Cref{lem:bound-difference-bounded} we have 
that \[\xi^2\eps/(c'^2\kappa^{10})< \E_{\x\sim \D_\x}[(\sigma(\vec w\cdot \x)-\sigma(\wstar\cdot\x))^2]\leq \|\vec w-\wstar\|_2^2\leq \xi^2\eps/(c'^2\kappa^{10})\;,\]
which is a contradiction.

Next, we consider the case where $\|\vec w\|_2\geq c'\kappa/(2\eps)$. 
From \Cref{prop:gradient-bound-regular}, we know that there is no approximate stationary point in this region, i.e., 
there is no point $\vec w$ with $\|\nabla F(\vec w)\|_{*,\vec w}\leq c'\sqrt{\eps}$.

For the last case we consider the case where $2/R\leq \|\vec w\|_2\leq c'\kappa/\eps$. 
From \Cref{prop:gradient-bound-regular}, we have that if either $\|\vec w-\wstar\|_{\vec w}\geq U\sqrt{\eps}/(c'\kappa^5)$ or $\|\vec w\|_2\geq 2\|\wstar\|_2$, then
\[
\|\nabla F(\vec w)\|_{*,\vec w}\geq\nabla F(\vec w)\cdot\frac{\vec w-\wstar}{\|\vec w-\wstar\|_{\vec w}}\geq c'\sqrt{\eps}\,, \]
where we used that for any two vectors $\vec z_1,\vec z_2\in \R^d$, it holds 
$\vec z_1\cdot \vec z_2\leq \|\vec z_1\|_{*,\vec w}\|\vec z_2\|_{\vec w}$.

In order to reach a contradiction, assume 
that $\E_{\x\sim \D_\x}[(\sigma(\vec w\cdot \x)-\sigma(\wstar\cdot\x))^2]\geq C\frac{\boundd^2U^2}{L^4 \boundc^3\kappa^{10}}\eps$ 
and that  $\|\nabla F(\vec w)\|_{*,\vec w}\leq  c' \sqrt{\eps}$, where $C>0$ is a sufficiently large absolute constant. 
It holds that  $\|\wstar-\vec w\|_{\vec w}\leq U\sqrt{\eps}/(c'\kappa^5)$  and $\|\vec w\|_2\leq 2\|\wstar\|_2$. 
Hence, we have that
\begin{equation*}
   \|\wstar-\vec w\|_{\vec w}=\frac{|\|\vec w\|_2-\cos\theta\|\wstar\|_2|}{\|\vec w\|_2^{3/2}}
     + 
     \frac{\|\wstar\|_2\sin\theta }{\sqrt{\|\vec w\|_2}}\lesssim \frac{U\sqrt{\eps}}{\kappa^5}\;.
\end{equation*}
Therefore, it holds that 
\[ \sin\theta \lesssim \frac{\sqrt{\eps\|\vec w\|_2}}{\|\wstar\|_2\kappa^5}\lesssim\frac{\sqrt{\eps}}{\sqrt{\|\wstar\|_2}\kappa^5}\;,\]
where we used that $\|\vec w\|_2\leq 2\|\wstar\|_2$, and using that 
$\sqrt{\eps/\|\wstar\|_2}$ is smaller than a sufficiently small absolute constant, we get that $\sin\theta\leq 1/2$, and hence, $\theta\in(0,\pi/4)$.
Using again \Cref{lem:bound-difference-bounded}, we have that
\begin{align*}
\E_{\x\sim \D_\x}[(\sigma(\vec w\cdot \x)-\sigma(\wstar\cdot\x))^2]&\lesssim \frac{\boundd^2}{L^4\boundc^3}\|\vec w-\wstar\|_{\vec w}^2 \lesssim \frac{\boundd^2U^2}{L^4 \boundc^3\kappa^{10}}\eps\;.
     \end{align*}
Therefore, we get again a contradiction. The proof then follows by noting 
     that $F(\vec w)\leq 2 \eps+ 2\E_{\x\sim \D_\x}[(\sigma(\vec w\cdot \x)-\sigma(\wstar\cdot\x))^2]$.

\end{proof}

\section{The Landscape of the $L_2^2$-Loss for Unbounded Activations}\label{sec:unbounded-activations}

For unbounded activations we essentially characterize 
the optimization landscape of the $L_2^2$-loss as a function of the weight vector $\vec w$.
Specifically, our main structural theorem in this section establishes (roughly) the following: 
even though the population $L_2^2$-loss 
$F(\vec w)=\E_{(\x,y)\sim \D}[(\sigma(\vec w\cdot \x)-y)^2]$ is not convex, 
any approximate stationary point $\vec w$ of $F(\vec w)$ will have error 
close to that of the optimal weight vector $\vec w^\ast$, i.e.,  
$F(\vec w) \leq O(\eps)$.
In more detail, while there exist ``bad'' stationary points in this case,
they all lie in a cone around $-\vec w^\ast$, i.e., in the opposite direction of
the optimal weight vector.  As we will see in \Cref{sec:optimization},
gradient descent initialized at the origin will always avoid such stationary points.
Our main structural result for unbounded activations is as follows:

\begin{theorem}[Stationary Points of $(\alpha,\lambda)$-Unbounded Activations] \label{prop:stationary-points-unbounded}
Let $\D$ be an $(\eps,W)$-corrupted, $(L,R)$-well-behaved distribution in $\R^d$. 
Let $\sigma$ be an $(\alpha,\lambda)$-unbounded activation and let 
$F(\vec w)=\E_{(\x,y)\sim \D}[(\sigma(\vec w\cdot \x)-y)^2]$.
Then,
if for some $\vec w\in \R^d$, with $\vec w \cdot \wstar \geq 0$ and $\|\vec w\|_2\leq W$ it holds
$\|\nabla F(\vec w)\|_2\leq 2\lambda \sqrt{\eps}$, then 
$F(\vec w)\leq C\big(\frac{\lambda}{\alpha}\big)^4\frac{1}{L^2 R^8}\eps$, 
for some absolute constant $C>0$.
\end{theorem}

\begin{remark}\emph{
An important feature of the above theorem statement is that the error guarantee
of the stationary points of the $L_2^2$-loss $F(\vec w)$ does
not depend on the size of the weight vector $\vec w^\ast$.
It only depends on the ratio of the constants $\lambda$ and
$\alpha$ (that for all activation functions discussed above 
is some small absolute constant).  For the special case
of ReLU activations, it holds that $\lambda/\alpha = 1$.}
\end{remark}

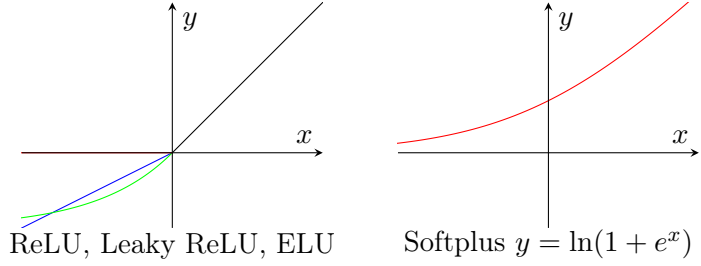
\begin{figure}[h]
\ifplots
\pgfplotsset{every axis title/.append style={at={(0.5,-0.25)}}} 
\usetikzlibrary{matrix}
\usepgfplotslibrary{groupplots}
\centering
\begin{tikzpicture}
 \begin{groupplot}[group style={group size=4 by 1},height=2cm,width=2cm,
 restrict y to domain=-4:4,xlabel=$x$,ylabel=$y$, legend pos=north west,axis y line =middle,
 axis x line =middle,
 				axis on top=true,
 				xmin=-2,
 				xmax=2,
 				ymin=-1,
 				ymax=2,
 				   y=1cm,
                  x=1cm,
                   yticklabels={,,},
 				    xticklabels={,,},
 				      ytick=\empty,
 				        xtick=\empty,]
				\nextgroupplot[title= {ReLU, Leaky ReLU, ELU}]
\addplot[color=black, samples=100, domain=0:4]
				{x};
				\addplot[color=blue, samples=100, domain=-4:0]
				{0.5*x};
				\addplot[color=green, samples=100, domain=-4:0]
				{2(exp(x)-1)};
				\addplot[color=red, samples=100, domain=-4:0]
				{0.0};
				\nextgroupplot[title=Softplus ${y=\ln(1+e^{x})}$]
				\addplot[color=red, samples=100, domain=-4:4]
				{ln(1+exp(x))};

			\end{groupplot}
\end{tikzpicture}
\fi
    \caption{Some well-known unbounded activation functions that satisfy \Cref{def:well-behaved-unbounded-intro} .}
    \label{fig:unbounded-activation-functions}
\end{figure}

\begin{figure}
    \centering
    \begin{tikzpicture}[scale=1.35]
      \coordinate (start) at (0.5,0);
      \coordinate (center) at (0,0);
      \coordinate (end) at (0.5,0.5);
      \draw (-2,1) node[left] {};
      \draw (-2,0.5) node[left] {};
      \draw (2,2) node[above] {};
      \draw (2,1.2) node[above] {};
      \draw[black,dashed, thick,red](-2,1) -- (2,1);
      \draw[black,dashed, thick, red](-2,0.5) -- (2,0.5);
      \draw[fill=blue, opacity=0.5,draw=none] (-1,0.5) -- (-1 ,1)--(-1.5,1)--(-1.5,0.5);
\draw[black,dashed, thick](-1,0) -- (-1,2);
      \draw[black,dashed, thick](-1.5,0) -- (-1.5,2);
      \draw[->] (-2,0) -- (2,0) node[anchor=north west,black] {};
      \draw[->] (0,-1) -- (0,2) node[anchor=south east] {};
      \draw[thick,->] (0,0) -- (-0.7,0.7) node[anchor= south east,below,right] {$\wstar$};
      \draw[black] (-1,-1) -- (2,2);
\draw[thick ,->] (0,0) -- (0,1) node[right=2mm,below] {$\bw$};
\draw (2,0.5) node[right] {$R/2$};
      \draw (2,1) node[right] {$R$};
     \draw (-1,2) node[above] {$-R/2$};
      \draw (-1.5,2) node[above] {$-R$};
     
    \end{tikzpicture}
    \caption{Using our distributional assumptions, we identify a region (``blue'') 
    that provides enough contribution to the gradient, so that an update step 
    will decrease the distance with the optimal one.}
    \label{fig:contribution-unbounded}
\end{figure}
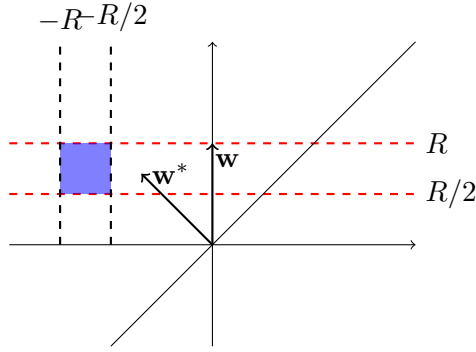

\paragraph{Proof of \Cref{prop:stationary-points-unbounded}}

\begin{remark}
  \label{rem:sigma-zero}
{\em We remark that we assume $\sigma(0) = 0$ to simplify the presentation.  If $\sigma(0) \neq 0$, then we can
always consider the loss function $\tilde \sigma(t) = \sigma(t) - \sigma(0)$ and also subtract $\sigma(0)$
from the labels $y$, i.e., $y' = y - \sigma(0)$.  Our results directly apply to the transformed instance.}
\end{remark}

For the proof of \Cref{prop:stationary-points-unbounded}, 
we need the following structural result for the gradient field of the $L_2^2$ loss for unbounded
activations.  We show that as long as $\| \vec w - \vec w^\ast\|_2$ is larger than roughly
$\sqrt{\eps}$ and $\vec w \cdot \vec w^\ast \geq 0$, then the gradient at $\vec w$ has large 
projection in the direction of $\vec w - \vec w^\ast$, i.e., it ``points in the right direction''.

\begin{proposition}\label{main-pro:gradient-bound}
  Let $\D$ be an $(\eps,W)$-corrupted, $(L,R)$-well-behaved distribution and $\sigma$ be
  an $(\alpha, \lambda)$-unbounded activation.
  For any $\vec w\in\R^d$ with $\vec w \cdot \vec w^\ast \geq 0$ and 
  $\|\vec w - \vec w^\ast\|_2 \geq C \lambda/ (\alpha^2 L R^4) \sqrt{\eps}$, it holds
 \( \nabla F(\vec w)\cdot(\vec w-\wstar)\gtrsim  ~ \alpha^2  L R^4 \|\vec w-\wstar\|_2^2
 \).
\end{proposition}

We remark that one can readily use the above proposition to obtain an efficient algorithm via
black-box optimization methods.  To achieve this, we can simply start by finding a stationary point
$\vec w^{(0)}$, and then repeat our search constraining on the set 
$\{\vec w: \vec w \cdot \vec w^{(0)} \leq 0\}$ to obtain some stationary point $\vec w^{(1)}$. 
Then we continue in the set $\{\vec w^{(0)} \cdot \vec w \leq 0, \vec w^{(1)} \cdot \vec w \leq 0\}$.  
Since for all the boundaries we have (using \Cref{main-pro:gradient-bound})
that the gradient has non-zero projection onto the direction $\vec w - \vec w^\ast$,
we obtain that by adding these linear constraints, we do not
introduce new stationary points. It is not hard to see that after $O(d)$ such iterations, we obtain
a stationary point that lies in the halfspace $\vec w^\ast \cdot \vec w \geq 0$, and therefore
is an approximately optimal solution.

\begin{proof}[Proof of \Cref{main-pro:gradient-bound}]
We have
\begin{align*}
\nabla F(\vec w)&\cdot(\vec w-\wstar)\\&=\underbrace{\E[(\sigma(\wstar\cdot \x)-y)\sigma'(\vec w\cdot \x)(\vec w\cdot \x-\wstar \cdot \x)]}_{I_1}\\
& +\underbrace{\E[(\sigma(\vec w\cdot \x)-\sigma(\wstar\cdot \x))\sigma'(\vec w\cdot \x)(\vec w\cdot \x-\wstar \cdot \x)]}_{I_2}\;.
\end{align*}
We start by bounding the contribution of $I_1$. 
We show that the ``noisy'' integral $I_1$ has small negative contribution 
that is bounded by some multiple of $\|\vec w-\wstar\|_2\sqrt{\eps}$. 
We show the following claim.
\begin{claim} \label{clm:I_1-unbounded}
It holds that  $I_1 \geq -2\lambda \|\vec w - \wstar\|_2\sqrt{\eps}$.
\end{claim}
\begin{proof}
Using the \CS\ inequality, we have that
\begin{align*}
   I_1&\geq -\E[|(\sigma(\wstar\cdot \x)-y)\sigma'(\vec w\cdot \x)(\vec w\cdot \x-\wstar \cdot \x)|] \\ 
   &\geq-\lambda   \E[|\sigma(\wstar\cdot \x)-y||\vec w\cdot \x-\wstar \cdot \x|]\\
    &\geq -\lambda   (\E[(\sigma(\wstar\cdot \x)-y)^2])^{ 1/2}(\E[(\vec w\cdot \x-\wstar \cdot \x)^2])^{1/2}\\
    &\geq -\lambda ~ \sqrt{\eps} ~ \|\vec w-\wstar\|_2 
    \left(\max_{\vec v\in \R^d,\|\vec v\|_2=1}\E[(\vec v\cdot \x)^2]\right)^{1/2}
    \geq -2\lambda \|\vec w-\wstar\|_2\sqrt{\eps}\;,
\end{align*}
where in the first inequality we used the fact that $\sigma$ is $\lambda$-Lipschitz, 
in the second that $F(\wstar)=\eps$, 
and in the last the fact that the distribution is isotropic. 
Hence, we have that
\begin{equation}\label{eq:bad-region}
    I_1\geq -2\lambda \|\vec w-\wstar\|_2\sqrt{\eps}\;.
\end{equation}
This completes the proof of \Cref{clm:I_1-unbounded}.
\end{proof}
Next we bound from below the contribution of the ``clean'' examples, i.e., $I_2$. 
We show that this positive contribution is bounded below by a multiple of $\|\vec w-\wstar\|_2^2$, 
which is enough to surpass the contribution of the negative region $I_2$.
We have the following claim.
\begin{claim}
It holds that  $I_2\geq \frac{7\alpha^2 L R^4}{64} \|\vec w-\wstar\|_2^2 $.
\end{claim}
\begin{proof}
We first notice that $(\sigma(\vec w\cdot \x)-\sigma(\wstar\cdot \x))\sigma'(\vec w\cdot \x)(\vec w\cdot \x-\wstar \cdot \x)\geq 0$ 
for every $\x\in \R^d$, because $\sigma$ is an increasing function. 
Without loss of generality, we may assume that $\vec w/\|\vec w\|_2 = \vec e_2$, 
and therefore $\wstar =\|\wstar\|_2 (\cos \theta \vec e_2 -\sin\theta \vec e_1)$.
For simplicity, we abuse notation and denote $\D_\x$ the marginal distribution 
on the subspace spanned by the vectors $\vec w,\wstar$. 
From the definition $(\alpha,\lambda)$-unbounded activations, we have that 
$\inf_{t\in(0,R/\|\vec w\|_2)}\sigma'(t)\geq \inf_{t\in(0,\infty)}\sigma'(t) \geq \alpha>0$. 
We now have that
\begin{align*}
   I_2&=\E[(\sigma(\vec w\cdot \x)-\sigma(\wstar\cdot \x))\sigma'(\vec w\cdot \x)(\vec w\cdot \x-\wstar \cdot \x)]\\
   &\geq \alpha^2\E[ (\vec w\cdot \x-\wstar \cdot \x)^2\1\{\x_1\in(-R,-R/2),\x_2\in(R/2,R)\}]\;, 
\end{align*}
where the region inside the indicator is the ``blue region'' of \Cref{fig:contribution-unbounded}. 
We remark that this region is not particularly ``special'': 
we can use other regions that contain enough mass depending on the distributional assumptions. 
At a high level, we only need a rectangle that contains enough mass 
and the gradient in these points is non-zero. 
Therefore, using the fact that the distribution is $(L,R)$-well-behaved, we have
\begin{align*}
\E\big[ (\vec w\cdot \x- &\wstar \cdot \x)^2 ~ \1\{\x_1\in(-R,-R/2),\x_2\in(R/2,R)\big] \\
&=\E\left[ (\|\vec q^{\|_{\vec w}}\|_2^2 \x_2^2+ \|\vec q^{\perp_{\vec w}}\|_2^2 \x_1^2) ~ \1\{\x_1\in(-R,-R/2),\x_2\in(R/2,R)\}\right]
\\&\geq \frac{7LR^4}{32}(\|\vec q^{\|_{\vec w}}\|_2^2+ \|\vec q^{\perp_{\vec w}}\|_2^2) = \frac{7LR^4}{64}\|\vec  w-\wstar\|_2^2\;,
\end{align*}
where $\vec q=\vec w-\wstar$.
Therefore, we proved that
\begin{equation}\label{eq:good-region}
    I_2\geq \frac{7\alpha^2 L R^4}{64} \|\vec w-\wstar\|_2^2\;.
\end{equation}
\end{proof}
\noindent Thus, combining \Cref{eq:bad-region,eq:good-region}, we have that
\[
 \nabla F(\vec w)\cdot(\vec w-\wstar)\geq \|\vec w-\wstar\|_2\left(\alpha^2 \frac{ 7L R^4}{64} \|\vec w-\wstar\|_2- 2\lambda \sqrt{\eps}\right)\;,
\]
which completes the proof of \Cref{main-pro:gradient-bound}.
\end{proof}

To prove \Cref{prop:stationary-points-unbounded}, in order to reach a contradiction we assume  
that $F(\vec w)\geq C\frac{\lambda^4}{\alpha^4 L^2 R^8}\eps$ for absolute constant $C>0$ 
and that $\|\nabla F(\vec w)\|_2\leq \frac{\alpha^2 L R^4}{100}\sqrt{\eps}$. 
From \Cref{main-pro:gradient-bound}, we have that 
\[
\frac{7\alpha^2 L R^4}{64}\sqrt{\eps}\geq \nabla F(\vec w)\cdot\frac{(\vec w-\wstar)}{\|\vec w-\wstar\|_2}\geq \frac{7\alpha^2 L R^4}{64} \left(\|\vec w-\wstar\|_2- \sqrt{\eps}\frac{\lambda 128}{7\alpha^2 L R^4}\right)\;,
\]
and therefore
\begin{equation}\label{eq:bound-a}
    \|\vec w-\wstar\|_2\leq \frac{128\lambda\sqrt{\eps}}{7\alpha^2 L R^4}\;.
\end{equation}
Moreover, we have
\begin{align*}
    F(\vec w)
    &=
    \E[(\sigma(\vec w\cdot \x)-y)^2]\leq 2\E[(\sigma(\wstar\cdot \x)-y)^2]+2\E[(\sigma(\vec w\cdot \x)- \sigma(\wstar\cdot \x))^2] \\
    &\leq 2\eps +2\lambda^2\|\vec w-\wstar\|_2^2\;.
\end{align*}
Therefore, using \Cref{eq:bound-a}, we get
\[
F(\vec w)\leq C\bigg(\frac{\lambda}{\alpha}\bigg)^4\frac{1}{ L^2 R^8}\eps\;,
\]
which leads to a contradiction. This completes the proof of \Cref{prop:stationary-points-unbounded}.\qed

\section{Optimizing the Empirical $L_2^2$-Loss}
\label{sec:optimization}
In this section, we show that given sample access to a distribution over labeled examples $\D$ on
$\R^d \times \R$, we can optimize the $L_2^2$ loss via (approximate) gradient descent. 
This allows us to efficiently find a weight vector that achieves error $O(\eps)$. 
Ideally, we would like to perform gradient descent with the population gradients, 
i.e., $\vec w^{(t+1)} \gets \vec w^{(t)} - \eta (\nabla F(\vec w^{(t)})) $.  
We do not have direct access to these gradients, 
but we show that it is not hard to estimate them using
samples from $\D$; see \Cref{alg:gd-2}.
We state our algorithmic result for sigmoidal activations in \Cref{ssec:sigmoidal-algorithm}
and for unbounded activations in \Cref{ssec:unbounded-algorithm}.

\subsection{Sigmoidal Activations} \label{ssec:sigmoidal-algorithm}

In this section, we prove our main algorithmic result for learning the sigmoidal 
activations discussed in \Cref{main-sec:bounded-plain-landscape}.  

\begin{theorem}[Learning Sigmoidal Activations] \label{thm:bounded-main-theorem}
    Let $\D$ be an $\eps$-corrupted, $(L,R)$-well-behaved distribution on 
    $\R^d \times \R$ and $\sigma(\cdot)$ be a $(\tau, \boundc, \boundd)$-sigmoidal activation.
    Set $\kappa = L^6R^6\mu^3\tau^4/\xi^2$ and let $c>0$ be a sufficiently small absolute constant.
    Then gradient descent (\Cref{alg:gd-2}) with step size $\eta = c\eps^{2.5}$, 
    regularization $\rho = (1/c)\eps^3/k^5$, truncation threshold $M=\xi/\mu$, $N = \wt{\Theta}( d/\eps  \log(1/\delta) ) ~ \poly(1/\kappa)$ samples, 
    and $T = \poly(1/(\eps \kappa)) $ iterations converges to a vector $\vec w^{(T)} \in \R^d$ that, 
    with probability $1-\delta$, satisfies \( F(\vec  w^{(T)} ) \leq \poly(1/\kappa) ~ \eps\,.  \) 
    \end{theorem}

\begin{Ualgorithm}
	\centering
	\fbox{\parbox{6in}{
			{\bf Input:}   Iterations: $T$, $N$ samples $(\x^{(i)}, y^{(i)})$ from $\D$, step size: $\eta$, bound $M$, regularization $\rho$. \\
			{\bf Output:}  A weight vector $\vec w^{(T)}$.
			\begin{enumerate}
				\item Let $\widehat{F}(\vec w)= \frac{1}{N}\sum_{i=1}^N(\sigma(\vec w\cdot \x^{(i)}) -\tilde{y}^{(i)})^2$, where $\wh{y}^{(i)} = \sgn(y^{(i)}) \min(|y^{(i)}|, M)$ 
				\item $\vec w^{(0)} \gets \vec 0$.
				\item For $t=0,\ldots, T$ do
				\begin{enumerate}
					\item   $\vec g\tth\gets \nabla \widehat F(\vec w\tth)$.
					\item $\vec w^{(t+1)} \gets \vec w\tth-\eta (\vec{g}\tth + \rho \vec w\tth)$.
				\end{enumerate}
				
			\end{enumerate}
			
	}}
	\caption{Gradient Descent Algorithm for Optimizing the $L_2^2$-Loss for Sigmoidal Activations}
	\label{alg:gd-2}
\end{Ualgorithm}

\subsubsection*{Proof of \Cref{thm:bounded-main-theorem}}
We will first assume that we have access to the population gradients of the $L_2^2$ objective and
then show that given samples from $\D$, gradient descent again converges to some approximately
optimal solution.  Before we start, we observe that, without loss of generality, we may assume that 
$\sigma(0) = 0$ (otherwise we can subtract $\sigma(0)$ from $y$ and the activation, see \Cref{rem:sigma-zero}).
Moreover, from \Cref{clm:upper-bound-value-bounded} we know that $|\sigma(t)| \leq \xi/\mu$.  Therefore, for every
example $(\x, y) \sim D$ we can ``truncate'' its label to $\wh{y} = \sgn(y) \min(|y|, \xi/\mu)$ and
the new instance will still be (at most) $\eps$-corrupted.  To simplify notation, from now on, we
will overload the notation and use $y$ instead of $\wh{y}$ (assuming that for every example 
$(\x,y)$ it holds that $|\sigma(\vec w \cdot \x) - y| \leq 2 \xi/\mu$).

\paragraph{Population Gradient Descent}
Recall that in \Cref{prop:gradient-bound-regular} we showed that the population gradient field
``points'' to the right direction, i.e., a step of gradient descent would decrease the distance
between the current $\vec w$ and a target $\vec v$.  Notice that in
\Cref{prop:gradient-bound-regular} we require that the target vector $\vec v$ satisfies $F(\vec v) =
O(\eps)$ and $\|\vec v\|_2 \leq O(1/\eps)$.  Indeed, from \Cref{lem:radius-approx}, we have that
there exists an approximately optimal target vector $\vec v\in \R^d$, such that $ \|\vec v\|_2\leq
U/\eps$ for some $U = O(\xi/(\mu L))$ and $F(\vec v) = O(\eps)$.

Moreover, recall the way we measure distances changes depending on how far $\vec w$ is from the origin (see
the three cases of \Cref{prop:gradient-bound-regular}).  We first show that, given any gradient
field that satisfies the properties given in \Cref{prop:gradient-bound-regular}, then gradient
descent with an appropriate step size will converge to the target vector $\vec v$.  The distance of
$\vec v$ and the guess $\vec w^{(T)}$ after $T$ gradient iterations will be measured with respect to
$\|\cdot\|_2$ when $\vec w^{(T)}$ is close to the origin and with respect to $\|\cdot \|_{\vec
w^{(T)}}$ (see \Cref{main-def:weighted-euclidean}) otherwise.  The proof of
\Cref{lem:gradient-field-bounded} can be found in \Cref{app:optimization}.
\begin{lemma}[Gradient Field Distance Reduction]
    \label{lem:gradient-field-bounded}
    Let $Z_1 > Z_0 \geq 1$. Let $\vec g: \R^d \mapsto \R^d$ be a vector field 
    with $\|\vec g(\vec w)\|_2 \leq B$ for every $\vec w \in \R^d$ with $\|\vec w\|_2\leq 2Z_1$.
    We assume that  $\vec g$ satisfies the following properties with respect to some unknown 
    target vector $\vec v$:
    \begin{enumerate}
    \item  
    If $\|\vec w\|_2 \leq Z_0$
    and $\|\vec w - \vec v\|_2 \geq \alpha_1$
    then 
    it holds that $\vec g(\vec w) \cdot ( \vec w - \vec v) \geq \alpha_2 \|\vec w - \vec v\|_2$,
    for $\alpha_1, \alpha_2 > 0$.
    \item 
    If $ Z_0 < \|\vec w\|_2 \leq Z_1$ 
    and $\|\vec w - \vec v\|_{\vec w} \geq \beta_1$
    it holds that $\vec g(\vec w) \cdot ( \vec w - \vec v) \geq \beta_2 \|\vec w - \vec v\|_{\vec w}$, 
    for $\beta_1, \beta_2 > 0$.
    \item 
    For some $\zeta \in (0,1), \gamma>0$, we have that if 
    $\|\vec w\|_2 \geq \zeta Z_1 $,
    it holds that $\vec g(\vec w) \cdot  \vec w \geq \gamma \|\vec w\|_{\vec w}$.
    \end{enumerate}
    We consider the update rule $\vec w^{(t+1)} \gets \vec w^{(t)} - \eta \vec g(\vec w^{(t)})$
    initialized  with $\vec w^{(0)} = \vec 0$
    and step size  
    \[
    \eta = \frac{1}{B^2} \min\left(\alpha_1 \alpha_2, \frac{\beta_1 \beta_2}{Z_1^{3/2}}, \frac{2 \gamma}{Z_1}, (1-\zeta) Z_1 B\right) \;.
    \]
    Let $T$ be any integer larger than $\Big\lceil\frac{\|\vec v\|^2}{\eta  \min(\alpha_1 \alpha_2, \beta_1 \beta_2/Z_1^{3/2})} \Big\rceil $.
    We have that $\|\vec w^{(T)}\|_2 \leq Z_1$.
    Moreover, if  $\|\vec w^{(T)} \|_2 \leq Z_0$ it holds that 
    $\|\vec w^{(T)} - \vec v \|_{2} \leq \eta B + \max(\alpha_1, (2 Z_0)^{3/2} \beta_1)$
    and if $\|\vec w^{(T)}\|_{2} > Z_0$ we have 
    $\|\vec w^{(T)} - \vec v \|_{\vec w^{(T)}} 
    \leq \sqrt{2} \eta B +  \max(\sqrt{2} \alpha_1, e^{3 Z_1^{3/2} \eta B} \beta_1) \,.
    $
\end{lemma}

We now show that the population gradient field, $\nabla F(\vec w)$ satisfies the assumptions 
of \Cref{lem:gradient-field-unbounded} for any $\|\vec w\|_2\leq \kappa/\eps$.  
We first show that $\|\nabla F(\vec w)\|_2$ is bounded.  We have 
\begin{align*}
    \|\nabla F(\vec w )\|_2&= \left\|\E[(\sigma(\vec w \cdot \x)-y)\sigma'(\vec w \cdot \x)\x] +\rho \|\vec w\|_2\vec w\right\|_2 
    \\
    &\leq \max_{\|\vec u\|_2=1}\E[(\sigma(\vec w \cdot \x)-y)\sigma'(\vec w \cdot \x)\vec u\cdot \x] +\rho\|\vec w\|_2^2
    \\ 
    &\leq  (\xi^2/\mu) \max_{\|\vec u\|_2=1}\E[|\vec u\cdot \x|] +\rho\|\vec w\|_2^2
    \leq \xi^2/\mu +\rho\|\vec w\|_2^2\lesssim 1/\kappa \;,
\end{align*}
where we used the fact that $| \sigma(\vec w \cdot \x) - y| \leq \xi/\mu$, 
$\sigma'(t) \leq \xi$, that the $\x$-marginal of $\D$ is isotropic and that $\rho =O(\eps^3)$.

From \Cref{prop:gradient-bound-regular} we have that for any vector  $\vec w\in\R^d$, with $\|\vec
w\|_2\leq 2/R$, it holds that if $\|\vec w - \vec w^\ast\|_2 \geq  \sqrt{ \eps }   /(c'\kappa)$ for
some absolute constant $c'>0$ it holds \( \nabla F_\rho(\vec w)\cdot(\vec w-\wstar)\geq  \sqrt{\eps}
\|\vec w-\wstar\|_2 \,. \) Furthermore, for any vector  $\vec w\in\R^d$, with $2/R\leq\|\vec
w\|_2\leq c'\kappa/\eps$, if $\|\vec w - \vec w^\ast\|_{\vec w} \geq  c' U \sqrt{ \eps } /\kappa$,
then \( \nabla F_\rho(\vec w)\cdot(\vec w-\wstar)\geq  c'\sqrt{\eps} \|\vec w-\wstar\|_{\vec w} \,.
\) Finally, for any vector $\vec w\in\R^d$, with $c''\kappa/\eps\leq\|\vec w\|_2$, it holds that \(
\nabla F_\rho(\vec w)\cdot\vec w\geq  c'\sqrt{\eps} \|\vec w\|_{\vec w} \), for some $0<c''\leq c'$.
Therefore, the true gradient field of $F$ satisfies the assumptions of
\Cref{lem:gradient-field-unbounded} with $B= O(1/\kappa)$, $Z_0 = O(1/R)$, $Z_1 =
\poly(LR\tau\mu/\xi)$, $\zeta = 1/2$, $\alpha_1, \beta_1  =  \poly(\xi/(\mu \tau L R) ) \sqrt{
\eps}$, and $\alpha_2, \beta_2,\gamma =  O(\sqrt{\eps})$.

\paragraph{Using Sample-Estimated Gradients}
In the following claim we show that since $\nabla F(\vec w)$ is a sub-exponential random variable, 
with roughly  $N = \wt{O}(d/\eps'^2)$ samples we can get $\eps$-estimates of the population
gradients.  The proof of \Cref{clm:estimation-gradient-bounded} can be found in
\Cref{app:optimization}.
\begin{claim}[Sample Complexity of Gradient Estimation]\label{clm:estimation-gradient-bounded}
    Fix $B,\eps' > 0$ with $\eps' \leq 1/\sqrt{B}$.  Using $N =  \wt{O}( (d/\eps'^2)
    \poly(\xi/(L \mu)) \log(B/\delta)) $ samples from $\D$, we define the empirical $L_2^2$-loss
    as $\widehat{F}(\vec w) = \sum_{i=1}^N ( \sigma(\vec w \cdot \x^{(i)})  - y^{(i)})^2$ and its 
    corresponding empirical gradient field \(\nabla \widehat{F}(\vec w) \).
    With probability at least $1-\delta$, for all $\vec w$ with $\|\vec w\| \leq B$, it holds $\| \vec
   \nabla \widehat{F}(\vec w) - \nabla F(\vec w) \|_2 \leq \eps'$ and  
   $\| \nabla \widehat F (\vec w) - \nabla F(\vec w) \|_{*, \vec w} \leq \eps'$.
\end{claim}
Using \Cref{clm:estimation-gradient-bounded}, we can estimate all the gradients with probability
$1-\delta'$ and accuracy $\eps'$, with  $N =  \wt{O}( (d/\eps'^2) \poly(\xi/(L \mu))
\log(B/\delta')) $ samples, for some parameters $\eps',\delta>0$ that we will choose below.  We now
show that the empirical gradients will also satisfy the required gradient field assumptions of
\Cref{lem:gradient-field-unbounded}.  Assume that we have  $\|\nabla \widehat{F}(\vec w) -  \nabla F (\vec w)
\|_2 \leq \eps'$ and $\|\nabla \widehat{F}(\vec w) - \nabla F( \vec w) \|_{*,\vec w} \leq \eps'$ for some
$\eps'$ to be specified later.  Using the triangle inequality we have that 
\begin{equation*}
\|\nabla \widehat{F}(\vec w) \|_2  \leq \|\vec \nabla F (\vec w) \|_2 + \eps'\;.
\end{equation*} 
Therefore, for $\eps' \leq 1/\kappa$ we obtain that $\|\nabla \widehat{F}(\vec w) \|_2  = O(1/\kappa) \leq O(B)
$.  We next show that the empirical gradient also points to the direction of $\vec w - \vec w^\ast$,
i.e., satisfies the assumptions of \Cref{lem:gradient-field-bounded}.  For the case where $\|\vec w
\| \leq Z_0$ we have that 
\begin{equation}
\label{eq:lower-bound-gradient-1}
    \nabla \widehat{F}(\vec w) \cdot(\vec w - \vec v)
    \geq  \nabla F(\vec w) \cdot (\vec w - \vec v ) 
    - \eps' \|\vec w - \vec v\|_2
    \geq  (\alpha_2  - \eps')  \|\vec w - \vec v\|_2
    \;.
\end{equation}
Therefore, we need to choose  $\eps' < \alpha_2$.  Similarly, for the case where $Z_0 < \|\vec w \|
\leq Z_1$  we have that 
\begin{align}
\label{eq:lower-bound-gradient-Z1}
    \nabla \widehat{F}(\vec w) \cdot(\vec w - \vec v)
    &\geq  \nabla F(\vec w) \cdot (\vec w - \vec v) 
    - (\nabla \widehat{F}(\vec w) - \nabla F(\vec w)) \cdot (\vec w - \vec v)  
    \nonumber \\ 
    &\geq  \beta_2 \|\vec w - \vec v\|_{\vec w}
    - \|\nabla \widehat{F}(\vec w) - \nabla F(\vec w)\|_{*,\vec w} \|\vec w - \vec v\|_{\vec w}
    \nonumber
    \\
    &\geq (\beta_2  - \eps') \|\vec w - \vec v\|_{\vec w}
    \;.
\end{align}
Finally, for $\| \vec w\|_2 \geq \zeta Z_1$ we similarly get the  lower bound
\begin{align}
\label{eq:lower-bound-gradient-outer}
    \nabla \widehat{F}(\vec w) \cdot \vec w  \geq  (\gamma - \eps') \|\vec w\|_{\vec w} \,.
\end{align}
Therefore, it suffices to choose $\eps' \leq \min(\alpha_2, \beta_2, \gamma)/2
= (\sqrt{ \eps}) \poly(LR \mu \tau/\xi)$.
Assuming that all the empirical gradients used by the gradient descent satisfy the error bound with 
$\eps'$ as above, from  \Cref{lem:gradient-field-bounded}, we obtain that with step size 
$\eta = \poly(\eps \tau \mu L R/\xi)$ after $T = \poly(1/(\eps \eta))$ iterations, we will have that 
if $\|\vec w^{(T)}\|_2 \leq Z_0$,
then  $\|\vec w^{(T)} - \vec v \|_2 \leq \poly(\xi/ (L R \tau \mu)) \sqrt{ \eps}$
which using \Cref{lem:bound-difference-bounded}, implies that  
$F(\vec w^{(T)}) \leq  \poly(\xi/(LR \mu \tau) ( \eps)$.
Similarly, if $\|\vec w^{(T)}\|_2 > Z_0$, we obtain that $\|\vec w^{(T)} - \vec v\|_{\vec w^{(T)}} 
\leq \poly(\xi/ (L R \tau \mu)) \sqrt{\eps}$
which again implies that $F(\vec w^{(T)}) \leq \poly(\xi/(LR \mu \tau)) \eps$.

\subsection{Unbounded Activations}

We now turn our attention to the case of unbounded activations.  Perhaps surprisingly, we show that
essentially the same algorithm (the only difference is that in this case we set the regularization
parameter to $0$) achieves sample complexity polylogarithmic in $1/\eps$.

\begin{theorem}[Unbounded Activations] \label{thm:unbounded-main-theorem}
Let $\D$ be an $(\eps, W)$-corrupted, $(L,R)$-well-behaved distribution on 
$\R^d \times \R$ and $\sigma(\cdot)$ be an $(\alpha, \lambda)$-unbounded activation.
Set $\kappa = \poly(L R\alpha/\lambda)/(W^2\log(W))$.  The gradient descent \Cref{alg:gd-3} with
step size $\eta = \kappa \eps$, truncation threshold $M=\wt{O}((W/L)$ $\max(\log(\lambda^2
W^2/\eps),1))$, $N = \wt{\Theta}( d/ \kappa \log(1/\delta) \max(\poly\log(1/\eps),1)) $ samples, and $T =  \poly(\log(1/\eps),1/\kappa) $ 
iterations converges to a vector $\vec w^{(T)} \in \R^d$ that, with probability $1-\delta$, satisfies 
\(
F(\vec  w^{(T)} ) \lesssim \frac{1}{L R^4} \Big(\frac{\lambda}{\alpha}\Big)^4 \eps\,.
\) 
\end{theorem}

\begin{Ualgorithm}
	\centering
	\fbox{\parbox{6in}{
			{\bf Input:}   Iterations: $T$, sample access from $\D$, step size: $\eta$, bound $M$. \\
			{\bf Output:}  A weight vector $\vec w^{(T)}$.
			\begin{enumerate}
				\item $\vec w^{(0)} \gets \vec 0$.
				\item For $t=0,\ldots, T$ do
				\begin{enumerate}
					\item  Use fresh $N$ samples from $\D$ truncated at $M$ to obtain $\vec g\tth$, an approximation of the population 
					gradient $\nabla F(\vec w\tth)$ (see \Cref{clm:estimation-gradient}).
					\item $\vec w^{(t+1)} \gets \vec w\tth-\eta \vec{g}\tth$.
				\end{enumerate}
				
			\end{enumerate}
			
	}}
	\caption{Gradient Descent Algorithm for Optimizing the $L_2^2$-Loss for Unbounded Activations}
	\label{alg:gd-3}
\end{Ualgorithm}
\label{ssec:unbounded-algorithm}

\subsubsection*{Proof of \Cref{thm:unbounded-main-theorem}}
We consider the population loss $F(\vec w)= (1/2)\Ey[(\sigma(\vec w\cdot \x)-y)^2]$. As we discussed in \Cref{rem:parameters}, we can assume that $\lambda \geq 1$ and $L\leq 1$ for making the presentation simpler. Moreover, we can also assume that
$W\geq 1$, because if $W<1$, then $\inf_{\|\vec w\|_2\leq 1}F(\vec w)\leq \inf_{\|\vec w\|_2\leq W}F(\vec w)$,
therefore the distribution is $(\eps,1)$-corrupted. Denote $\vec w^\ast$ to be a vector with $\|\wstar\|_2\leq W$ that achieves $F(\vec w^\ast) \leq \eps$.
First, we assume that $\eps\geq \lambda^2 W^2/C$, for some $C\geq 10$. 
Then, any solution $\vec w\in \R^d$ achieves error 
$F(\vec w)\leq 2\eps$. To see this, note that 
\begin{align*}
    F(\vec w)\leq 2F(\wstar) + \Exx[(\sigma(\vec w\cdot \x)-\sigma(\wstar\cdot \x))^2]\leq 2\eps + 4\lambda^2W^2\leq  2C\eps\;.
\end{align*}
Next we consider the case where $\eps\leq  \lambda^2 W^2/C$.
First, using the exponential concentration properties of well-behaved distributions, we observe that
we can truncate the labels $y$ such that $|y|\leq M$, for some $M>0$, without increasing the
level of corruption by a lot.  Given the exponential concentration of the distribution and the fact
that $\|\vec w^\ast\|_2 \leq W$, we show that we can pick $M=\Theta(W\max(\log(W/\eps),1))$ so that
the instance $\wt D$ of $(\x, \wt y)$, where  $\wt y= \sign(y) \min(|y|,M)$, is at most $O(\eps)$-corrupted. Formally, we show:
\begin{claim}\label{clm:truncate-labels-unbounded}
    Let $M=\Theta(W\max(\log(W/\eps),1))$. Denote by $\mathrm{tr}(y)= \sign(y) \min(|y|,M)$. Then, it holds that:
    \[
      \Ey[(\mathrm{tr}(y)-\sigma(\wstar\cdot \x))^2]\leq O(\eps)\;.
    \]
\end{claim}
The proof of the above claim can be found on \Cref{app:optimization}.
Thus, we will henceforth
assume that $|y| \leq M$, and keep denoting the instance distribution as $\D$. 

We first show that, given any gradient field that satisfies the properties given in \Cref{main-pro:gradient-bound},
then gradient descent with an appropriate step size will converge to the target vector $\vec v$. 
The proof of \Cref{lem:gradient-field-unbounded} can be found in \Cref{app:optimization}.
\begin{lemma}[Gradient Field Distance Reduction]
    \label{lem:gradient-field-unbounded}
    Let $\vec g: \R^d \mapsto \R^d$ be a vector field and fix $W, B \geq 1$.
    We assume that $\vec g$ satisfies the following properties with respect to some unknown target vector $\vec v$
    with $\|\vec v\|_2 \leq W$. Fix parameters $\alpha_1, \alpha_2 > 0$.
    For every $\vec w \in \R^d$ with $\|\vec w\|_2 \leq 2 W$ it holds that $\|g(\vec w)\|_2 \leq B\max(\|\vec w - \vec v\|_2,\alpha_1)$.
    Moreover, if $\|\vec w - \vec v\|_2 \geq \alpha_1$ and $\theta(\vec w, \vec v) \in (0, \pi/2)$ then it holds that 
    $\vec g(\vec w) \cdot (\vec w - \vec v) \geq \alpha_2 \|\vec w - \vec v\|_2^2 $. 
    We consider the update rule $\vec w^{(t+1)} \gets \vec w^{(t)} - \eta \vec g(\vec w^{(t)})$ 
    initialized with $\vec w^{(0)} = \vec 0$ and step size  
    \(
    \eta \leq  \alpha_2/B.
    \)
    Let $T$ be any integer larger than  $\lceil (W^2+\log(1/\alpha_1))/(\eta \alpha_2) \rceil$, it 
    holds that $ \| \vec w^{(T)} - \vec v\|_2  \leq  (1+\eta B)\alpha_1 $.
    \end{lemma}
We first show that \Cref{alg:gd-3} with the population gradients would converge after
$\polylog(1/\eps)$ iterations to an approximately optimal solution. 
We have  
\begin{align*}
    \|\nabla F(\vec w )\|_2&= \|\E[(\sigma(\vec w \cdot \x)-y)\sigma'(\vec w \cdot \x)\x]\|_2 
    = \max_{\|\vec u\|_2=1}\E[(\sigma(\vec w \cdot \x)-y)\sigma'(\vec w \cdot \x)\vec u\cdot \x]\;.
\end{align*}
Let $\vec w$ be any vector with $\|\vec w\|_2 \leq W$ and 
let $\vec u$ be any unit vector. 
By adding and subtracting $\sigma(\wstar \cdot \x)$, we get:
\begin{align*}
  \E&[(\sigma(\vec w \cdot \x)-y)\sigma'(\vec w\cdot \x)\vec u\cdot \x]
  \\
  &= \E[(\sigma(\vec w \cdot \x)-y)\sigma'(\vec w \cdot \x)\vec u\cdot \x] +
  \E[(\sigma(\vec w\cdot \x)-\sigma(\wstar \cdot \x))\sigma'(\vec w\cdot \x)\vec u\cdot \x]
  \\ &\leq \lambda  \E[|(\sigma(\wstar\cdot \x)-y)||\vec u\cdot \x|]+
  \E[(\sigma(\vec w \cdot \x)-\sigma(\wstar \cdot \x))\sigma'(\vec w\cdot \x)\vec u\cdot \x]
  \\ &\leq 
  \lambda 
  \E[(\sigma(\wstar \cdot \x)-y)^2]^{1/2}\E[(\vec u\cdot \x)^2]^{1/2} +
  \E[(\sigma(\vec w\cdot \x)-\sigma(\wstar \cdot \x))^2\sigma'(\vec w\cdot \x)^2]^{1/2}\E[(\vec u\cdot \x)^2]^{1/2}
  \\ 
  &\leq \lambda  \sqrt{\eps} + \lambda \|\vec w - \wstar \|_2 \;,
\end{align*}
where for the third inequality we used the fact
the $\x$-marginal of $\D$ is isotropic and 
that $\E[(\sigma(\wstar \cdot \x)-y)^2]^{1/2} \leq \sqrt{\eps}$.
Hence, we have that  $\|\nabla F(\vec w)\|_2 \leq  \lambda  (\sqrt{\eps} +\|\vec w - \wstar \|_2)$.
Using \Cref{main-pro:gradient-bound}, we have that the true gradients point in the right
direction: we have that for any vector 
$\vec w\in\R^d$, with $\theta(\vec w,\wstar)\in(0,\pi/2)$, it holds
that if $\|\vec w - \vec w^\ast\|_2 
\geq  c' \lambda \sqrt{ \eps }   /(L R^4 \alpha^2) $, 
for some absolute constant $c'>0$, then it holds that
 \( \nabla F(\vec w)\cdot(\vec w-\wstar)\geq   (1/2)\alpha^2LR^4\|\vec w-\wstar\|_2^2 \,. \)
 Therefore, the true gradient field of $F$ satisfies the assumptions of \Cref{lem:gradient-field-unbounded}
 with $B= O(\lambda)$, $\alpha_1 =  c' \sqrt{ \eps}/(L R^4) \lambda/\alpha^2$
 and $\alpha_2 =  (1/2)\alpha^2LR^4$.
 
In the following claim, we show that with roughly 
$\wt{O}(d\|\vec w-\wstar\|_2^2/\eps^2)$ samples in each iteration, we can get $\eps'$-estimates of
the population gradients. Its proof can be found in \Cref{app:optimization}.
\begin{claim}\label{clm:estimation-gradient}
Let $\vec w$ be such that $\|\vec w-\wstar\| \geq \sqrt{\eps}$.
Using $N =\wt{O}(d\lambda^4\|\vec w-\wstar\|_2^2W^2\max(\log^2(1/\eps),1)/(L\eps')^2)$ samples, we can compute an empirical 
gradient $\vec g(\vec w)$ such that $\| \vec g(\vec w) - \nabla F(\vec w) \|_2 \leq \eps$ with probability $1-\delta$.
\end{claim}

We now show that the empirical gradients will also satisfy the required 
gradient field assumptions of \Cref{lem:gradient-field-unbounded}.
Assume that we have  $\|\vec g(\vec w) -  \nabla F (\vec w) \|_2 \leq \eps'$
for some $\eps'$ to be specified later.
Using the triangle inequality, we get for any 
$\vec w$ with $\|\vec w\|\leq 2 W$ it holds that 
\begin{equation*}
\|\vec g(\vec w) \|_2  \leq \|\vec \nabla F (\vec w) \|_2 + \eps' \,.
\end{equation*} 
Therefore, for $\eps' \lesssim \lambda(\sqrt{\eps} +\|\vec v-\vec w\|_2)$
we obtain that 
$\|\vec g(\vec w) \|_2  = O(\lambda(\sqrt{\eps} +\|\vec v-\vec w\|_2))$.
We next show that the empirical gradient also points to the direction
of $\vec w - \vec w^\ast$. It holds that 
\begin{equation*}
    \vec g(\vec w) \cdot(\vec w - \wstar)
    \geq  \nabla F(\vec w) \cdot (\vec w - \wstar) 
    - \eps' \|\vec w - \wstar\|_2
    \geq  ( (1/2)\alpha^2 L R^4\|\vec w-\wstar\|_2  - \eps')  \|\vec w - \wstar\|_2
    \;.
\end{equation*}
Therefore, we need to choose  $\eps' = O(\alpha^2 L R^4\|\vec w-\wstar\|_2)$.
Using that  $\|\vec w - \vec w^\ast\|_2 
\geq  c' \lambda \sqrt{ \eps }   /(L R^4 \alpha^2) $, we have that the estimated
gradient field of $F$ satisfies the assumptions of \Cref{lem:gradient-field-unbounded}
 with $B= O(\lambda)$, $\alpha_1 =  c' \sqrt{ \eps}/(L R^4) \lambda/\alpha^2$
 and $\alpha_2 =   (1/2)\alpha^2LR^4 $.
Conditioning that all the empirical gradients used by the gradient descent 
satisfy the error bound with $\eps'$ as above from 
\Cref{lem:gradient-field-unbounded} we obtain that 
with step size $\eta = \poly(L R\alpha/\lambda)$
after $T = \poly(1/(\alpha L R)) W^2\log(1/\alpha_1)$
iterations we will have that $\|\vec w^{(T)} - \vec w^\ast\|_2 \leq 
\poly(1/(LR)) \lambda/\alpha^2 \sqrt{\eps} $ which implies that 
\begin{align*}
F(\vec w^{(T)}) &\leq   2 
\E[(\sigma(\vec w^{\ast} \cdot \x) - y) ^2] 
+ 
2 \E[(\sigma(\vec w^\ast \cdot \x) - \sigma(\vec w^{(T)} \cdot \x ))^2] 
\\
&= 
\poly(1/(LR)) ~ (\lambda/\alpha)^4 ~ O(\eps),
\end{align*}
where we used the fact that $\sigma$ is $\lambda$-Lipschitz and that
the $\x$-marginal of $\D$ is isotropic.
Since we have to do a union bound over all $T$ iterations, 
and we draw fresh samples in each round, we need to divide $\delta$ by $T$ in each round; 
hence, the total sample complexity is
$N = \poly(\lambda/(\alpha LR) ) ~ \wt{O} (dW^2\log^3(1/\eps) \log(1/ \delta) )$ 
and the runtime $\poly(\lambda/(\alpha LR))\log(1/\eps)N$.

 \bibliographystyle{alpha}
\bibliography{mydb}
\appendix

\section{Omitted Proofs from \Cref{main-sec:bounded-plain-landscape}}\label{app:bounded-plain-landscape}

\subsection{Proof of \Cref{theorem:intro_bad_stationary}}
We restate and prove the following proposition.
\begin{proposition}[Bounded Activations Have ``Bad'' Stationary Points] 
\label{theorem:bad_stationary}
Fix $\eps \in (0,1]$ and let $r(t)$ be the ramp activation.
There exists a distribution $\D$ on $\R^d \times \{\pm 1 \}$
with standard Gaussian $\x$-marginal such that there exists a vector $\vec v$ with $F(\vec v) = (1/2) \E_{(\x, y) \sim \D}[(y - r(\vec v \cdot \x))^2] \leq O(\eps)$ and a vector $\vec u$ with $\nabla F(\vec u) = 0$,
$\nabla^2 F(\vec u)  \preceq 0$, i.e., $\vec u$ is a local-minimum of $F$, and  $F(\vec u) \geq 1/2$.
\end{proposition}
\begin{proof}
We define the following deterministic noise function 
\\
\begin{minipage}[c]{0.5\textwidth}
\[
\wh{r}(t)= 
\begin{cases} 
-1  \quad \text{ if }   t \leq -2 \\
+1  \quad \text{ if }   -2 \leq t \leq -1 \\
-t \quad \text{ if } -1 \leq t \leq 1 \\
-1 \quad \text{ if } -1 \leq t \leq 2 \\
+1 \quad\text{ if } t \geq  2
\end{cases}
\]
\end{minipage}
~~~~~~
 \begin{minipage}[c]{.5\textwidth}
 \ifplots
 \begin{tikzpicture}
 \begin{groupplot}[group style={group size=3 by 2},height=3cm,width=3cm,
 restrict y to domain=-4:4,xlabel=$t$,ylabel=$y$, legend pos=north west,axis y line =middle,
 axis x line =middle,
 				axis on top=true,
 				xmin=-4,
 				xmax=4,
 				ymin=-1.5,
 				ymax=1.5,
 				y=1cm,
                x=0.6cm,
                  yticklabels={,,},
 				    xticklabels={,,},
 				      ytick=\empty,
 				        xtick=\empty,]
				  \nextgroupplot[title=${\wh{r}(t)}$]
				\addplot[color=red, samples=100, domain=-4:-2] {-1};
				\addplot[color=red, samples=100, domain=-2:-1] {+1};
				\addplot[color=red, samples=100, domain=-1:1] {-x};
				\addplot[color=red, samples=100, domain=1:2] {-1};
				\addplot[color=red, samples=100, domain=2:4] {+1};
				\addplot[color=blue,dashed, samples=100, domain=-4:-1] {-1};
				\addplot[color=blue,dashed, samples=100, domain=-1:1] {x};
				\addplot[color=blue,dashed, samples=100, domain=1:4] {1};
			\end{groupplot}
\end{tikzpicture}
\fi
 \end{minipage}
 \\
 In what follows, we denote by $\gamma(t) = \frac{1}{\sqrt{2 \pi}} e^{-t^2/2}$ the density of the standard normal distribution. It suffices to consider noise that only depends on a single direction,
 i.e., $y(\vec x) = \wh{r}(\vec x_1/\eps)$. We first show that there exists a vector with small $L_2^2$ loss.
 Take $\vec v = \vec e_1/\eps$, where $\vec e_1 = (1,0,\ldots,0)$ is the first vector of the standard orthogonal basis of $\R^d$. 
 It holds that
 \begin{align*}
 F(\vec v)
 &= (1/2) \Exx[ (r(\x_1/\eps) - y(\x_1))^2]
 = (1/2) \Exx[ (r(\x_1/\eps) - \wh{r}(\x_1/\eps))^2]
 \\
 &= (1/2) \int_{-\eps}^{\eps} (t/\eps + t/\eps)^2 \gamma(t) d t +\int_{-2\eps}^{-\eps} 4 \gamma(t) d t
 \leq \frac{\sqrt{2}}{\sqrt{\pi} \eps^2}  \int_{-\eps}^{\eps} t^2  d t +4\eps
 = O(\eps) \,.
 \end{align*}
 We next show that there exists a ``bad'' stationary point $\vec u$, i.e.,
 a $\vec u$ with $\nabla F(\vec u) = 0$ and $F(\vec u) \geq 1/2$.
 We have 
 \[
 \nabla F(\vec u) = 
 \Ex[(r(\vec u \cdot \x) - y(\x) ) r'(\vec u \cdot \x) \x] \,.
 \]
 We shall take $\vec u = - \vec e_1/\eps$, i.e., the opposite direction of the almost
 optimal vector $\vec v$ that we used above.  Using the coordinate-wise independence of the Gaussian
 distribution, we have that for every orthogonal direction $\vec e_i$, $i\geq 2$, it holds that 
 $\nabla F(\vec u) \cdot \vec e_i = 0$.  For the direction $\vec e_1$, we obtain
 \begin{align*}
 \nabla F(\vec u) \cdot \vec e_1 &= 
 \Ex[ ( r(-\x_1/\eps) - \wh{r}(\x_1/\eps) ) ~ \1\{|\x_1| \leq \eps\} \x_1] 
 \\
 &=  \Ex[ ( r(-\x_1/\eps) - \wh{r}(-\x_1/\eps) ) ~ \1\{|\x_1| \leq \eps\} \x_1] 
 = 0 \,.
 \end{align*}
 We next proceed to show that $\vec u$ is a local minimum of the population $L_2^2$-loss $F$.  We
 compute the Hessian of $F$ at $\vec u$.  In what follows, we denote by $\delta(t)$ the standard
 Dirac delta function.  We have that
 \begin{align*}
 \nabla^2 F(\vec u) 
 &= \Exx[ (r'(\vec u \cdot \x))^2 \vec x \vec x^T]
 + 
 \Exx[ (r(\vec u \cdot \x) - y(\vec x)) 
 (\delta(\vec u\cdot \x + 1) - \delta(\vec u\cdot \x - 1) )
  \vec x \vec x^T]
  \,.
 \end{align*}
 For $i \neq j$, using the fact that the Gaussian marginals are independent, we have that $(\nabla^2 F(\vec u))_{ij} = 0$.
 We next observe that the second term of the Hessian vanishes.  We have 
 \begin{align*}
 \Exx[ (r(\vec u &\cdot \x) - y(\vec x)) 
 (\delta(\vec u\cdot \x + 1) - \delta(\vec u\cdot \x - 1) )
  \x_i^2] 
  \\
  &= 
 \Exx[ (r(-\x_1/\eps) - r(\x_1/\eps)) 
 (\delta(\x_1/\eps + 1) - \delta(\x_1/\eps - 1) )
  \x_i^2] 
  = 0 \,,
  \end{align*}
  where we used the fact that both $r(-\x_1/\eps), \wh{r}(\x_1/\eps)$ are continuous at $\x_1 = \pm \eps$, and it holds $r(\pm 1) = \wh{r}(\pm 1)$.

To complete the proof, we need to show that the $L_2^2$ loss 
of $\vec u$ is large. It holds that
 \begin{align*}
 F(\vec u) &= 
 \frac12 \Ex[ ( r(-\x_1/\eps) - r'(\x_1/\eps) )^2]
 \\
 &\geq 
 \frac12 
 \left( \int_{-\infty}^{-\eps}  ( r(-t/\eps) - r'(t/\eps) )^2
 \gamma(t) d t 
 +
 \int_{\eps}^{+\infty} 
 ( r(-t /\eps) - r'(t/\eps) )^2
 \gamma(t) d t 
 \right)
 \\
 &\geq 
 \frac12 
 \left( \int_{-\infty}^{-\eps}  4 \gamma(t) d t 
 +
 \int_{\eps}^{+\infty}  4 \gamma(t) d t  \right)
 \geq 4 \int_{1}^\infty \gamma(t) d t \geq 1/2 \,. 
 \end{align*}
This completes the proof of \Cref{theorem:intro_bad_stationary}.
\end{proof}

\subsection*{Proof of \Cref{lem:radius-approx}}
We restate and prove the following lemma.
\begin{lemma}[Radius of Approximate Optimality of Sigmoidal Activations]
Let $\D$ be an $(L,R)$-well-behaved distribution in $\R^d$ and let $\sigma(t)$ be a $(\tau, \boundc,
\boundd)$-sigmoidal activation function.  There exists a vector $\vec v$ with $\|\vec v\|_2 \leq
1/\eps $ and $F(\vec v) \leq (1+O(\frac{\xi}{ \mu L }))\eps $.
\end{lemma}
\begin{proof}
We first observe that since $F(\vec w)$ is a continuous function of $\vec w$,
there exists a vector $\vec w^\ast$ with $\|\vec w^\ast\|_2 < +\infty$ 
such that $F(\vec w^\ast) \leq 2 \eps$.
If $\eps \geq 1$, then the zero vector achieves $4\eps$ error since,  
using the inequality $(a+b)^2 \leq 2 a^2 + 2 b^2$, we have that 
$ F(\vec 0) = \E[ y^2 ] \leq 2 \E[ (y - \sigma(\vec w^\ast\cdot \x))^2 ]  +  2 \E[ \sigma(\vec w^\ast\cdot \x)^2 ]  
\leq 2 \eps + 2\eps 
\leq 4 \eps $. 
Denote $ 1/\eps :=\rho$.
If $\|\vec w^\ast\|_2 \leq \rho$, then we are done.
If $\|\vec w^\ast \|_2 > \rho$, then we consider 
a scaled down version of $\vec w^\ast$, namely,
the vector $\vec v = \rho ~ \vec w^\ast/\|\vec w^\ast\|_2$.
Using again the inequality $(a+b)^2 \leq 2 a^2 + 2 b^2$, 
we have 
\[
F(\vec v) \leq 2 F(\vec w^\ast) + 2 \Exx [ (\sigma(\vec v \cdot \x) - \sigma (\vec w^\ast \cdot \x))^2] 
\,.
\]
Since $\vec v$ is parallel to $\vec w^\ast$ and its norm is smaller than $\|\vec w^\ast\|$, 
we have 
that 
\begin{align*}
\Exx[(\sigma(\vec v \cdot \x) - \sigma(\vec w^\ast \cdot \x))^2]
&\leq  
\frac{1}{L} 
\int_{-\infty}^{+\infty}
(\sigma(\|\vec v\|_2 t) - \sigma(\|\vec w^\ast\|_2 |t| )^2  d t
\\
&\leq
\frac{1}{L} 
\int_{-\infty}^{+\infty}
\Big(\int_{\|\vec v\| t}^{+\infty} \sigma'(z) d z \Big)^2  d t
\\
&\leq
\frac{1}{L} 
\int_{-\infty}^{+\infty} (\xi e^{-\mu \|\vec v\| |t|})^2  d t
\\
&=
\frac{\xi}{ \mu L \|\vec v\|} 
\,,
\end{align*}
where we used the fact that the density of $1$-dimensional marginals of $\D_\x$ 
is bounded from above by $1/L$, see \Cref{def:bounds}, and the 
fact that $\sigma'(t) \leq \xi e^{-\mu |t|}$.
We see that by choosing $\|\vec v\|_2 = 1/\eps$
it holds that 
$\Exx[(\sigma(\vec v \cdot \x) - \sigma(\vec w^\ast \cdot \x))^2] = O(\xi/(\mu L))\eps$.
Thus, combining the above we have that there always exists a 
$\vec v$ with $\|\vec v\|_2 =O(1/ \eps$
that achieves $L_2^2$ loss $F(\vec v) \leq  (1+O(\frac{\xi}{\mu L}))\eps$.
\end{proof}

\subsection{Proof of \Cref{prop:gradient-bound-ramp}}
We restate and prove the following proposition.
\begin{proposition}[Gradient of the ``Vanilla'' $L_2^2$ Loss (Inside a Ball)]
 Let $\D$ be an $(L,R)$-well-behaved distribution
 and let $\sigma$ be a $(\tau, \boundc, \boundd)$-sigmoidal activation. 
 Let $\eps\in(0,1)$ and let $\wstar \in \R^d$ with $F(\wstar)\leq \eps$ and $F(\wstar)$ is less than a sufficiently small multiple of $L^2R^6\tau^4/\boundd^2$. Denote $\kappa=L^6 R^6\boundc^3\tau^4/\boundd^2$.
There exists a universal constant $c' > 0$, such that for any $\vec w\in \R^d$ with $\| \vec w\|_2\leq c'\kappa/\eps$, it holds that
\begin{enumerate}
\item 
 When $\|\vec w\|_2\leq 2/R$ and $\|\wstar-\vec w\|_2\geq \sqrt{\eps}\boundd/(c'L R^4\tau^2)$, then
 \[
\nabla F(\vec w)\cdot(\vec w-\wstar) \geq c'\sqrt{\eps}\|\wstar-\vec w\|_2 \,. \]
\item  When $\|\vec w\|_2 \geq 2/R$ and either $\|\wstar-\vec w\|_{\vec w}\geq \sqrt{\eps/\kappa}$ or $\|\vec w\|_2\geq 2\|\wstar\|_2$, then
\[
\nabla F(\vec w)\cdot(\vec w-\wstar) \geq c' \sqrt{\eps}\|\wstar-\vec w\|_{\vec w}\,. \]
  \end{enumerate}
\end{proposition}

\begin{proof}[Proof of \Cref{prop:gradient-bound-ramp}]
We decompose the gradient into a part that corresponds to the contribution of the ``true'' labels, i.e.,
$\sigma(\wstar \cdot \x)$ (see $I_2$ below), and a part corresponding to the noise (see $I_1$ below).  We have that
\begin{align}
    \nabla F(\vec w)\cdot(\vec w-\wstar)&=\underbrace{\E[(\sigma(\wstar\cdot \x)-y)\sigma'(\vec w\cdot \x)(\vec w\cdot \x-\wstar \cdot \x)]}_{I_1} 
    \nonumber
    \\ &+\underbrace{\E[(\sigma(\vec w\cdot \x)-\sigma(\wstar\cdot \x))\sigma'(\vec w\cdot \x)(\vec w\cdot \x-\wstar \cdot \x)]}_{I_2}\;.
    \label{eq:gradient-decomposition-bounded}
\end{align}
In the following subsections, we bound $I_1$ and $I_2$ from below. We start by bounding from below the effect of the noise, i.e., the  contribution of $I_1$.
\paragraph{Estimating the Effect of the Noise}
We start by showing that the noise cannot affect the gradient by a lot, i.e., we bound the contribution of $I_1$. We prove the following lemma:
\begin{lemma}\label{lem:noise-coontribution-ramp}
 Let $\D$ be an $(L,R)$-well-behaved distribution and $\sigma$ be a $(\tau,\mu,\xi)$-sigmoidal activation. For any vector $\vec w\in \R^d$, it holds
 that
 \[
 I_1\geq -\sqrt{8\boundd^2\eps}\min(\|\wstar-\vec w\|_{\vec w}/(\boundc^{3/2}L^{2}),\|\wstar-\vec w\|_2 )\;.
 \]
\end{lemma}
\begin{proof}
Using the \CS\ inequality, we obtain:
\begin{align}
\label{eq:I_1-bound}
   I_1&\geq
   -\E[|(\sigma(\wstar\cdot \x)-y)\sigma'(\vec w\cdot \x)(\vec w\cdot \x-\wstar \cdot \x)|]
   \nonumber
   \\
   &\geq -  (\E[(\sigma(\wstar\cdot \x)-y)^2])^{
    1/2}
    (\E[(\vec w\cdot \x-\wstar \cdot \x)^2\sigma'(\vec w\cdot \x)^2])^{1/2}
   \nonumber
    \\
    &= -\sqrt{F(\wstar)}  
    ~
    (\E[(\vec w\cdot \x-\wstar \cdot \x)^2\sigma'(\vec w\cdot \x)^2])^{1/2} \nonumber
       \\
    &\geq -\sqrt{\eps}  
    ~
    (\E[(\vec w\cdot \x-\wstar \cdot \x)^2\sigma'(\vec w\cdot \x)^2])^{1/2} \,,
\end{align}
where we used that $F(\wstar)\leq \eps$, from the assumptions of \Cref{prop:gradient-bound-ramp}.
We proceed to bound the term  
$\E[(\vec w\cdot \x-\wstar \cdot \x)^2\sigma'(\vec w\cdot \x)^2]$.
Note that we can use the upper bound on the derivative of the activation function,
i.e., $\sigma'(t) \leq \boundd$ for all $t \in \R$.  However, this would 
result in $\E[(\vec w\cdot \x-\wstar \cdot \x)^2\sigma'(\vec w\cdot \x)^2]
\leq O(\boundd^2\|\vec w - \wstar\|_2^2)$. While, this was sufficient for the case of unbounded
activations of the \Cref{def:well-behaved-unbounded-intro}, for bounded activation functions, 
we need a tighter estimate that takes into account the fact that the functions have exponential tails outside the interval $[-1/\boundc, +1/\boundc]$. 
Recall that we denote by $\vec q^{\|_{\wstar}}$ the component  
of $\vec q\in \R^d$ parallel to $\wstar\in \R^d$, i.e., 
$\vec q^{\|_{\wstar}} = \proj_{\wstar} \vec q$. Similarly, 
we denote $\vec q^{\perp_{{\wstar}}} = \proj_{{\wstar}^\perp} \vec q$.
We prove the following.
\begin{lemma}\label{lem:noise-contribution-new}
 Let $\D$ be an $(L,R)$-well-behaved distribution and $\sigma$ be a $(\tau,\mu,\xi)$-sigmoidal activation. For any vectors $\vec v,\vec w\in \R^d$, it holds
 that
\[
\Exx[(\vec w \cdot \x - \vec v \cdot \x)^2 (\sigma'(\vec w\cdot \x))^2]
\leq  8  \boundd^2
     \min
     \left(\frac{1}{L^4\boundc^3}\|\vec v-\vec w\|_{\vec w}^2
     ,
     \|\vec v-\vec w\|_2^2
     \right)\;.
\]
\end{lemma}

\begin{proof}
First note that $
    \Exx[ (\vec w \cdot \vec x- \vec v \cdot \vec x)^2  \sigma'(\vec w \cdot \vec x)^2 ] 
    \leq
   \boundd^2 \Exx[ (\vec w \cdot \vec x- \vec v \cdot \vec x)^2 \exp(-2|\vec w\cdot \x| \boundc) ]$, from the assumption that $|\sigma'(t)|\leq \xi \exp(-\mu|t|)$.
   It holds that
    \begin{align*}
    \Exx[ (\vec w \cdot \vec x- \vec v \cdot \vec x)^2  \exp(-2|\vec w\cdot \x| \boundc)
    ]\leq \Exx[ (\vec w \cdot \vec x- \vec v \cdot \vec x)^2]= \|\vec w-\vec v\|_2^2\;,
\end{align*}
where  we used the fact that the distribution $\D_\x$ is isotropic, i.e., for any vector $\vec u\in \R^d$, it holds that
$\Exx[(\vec u\cdot \x)^2] = \|\vec u\|_2^2$.

Next show that by using the tails of the distribution $\D_\x$, we can prove a tighter upper bound for some cases. We use that the distribution is $(L,R)$-well-behaved, and we prove the following claim
that bounds from above the expectation.
\begin{claim}\label{clm:bound-square-conce}
Let $\D_\x$ be an $(L,R)$-well-behaved distribution. Let $b>0$ and $\vec u, \vec v$ be unit norm
orthogonal vectors. It holds that
\[
\Exx[ (\vec u\cdot\x)^2 \exp(-b |\vec v\cdot \x| ) ]\leq \frac{8}{L^4b} \quad \Exx[ (\vec v\cdot\x)^2 \exp(-b|\vec v\cdot \x|)]\leq \frac{8}{L^2b^3} \;.
\]
\end{claim}
\begin{proof}
Without loss of generality take $\vec u = \vec e_1$ and $\vec v = \vec e_2$.
Using the fact that the distribution is $(L,R)$-well-behaved, we have that 
the $2$-dimensional projection on the subspace $V$ spanned by $\vec v, \vec u$ is bounded
from above by $(1/L)\exp(- L\|\x_V\|_2)$ everywhere. We have that
\begin{align*}
    \Exx[ \x_1^2\exp(-b \vec v\cdot \x ) ]&\leq (1/L)\int_{\x_2\in \R}\int_{\x_1\in \R}\x_1^2\exp(-b \vec v\cdot \x )\exp(- L\|\x_V\|_2)\d\x_1\d\x_2
    \\&\leq (4/L)\int_{0}^{\infty}\x_1^2\exp(-L|\x_1|)\int_{0}^{\infty} \exp(-b|\x_2|)\d\x_1\d\x_2
    \\&= 8/(L^4b)\;.
\end{align*}
Putting the above estimates together, we proved the first part of \Cref{clm:bound-square-conce}. For the other part, it holds that 
\begin{align*}
    \Exx[ (\vec v\cdot\x)^2\exp(-b \vec v\cdot \x ) ]\leq (4/L)\int_{0}^\infty\int_{0}^{\infty} \x_2^2\exp(-L|\x_1|)\exp(-b |\x_2|) \d \x_2\d \x_1=  8/(L^2b^3)\;.
\end{align*}
This completes the proof of \Cref{clm:bound-square-conce}.
\end{proof}
    Let $\vec q=\vec v-\vec w$ and note that 
    $(\vec q \cdot \vec x)^2 = (\vpar{q}{w} \cdot \x)^2 + (\vperp{q}{w}\cdot\x)^2$. Using \Cref{clm:bound-square-conce}, it holds that
    \begin{align*}
    \Exx[ (\vec q\cdot \x)^2 \exp(-2\vec w\cdot \x \boundc)
    ]
     &= \Exx[ (\vpar{q}{w} \cdot \x)^2 + (\vperp{q}{w}\cdot\x)^2) \exp(-2\vec w\cdot \x \boundc) ]
     \\&\leq     \frac{8}{L^4}( 
     \|\vec q^{\|_{\vec w}}\|_2^2 \frac{1}{\boundc^3\|\vec w\|_2^3}
     + 
     \|\vec q^{\perp_{\vec w}}\|_2^2 \frac{1}{\boundc\|\vec w\|_2})\leq \frac{8\|\vec q\|_{\vec w}^2}{L^4\boundc^3}\;,
\end{align*}
where we used that $a^2+b^2\leq (a+b)^2$ for $a,b\geq 0$. This completes the proof of \Cref{lem:noise-contribution-new}.
\end{proof}
\Cref{lem:noise-coontribution-ramp} follows from combining \Cref{eq:I_1-bound} along with \Cref{lem:noise-contribution-new}.
\end{proof}
\paragraph{Estimating the Contribution of the ``Noiseless'' Gradient}
In order to compute the contribution of the gradient when there is no noise to the instance, we show that in fact the contribution of $I_2$ is bounded from below by the contribution of a ramp function instead of $\sigma$. To show this, we use the property that 
$\sigma'(t)\geq \tau$, for all $t\in[-1,1]$, see \Cref{def:well-behaved-bounded-intro}.
\begin{claim}\label{clm:sig2ramp} It holds that $I_2\geq \tau^2\E_{\x\sim \D_\x}[(r(\vec w\cdot \x)-r(\vec v\cdot\x))r'(\vec w\cdot \x) (\vec w-\vec v)\cdot \x ]$.
\end{claim}
\begin{proof}
We have that
\begin{align*}
    I_2&=\E_{\x\sim \D_\x}[(\sigma(\vec w\cdot \x)-\sigma(\vec v\cdot\x))\sigma'(\vec w\cdot \x) (\vec w-\vec v)\cdot \x]
    \\&=\E_{\x\sim \D_\x}[|(\sigma(\vec w\cdot \x)-\sigma(\vec v\cdot\x))\sigma'(\vec w\cdot \x)(\vec w-\vec v)\cdot \x|]
    \\&\geq \tau E_{\x\sim \D_\x}[|(\sigma(\vec w\cdot \x)-\sigma(\vec v\cdot\x))(\vec w-\vec v)\cdot \x|\1\{|\vec w\cdot \x|\leq 1\}]\;,
\end{align*}
where we used that $\sigma$ is non-decreasing and that $\sigma(t)\geq \tau$ for $t\in[-1,1]$ from the assumptions of $(\tau,\boundc,\boundd)$-sigmoidal activations. Moreover, note that from the fundamental theorem of calculus, we have that
\begin{align*}
    \sigma(\vec w\cdot \x)-\sigma(\vec v\cdot\x)&=\int_{\vec v\cdot \x}^{\vec w\cdot\x}\sigma'(t)\d t
    \geq \tau\int_{\vec v\cdot \x}^{\vec w\cdot\x}\1\{|t|\leq 1\}\d t=\tau ( r(\vec w\cdot \x)-r(\vec v\cdot\x))\;,
\end{align*}
where we used again that $\sigma(t)\geq \tau$ for $t\in[-1,1]$. Therefore, we have that
\[
I_2\geq \tau^2\E_{\x\sim \D_\x}[(r(\vec w\cdot \x)-r(\vec v\cdot\x))r'(\vec w\cdot \x) (\vec w-\vec v)\cdot \x ]\;.
\]
This completes the proof of \Cref{clm:sig2ramp}.
\end{proof}
To bound $I_2$ from below, we consider three cases depending on how far the vector $\vec w$ is from the target vector $\wstar$.
The first two cases correspond to $\theta(\vec w,\wstar)\in (0,\pi/2)$. In the first case, we have that either $\|\proj_{\vec w^\perp}\wstar\|_2 \geq 2K/R$, for any $K\geq 1$, or $\|\proj_{\vec w}\wstar\|_2\geq 2\|\vec w\|_2$ and in the second case when both of the aforementioned conditions are false. The last case corresponds to $\theta(\vec w,\wstar)\in (\pi/2,\pi)$. Notice that when $\vec w$ is close to the target $\wstar$ then its projection
onto the orthogonal complement of $\vec w$, i.e., $\|\proj_{\vec w^\perp}\wstar\|_2 $, 
will be small and also its projection on the direction of $\vec w$, i.e., $\|\proj_{\vec w}\wstar\|_2$, 
will be close to $\|\vec w\|_2$.

If either of the conditions of the first case are satisfied or when $\vec w\cdot\wstar\leq 0$, then there is a large enough region that we can substitute the $r(\wstar\cdot \x)$ by the constant function $r(\wstar\cdot \x)=1$ (left plot in \Cref{fig:contribution_large-angle}), these conditions corresponds to the case that the vectors $\vec w$ and $\wstar$ are ``far'' apart from each other. Whereas if both of these conditions are not satisfied then there exists a large enough region that $r(\wstar\cdot \x)=\wstar\cdot \x$ (right plot in \Cref{fig:contribution_large-angle}), in this case the vectors $\vec w$ and $\wstar$ are ``close'' to each other.
Without loss of generality, we may assume 
that $\vec w/\|\vec w\|_2 = \vec e_2$ and $\wstar =\|\wstar\|_2 (\cos \theta \vec e_2 -\sin\theta \vec e_1)$.
For simplicity, we abuse notation and denote by $\D_\x$ the marginal distribution on the subspace spanned by the vectors 
$\vec w,\wstar$.

\paragraph{Case 1: $\vec w$ and $\wstar$ are ``close''}
We now handle the case where $\theta(\vec w,\wstar)\in(0,\pi/2)$, $\|\proj_{\vec w^\perp}\wstar\|_2
\leq 2K/R$ and $\|\proj_{\vec w}\wstar\|_2\leq 2\|\vec w\|_2$, i.e., the case where $\vec w$ and
$\wstar$ are close to each other but still not close enough to guarantee small $L^2$ error, i.e., the $L_2^2$ error of $\vec w$,
$F(\vec w)$ is much larger than $\eps$.  

\begin{lemma}\label{lem:lower-bound-square}
Let $\D$ be an $(L,R)$-well-behaved distribution. For any vector $\vec w\in \R^d$, let $\theta=\theta(\wstar,\vec w)\in(0,\pi/2)$. If $\|\proj_{\vec w^\perp}\wstar\|_2 \leq 2K/R$, for $K\geq 1$, and $\|\proj_{\vec w}\wstar\|_2\leq 2\|\vec w\|_2$, it holds:
\begin{itemize}
    \item If $\|\vec w\|_2\geq 2/R$, then
 $I_2\geq \frac{\tau^2L R^3 }{4096K^3}\|\vec w-\wstar\|_{\vec w}^2$.
\item Otherwise, $
 I_2\geq \frac{\tau^2L R^4 }{2048 K^3}\|\vec w-\wstar\|_2^2$.
\end{itemize}
\end{lemma}
\begin{proof}
Using \Cref{clm:sig2ramp}, we have that
\begin{align*}
  I_2&\geq \tau^2\E_{\x\sim \D_\x}[(r(\vec w\cdot \x)-r(\wstar\cdot\x))r'(\vec w\cdot \x) (\vec w-\wstar)\cdot \x]
\\&\geq 
\tau^2\E_{\x\sim \D_\x}[(\vec w\cdot \x-r(\wstar\cdot\x)) (\vec w-\wstar)\cdot \x \1\{0\leq \vec w\cdot \x \leq 1\}]\;.  
\end{align*}
Note that in the last inequality we used that because $r(\cdot)$ is a non-decreasing function, it holds that
$(r(\vec w\cdot \x)-r(\wstar\cdot\x)) (\vec w-\wstar)\cdot \x \geq 0$ for any value of $\x\in \R^d$. Moreover, we assume without loss of generality that $\vec w/\|\vec w\|_2 = \vec e_2$, and therefore $\wstar =\|\wstar\|_2( \cos \theta \vec e_2 -\sin\theta \vec e_1)$.
Note that in this case the condition $\|\proj_{\vec w}\wstar\|_2\leq 2\|\vec w\|_2$ is equivalent to $2\|\vec w\|_2\geq |\cos\theta| \|\wstar\|_2$.

Consider the region $-R /(4K)\leq \x_1\leq -R/(8K)$ and $\x_2\leq 1/(4\|\vec w\|_2)$ which is chosen to guarantee that $0\leq \wstar\cdot \x \leq 1$, which holds because
\[ 0\leq \wstar\cdot \x\leq \|\wstar\|_2\cos\theta \x_2+ R\|\wstar\|_2\sin\theta /(4K)\leq 1/2+\cos\theta\|\wstar\|_2/(4\|\vec w\|_2)\leq 1\;.\] 
We show the following claim which will be used to bound from below $I_2$ for this case.
\begin{claim}\label{clm:orthogonal-deriv}
Let $\D_\x$ be an $(L,R)$-well-behaved distribution. For any vectors $\vec w,\vec v\in \R^d$ and $a,b\in [0,R]$, let $\vec q=\vec w-\vec v$, it holds
\[
\E_{\x\sim \D_\x}\left[(\vec q\cdot\x)^2\1\left\{\frac{\vec w}{\|\vec w\|_2}  \cdot\x\in(0,a),
\frac{\vec v^{\proj_{\vec w}}}{\|\vec v^{\proj_{\vec w}}\|}_2  \cdot\x\in(-b,0)\right\} \right]\geq \frac{L a b}{8}(\|\vec q^{\|_{\vec w}}\|_2^2a^2+ \|\vec q^{\perp_{\vec w}}\|_2^2b^2)\;.
\]
\end{claim}
\begin{proof}
We have that $\vec q=\vec q^{\|_{\vec w}}+\vec q^{\perp_{\vec w}}$ and using the Pythagorean theorem we have that $((\vec w-\vec v) \cdot \x)^2=\|\vec q^{\|_{\vec w}}\|_2^2 (\vec e_1\cdot \x)^2+ \|\vec q^{\perp_{\vec w}}\|_2^2 (\vec e_2\cdot \x)^2$. Therefore, we have that
\begin{align*}
    \E_{\x\sim \D_\x}[(&(\vec w-\vec v)\cdot\x)^2\1\{\vec e_1\cdot\x\in(0,a),\vec e_2\cdot\x\in(-b,0)\}]
        \\&\geq \E_{\x\sim \D_\x}[((\vec w-\vec v)\cdot\x)^2\1\{\vec e_1\cdot\x\in(\frac{a}{2},a),\vec e_2\cdot\x\in(-b,-\frac{b}{2})\}]
    \\&\geq \E_{\x\sim \D_\x}[(\|\vec q^{\|_{\vec w}}\|_2^2 (\vec e_1\cdot \x)^2+ \|\vec q^{\perp_{\vec w}}\|_2^2 (\vec e_2\cdot \x)^2)\1\{\vec e_1\cdot\x\in(\frac{a}{2},a),\vec e_2\cdot\x\in(-b,-\frac{b}{2})\}]
    \\&\geq \frac{1}{4}(\|\vec q^{\|_{\vec w}}\|_2^2a^2+ \|\vec q^{\perp_{\vec w}}\|_2^2b^2\E_{\x\sim \D_\x}[\1\{\vec e_1\cdot\x\in(\frac{a}{2},a),\vec e_2\cdot\x\in(-b,-\frac{b}{2}\}]
    \\&\geq \frac{L a b}{8}(\|\vec q^{\|_{\vec w}}\|_2^2a^2+ \|\vec q^{\perp_{\vec w}}\|_2^2b^2)\;,
\end{align*}
where in the last inequality we used that the distribution $\D_\x$ is $(L,R)$-well-behaved.
\end{proof}
For the case where $\|\vec w\|_2\geq 2/R$. Using the \Cref{clm:orthogonal-deriv}, we have that
\begin{align*}
         \E_{\x\sim \D_\x}[( (\vec w-\vec v)\cdot \x)^2 &\1\{0\leq \x_2 \leq 1/(4\|\vec w\|_2), -R/(4K)\leq \x_1\leq -R/(8K)\}]
         \\&\geq \frac{L R}{128K\|\vec w\|_2}\left(\|\vec q^{\|_{\vec w}}\|_2^2\frac{1}{16\|\vec w\|_2^2} +\|\vec q^{\perp_{\vec w}}\|_2^2\frac{R^2}{K^216} \right)
         \\&\geq \frac{L R^3}{K^32048}\left(\frac{\|\vec q^{\|_{\vec w}}\|_2^2}{\|\vec w\|_2^3} +\frac{\|\vec q^{\perp_{\vec w}}\|_2^2}{\|\vec w\|_2} \right)\geq \frac{L R^3}{4096K^3}\|\vec q\|_{\vec w}^2\;.
\end{align*}
For the case  where $\|\vec w\|_2 \leq 2/R$, using \Cref{clm:orthogonal-deriv}, we have
\begin{align*}
         \E_{\x\sim \D_\x}[( &(\vec w-\vec v)\cdot \x)^2 \1\{0\leq \x_2 \leq 1/(4\|\vec w\|_2), -R/(4K)\leq \x_1\leq -R/(8K)\}]
         \\&\geq \E_{\x\sim \D_\x}[( (\vec w-\vec v)\cdot \x)^2 \1\{R/8\leq \x_2 \leq R/4, -R/(4K)\leq \x_1\leq -R/(8K)\}]
          \\&\geq \frac{L R^2}{K128}\left(\|\vec q^{\|_{\vec w}}\|_2^2\frac{R^2}{16} +\|\vec q^{\perp_{\vec w}}\|_2^2\frac{R^2}{K^216} \right)\geq \frac{L R^4}{K^32048}\|\vec q\|_2^2\;.
\end{align*}
This completes the proof of \Cref{lem:lower-bound-square}.
\end{proof}
From \Cref{lem:noise-coontribution-ramp}, we have that
$I_1\geq -\sqrt{8 \eps/\boundc^3}(\boundd/L^2) \|\wstar-\vec w\|_{\vec w}$. Using \Cref{lem:lower-bound-square}, for $K=1$, we have that
     \begin{align*}
    I_1+I_2&\geq \frac{\tau^2L R^3 }{4096}\|\wstar-\vec w\|_{\vec w}\left( 
    \|\wstar-\vec w\|_{\vec w}-\sqrt{\eps}\frac{4096\sqrt{8}\boundd}{L^{3}\boundc^{3/2}R^3\tau^2}\right)
    \\&\geq 2(\boundd/\boundc^{3/2})\sqrt{8\eps}\|\wstar-\vec w\|_{\vec w}\geq \sqrt{\eps}\|\wstar-\vec w\|_{\vec w}\;,
\end{align*}
where in the second inequality we used that $\|\wstar-\vec w\|_{\vec w}\geq \sqrt{\eps}\frac{8192\sqrt{8}\boundd}{L^{3}\boundc^{3/2}R^3\tau^2}$ and that $\xi\geq 1,\mu\leq 1$. Moreover, note that in the special case that $\|\vec w\|_2\geq 2\|\wstar\|_2$, we have that $\|\wstar-\vec w\|_{\vec w}\geq (1/2)\|\vec w\|_2^{-1/2}$. Therefore, it holds
     \begin{align*}
    I_1+I_2&\geq \frac{\tau^2L R^3 }{4096}\|\wstar-\vec w\|_{\vec w}\left( 
    \|\wstar-\vec w\|_{\vec w}-\sqrt{\eps}\frac{4096\sqrt{8}\boundd}{L^{3}\boundc^{3/2}R^3\tau^2}\right)
    \\&\geq \frac{\tau^2L R^3 }{4096}\|\wstar-\vec w\|_{\vec w}\left( 
    (1/2)\|\vec w\|_2^{-1/2}-\sqrt{\eps}\frac{4096\sqrt{8}\boundd}{L^{3}\boundc^{3/2}R^3\tau^2}\right)
    \\&\geq \frac{\tau^2L R^3 }{16384}\frac{\|\wstar-\vec w\|_{\vec w}}{\|\vec w\|_2^{1/2}}\;,
\end{align*}
where in the third inequality we used that $\eps\|\vec w\|_2\boundd^2/(L^{6}R^6\boundc^3\tau^2)$ is less than a sufficient small constant. Using that $\|\vec w\|_2\leq C L^6 R^6\boundc^3\tau^4/(\eps\boundd^2)$ for a sufficiently small constant $C>0$, from the assumptions of \Cref{prop:gradient-bound-ramp}, we get that  $I_1+I_2\geq  c'\frac{\boundd}{\boundc^{3/2}}\sqrt{\eps}\|\wstar-\vec w\|_{\vec w}\geq c'\sqrt{\eps}\|\wstar-\vec w\|_{\vec w}$, where we used that $\xi\geq 1,1\geq \mu$ by assumption.

Moreover, for the case that $\|\vec w\|_2\leq 2/R$, we have
$I_1\geq -\sqrt{\eps}\boundd\| \wstar-\vec w\|_2$. Using \Cref{lem:lower-bound-square}, we have that
     \begin{align*}
    I_1+I_2&\geq \frac{\tau^2L R^4}{2048}\|\wstar-\vec w\|_2\left( \|\wstar-\vec w\|_2-\sqrt{\eps}\frac{2048\boundd}{LR^4\tau^2}\right)
   \\&\geq \sqrt{\eps}\|\wstar-\vec w\|_2 \;,
\end{align*}
where in the last inequality we used that $\|\wstar-\vec w\|_2\geq \sqrt{\eps}\frac{4096\boundd}{LR^4\tau^2}$ and that $\xi\geq 1$.

\paragraph{Case 2: $\vec w$ is far from the target $\wstar$}
We now handle the case where our current guess $\vec w$ is far from the target weight vector $\vec
v$.  In particular, we assume that either $\|\proj_{\vec w^\perp}\wstar\|_2 \geq 2K/R$ for some $K\geq 1$, or
$\|\proj_{\vec w}\wstar\|_2\geq 2\|\vec w\|_2$.  In this case there is a large enough region for which we can substitute the $r(\wstar\cdot \x)$ by the constant function $r(\wstar\cdot \x)=1$ (left
plot in \Cref{fig:contribution_large-angle}).
 First, we handle the case where $\|\proj_{\vec w^\perp}\wstar\|_2 \geq 2K/R$ or $\|\proj_{\vec w}\wstar\|_2\geq 2\|\vec w\|_2$. We prove the following:
\begin{lemma}\label{lem:lower-bound-constant-ramp}
Let $\D$ be an $(L,R)$-well-behaved distribution. Let $\theta=\theta(\wstar,\vec w)\in(0,\pi/2)$. There exists a sufficiently small universal constant $c' > 0$ such that for any $\vec w\in \R^d$, if $\|\proj_{\vec w^\perp}\wstar\|_2 \geq 2K/R$, for $K\geq 1$, or $\|\proj_{\vec w}\wstar\|_2\geq 2\|\vec w\|_2$, it holds:
\begin{itemize}
    \item If $\|\vec w\|_2\geq 2/R$, then
 $I_2\geq \frac{\tau^2L R^2 }{72\|\vec w\|_2^{1/2}}\|\vec w-\wstar\|_{\vec w}$.
\item Otherwise, $
 I_2\geq \frac{\tau^2L R^3 }{144 }\|\vec w-\wstar\|_2$.
\end{itemize}
\end{lemma}
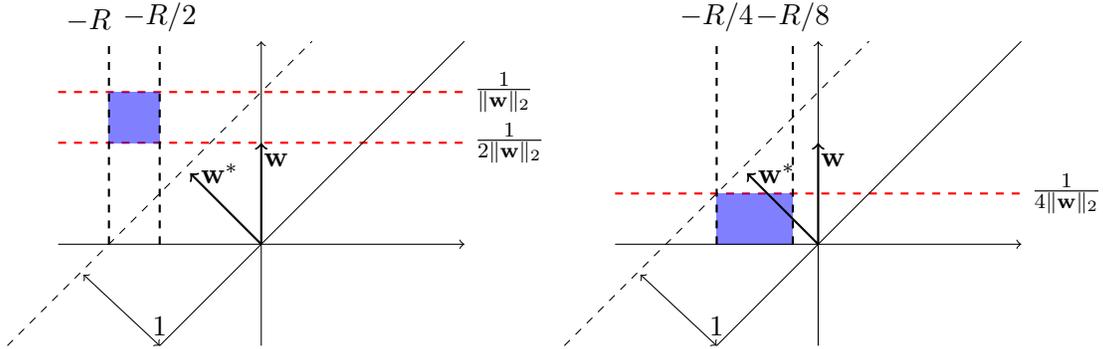
\begin{figure}[h]
    \centering
    \begin{tikzpicture}[scale=1.35]
      \coordinate (start) at (0.5,0);
      \coordinate (center) at (0,0);
      \coordinate (end) at (0.5,0.5);
      \draw (-2,1) node[left] {};
      \draw (-2,0.5) node[left] {};
      \draw (2,2) node[above] {};
      \draw (2,1.2) node[above] {};
      \draw[black,dashed, thick,red](-2,1.5) -- (2,1.5);
      \draw[black,dashed, thick, red](-2,1) -- (2,1);
      \draw[fill=blue, opacity=0.5,draw=none] (-1,1) -- (-1 ,1.5)--(-1.5,1.5)--(-1.5,1);
\draw[black,dashed, thick](-1,0) -- (-1,2);
      \draw[black,dashed, thick](-1.5,0) -- (-1.5,2);
      \draw[->] (-2,0) -- (2,0) node[anchor=north west,black] {};
      \draw[->] (0,-1) -- (0,2) node[anchor=south east] {};
      \draw[thick,->] (0,0) -- (-0.7,0.7) node[anchor= south east,below,right] {$\wstar$};
      \draw[black] (-1,-1) -- (2,2);
            \draw[black,dashed] (-2.5,-1) -- (0.5,2);
                  \draw[<->] (-1.75,-0.3) -- (-1,-1) node[anchor=south east,above] {$1$};
\draw[thick ,->] (0,0) -- (0,1) node[right=2mm,below] {$\bw$};
\draw (2,1) node[right] {$\frac{1}{2\|\vec w\|_2}$};
      \draw (2,1.5) node[right] {$\frac{1}{\|\vec w\|_2}$};
     \draw (-1,2) node[above] {$-R/2$};
      \draw (-1.7,2) node[above] {$-R$};
     
    \end{tikzpicture}
     \begin{tikzpicture}[scale=1.35]
      \coordinate (start) at (0.5,0);
      \coordinate (center) at (0,0);
      \coordinate (end) at (0.5,0.5);
      \draw (-2,1) node[left] {};
      \draw (-2,0.5) node[left] {};
      \draw (2,2) node[above] {};
      \draw (2,1.2) node[above] {};
\draw[black,dashed, thick, red](-2,0.5) -- (2,0.5);
      \draw[fill=blue, opacity=0.5,draw=none] (-1,0) -- (-1 ,0.5)--(-0.25,0.5)--(-0.25,0);
\draw[black,dashed, thick](-1,0) -- (-1,2);
      \draw[black,dashed, thick](-0.25,0) -- (-0.25,2);
      \draw[->] (-2,0) -- (2,0) node[anchor=north west,black] {};
      \draw[->] (0,-1) -- (0,2) node[anchor=south east] {};
      \draw[thick,->] (0,0) -- (-0.7,0.7) node[anchor= south east,below,right] {$\wstar$};
      \draw[black] (-1,-1) -- (2,2);
            \draw[black,dashed] (-2.5,-1) -- (0.5,2);
                  \draw[<->] (-1.75,-0.3) -- (-1,-1) node[anchor=south east,above] {$1$};
\draw[thick ,->] (0,0) -- (0,1) node[right=2mm,below] {$\bw$};
\draw (2,0.5) node[right] {$\frac{1}{4\|\vec w\|_2}$};
     \draw (-0.25,2) node[above] {$-R/8$};
      \draw (-1,2) node[above] {$-R/4$};
     
    \end{tikzpicture}
    \caption{Using our distributional assumptions, there exists a region (``blue'') that provides enough contribution to the gradient, the left plot corresponds to \Cref{lem:lower-bound-constant-ramp} where the angle between the current vector and the target one is large, in this case we take a region where $r(\wstar\cdot \x)=1$, the right plot corresponds to \Cref{lem:lower-bound-square}, in this case the angle is small, and we take a region where $r(\wstar\cdot \x)=\wstar\cdot\x$.}
    \label{fig:contribution_large-angle}
\end{figure}

\begin{proof}
From \Cref{clm:sig2ramp}, we have that
\begin{align*}
  I_2\geq \tau^2\E_{\x\sim \D_\x}[(r(\vec w\cdot \x)-r(\wstar\cdot\x))&r'(\vec w\cdot \x) (\vec w-\wstar)\cdot \x]\\
  &= \tau^2\E_{\x\sim \D_\x}[(\vec w\cdot \x-r(\wstar\cdot\x)) (\vec w-\wstar)\cdot \x ~ \1\{|\vec w\cdot \x| \leq 1\}] \\
&\geq  \tau^2\E_{\x\sim \D_\x}[(\vec w\cdot \x-r(\wstar\cdot\x)) (\vec w-\wstar)\cdot \x~  \1\{0\leq \vec w\cdot \x \leq 1\}]\;.  
\end{align*}
Note that the last inequality holds because $r$ is a non-decreasing function, and thus we have that
$(r(\vec w\cdot \x)-r(\wstar\cdot\x)) (\vec w\cdot \x-\wstar\cdot \x) \geq 0$ for all $\x\in \R^d$. Let $c=\min(1/\|\vec w\|_2,R)$, and note that the condition $  \|\proj_{\vec w}\wstar\|_2\geq 2\|\vec w\|_2$ is equivalent to $2\|\vec w\|\leq \cos\theta \|\wstar\|_2$.
We consider the following subset $-R\leq \x_1\leq -R/2$ and $c/2\leq \x_2\leq c$ which is chosen such $\wstar\cdot\x\geq 1$; which holds because
$\wstar\cdot \x \geq \|\wstar\|_2(\cos\theta \x_2 +R\sin\theta/2)\geq \|\wstar\|_2( \cos\theta c/2+ R\sin\theta/2)$ and the last part is always greater than $1$ if at least one of the following hold:
$\|\wstar\|_2\sin\theta \geq K/R$ or $2\|\vec w\|\leq \cos\theta \|\wstar\|_2$. Therefore, it holds that $r(\wstar\cdot \x)=1$. Hence, we can write
\begin{align*}
    \E_{\x\sim \D_\x}[(&\vec w\cdot \x-r(\wstar\cdot\x)) (\vec w-\wstar)\cdot \x ~ \1\{0\leq \vec w\cdot \x \leq 1\}]\\&\geq
       \E_{\x\sim \D_\x}
       \left[(\vec w\cdot \x-r(\wstar\cdot\x)) (\vec w-\wstar)\cdot \x ~ \1\{c/2\leq \x_2 \leq c, -R\leq \x_1\leq -R/2\}\right]
       \\&\geq \E_{\x\sim \D_\x}\left[(\vec w\cdot \x-1) (\vec w-\wstar)\cdot \x ~ 
       \1\{c/2\leq \x_2 \leq 2c/3, -R\leq \x_1\leq -R/2\} \right]\;.
\end{align*}

Notice that in this case $\vec w\cdot \x\leq \wstar\cdot \x$ and that $(1-\vec w\cdot \x)\geq 1/3$. Therefore,
\begin{align*}
    \E_{\x\sim \D_\x}[(&\vec w\cdot \x-r(\wstar\cdot\x)) (\vec w-\wstar)\cdot \x ~ \1\{0\leq \vec w\cdot \x \leq 1\}]
       \\&\geq (1/3)\E_{\x\sim \D_\x}\left[ |(\vec w-\wstar)\cdot \x| 
       \1\{c/2\leq \x_2 \leq 2c/3, -R\leq \x_1\leq -R/2\} \right]\;.
\end{align*}
Let $\vec q=\vec w-\wstar$. By using the inequality $(a^2+b^2)^{1/2}\geq (1/2) (|a|+|b|)$ for any $a,b\in \R$ and that $(\vec q\cdot \x)^2=(\vec q^{\|_{\vec w}}\cdot \x)^2+(\vec q^{\perp_{\vec w}}\cdot \x)^2$, we have that
\begin{align*}
    \E_{\x\sim \D_\x}[ |(\vec w-\wstar)\cdot &\x| 
       \1\{c/2\leq \x_2 \leq 2c/3, -R\leq \x_1\leq -R/2\} ]
       \\&=\E_{\x\sim \D_\x}[ \sqrt{(\vec q\cdot \x)^2} 
       \1\{c/2\leq \x_2 \leq 2c/3, -R\leq \x_1\leq -R/2\} ]
       \\&\geq \frac 12 \E_{\x\sim \D_\x}[(|\vec q^{\|_{\vec w}}\cdot \x|+|\vec q^{\perp_{\vec w}}\cdot \x|)
       \1\{c/2\leq \x_2 \leq 2c/3, -R\leq \x_1\leq -R/2\} ]
           \\&\geq \frac 12 \left(\|\vec q^{\|_{\vec w}}\|_2\frac{c}{2}+\|\vec q^{\perp_{\vec w}}\|_2\frac{R}{2}\right)\E_{\x\sim \D_\x}[
       \1\{c/2\leq \x_2 \leq 2c/3, -R\leq \x_1\leq -R/2\} ]
       \\&\geq \frac{c L R }{24} \left(\|\vec q^{\|_{\vec w}}\|_2\frac{c}{2}+\|\vec q^{\perp_{\vec w}}\|_2\frac{R}{2}\right)\;,
\end{align*}
where we used that the distribution is $(L,R)$-well-behaved.
For the case that $c=1/\|\vec w\|_2$, we have that 
\[
I_2\geq \frac{\tau^2 L R}{72 \|\vec w\|_2}  \left(\|\vec q^{\|_{\vec w}}\|_2\frac{1}{2\|\vec w\|_2}+\|\vec q^{\perp_{\vec w}}\|_2\frac{R}{2}\right)\geq  \frac{\tau^2 L R^2}{72 \|\vec w\|_2^{1/2}} \|\vec q\|_{\vec w}\;.
\]
Finally, for the case that $c=R$, we have that 
\[
I_2\geq \frac{\tau^2 L R^2}{72}  \left(\|\vec q^{\|_{\vec w}}\|_2\frac{R}{2}+\|\vec q^{\perp_{\vec w}}\|_2\frac{R}{2}\right)\geq  \frac{\tau^2 L R^3}{144 } \|\vec q\|_{2}\;,
\]
where we used that $(a+b)^2\geq a^2+b^2$ for $a,b\geq 0$.
This completes the proof of \Cref{lem:lower-bound-constant-ramp}.
\end{proof}
To prove \Cref{prop:gradient-bound-ramp} for this case (the case for which the assumptions of \Cref{lem:lower-bound-constant-ramp} hold), we use the bound of the contribution of the noise to the gradient from \Cref{lem:noise-coontribution-ramp}, i.e., that
$I_1\geq -\sqrt{8 \eps/\boundc^3}(\boundd/L^2) \|\wstar-\vec w\|_{\vec w}$, given that $\|\vec w\|_2\geq 2/R$.
     Putting together our estimates for $I_1$ and $I_2$, we obtain:
     \begin{align}
    I_1+I_2&\geq \frac{\tau^2L R^2}{72\|\vec w\|_2^{1/2}} \|\vec q\|_{\vec w}\left(1-\sqrt{\eps\|\vec w\|_2}\frac{\boundd 72\sqrt{8}}{L^{3}R^2\boundc^{3/2}\tau^2}\right)\nonumber
    \\&\geq \frac{\tau^2L R^2}{144\|\vec w\|_2^{1/2}}\|\vec q\|_{\vec w}\label{eq:lower-bound-large-angle}\;,
\end{align}
where, we used that $\sqrt{\eps\|\vec w\|_2}\boundd^2/(L^{3}R^2\boundc^{3/2}\tau^2)$ is 
less than a sufficiently small constant from the assumptions of \Cref{prop:gradient-bound-ramp}.  
Observe that from the assumptions of \Cref{prop:gradient-bound-ramp}, there exists a constant $C>0$,such that $\| \vec w\|_2\leq C L^6 R^6\boundc^3\tau^4/(\eps\boundd^2)$ hence, we have that
\[
I_1+I_2\geq \sqrt{\eps}\|\vec q\|_{\vec w}\;,
\]
where we used that $L,R,\mu\leq 1$ and $\xi\geq 1$.

Finally, similar to the previous case, for the case that $\|\vec w\|_2\leq 2/R$, it holds 
\begin{align*}
    I_1+I_2&\geq \frac{\tau^2L R^3}{144} \|\vec q\|_2\left(1-\sqrt{\eps}\frac{144\boundd}{L R^3\tau^2}\right)\geq \frac{\tau^2L R^3}{288}\|\vec q\|_2 \;,
\end{align*}
where we used that $(1-\sqrt{\eps}\frac{144\boundd}{L R^3\tau^2})\geq 1/2$.

\paragraph{Case 3: angle of $\vec w$ and $\wstar$ is greater than $\pi/2$}
We now handle the case where $\theta(\vec w,\wstar)\in(\pi/2,\pi)$. This case is very similar to the second case, i.e., \Cref{lem:lower-bound-constant-ramp}.
\begin{lemma}\label{clm:lower-bound-negative angle}
Let $\D$ be an $(L,R)$-well-behaved distribution. For any vector $\vec w\in \R^d$ with $\theta(\wstar,\vec w)\in(\pi/2,\pi)$. We show
\begin{itemize}
    \item If $\|\vec w\|_2\geq 2/R$, then
 $I_2\geq \frac{\tau^2L R^2 }{16\|\vec w\|_2^{1/2}}\|\vec w-\wstar\|_{\vec w}$.
\item Otherwise, we have that $
 I_2\geq \frac{\tau^2L R^4 }{16 }\|\vec w-\wstar\|_2$.
\end{itemize}
\end{lemma}
\begin{proof}
Let $c=\min(1/\|\vec w\|_2,R)$.  We consider the $R/2\leq \x_1\leq R$ and $c/2\leq \x_2\leq c$ so that $0\geq \wstar\cdot\x$. Similarly to the previous case, using \Cref{clm:sig2ramp}, we get that $I_2\geq \tau^2  \E_{\x\sim \D_\x}[(\vec w\cdot \x-r(\wstar\cdot\x)) (\vec w-\wstar)\cdot \x ~ \1\{0\leq \vec w\cdot \x \leq 1\}]$ and for this case it holds that $r(\wstar\cdot \x)\leq 0$. Hence, we have that
\begin{align*}
    \E_{\x\sim \D_\x}[(&\vec w\cdot \x-r(\wstar\cdot\x)) (\vec w-\wstar)\cdot \x ~ \1\{0\leq \vec w\cdot \x \leq 1\}]\\&\geq
       \E_{\x\sim \D_\x}
       \left[(\vec w\cdot \x-r(\wstar\cdot\x)) (\vec w-\wstar)\cdot \x ~ \1\{c/2\leq \x_2 \leq c, R/2\leq \x_1\leq R\}\right]
       \\&\geq  \frac{c\|\vec w\|_2}{2}\E_{\x\sim \D_\x}\left[(\vec w-\wstar)\cdot \x ~ 
       \1\{c/2\leq \x_2 \leq c, R/2\leq \x_1\leq R\} \right]\;.
\end{align*}
Note that $(\vec w-\wstar)\cdot \x=(\|\vec w\|_2-\cos\theta \|\wstar\|_2)\x_2 + \sin\theta\|\wstar\|_2 \x_1\geq (\|\vec w\|_2-\cos\theta \|\wstar\|_2)c/2 + \sin\theta\|\wstar\|_2 R/2$, when $R/2\leq \x_1\leq R$ and $c/2\leq \x_2\leq c$. Denote by $\vec q=\vec w-\wstar$ we get that $(\|\vec w\|_2-\cos\theta \|\wstar\|_2)c/2 + \sin\theta\|\wstar\|_2 R/2= \|\vec q^{\|_{\vec w}}\|_2 c/2 +\|\vec q^{\perp_{\vec w}}\|_2 R/2$.
Furthermore, using that the distribution $\D$ is $(L,R)$-well-behaved, we have that 
$    \E_{\x\sim \D_\x}\left[
       \1\{c/2\leq \x_2 \leq c, R/2\leq \x_1\leq R\} \right]\geq L \frac{R c}{4}$. Hence,
       \begin{align*}
    I_2\geq \frac{\tau^2L R c^2\|\vec w\|_2}{16}\left(\|\vec q^{\|_{\vec w}}\|_2 c +\|\vec q^{\perp_{\vec w}}\|_2 R\right) \;,
\end{align*}
which completes the proof of \Cref{clm:lower-bound-negative angle}.
 \end{proof}
 Therefore, if $\|\vec w\|_2\geq 2/R$, we have
$I_1\geq -\sqrt{8 \eps/\boundc^3}(\boundd/L^2) \|\wstar-\vec w\|_{\vec w}$.
     Putting together our estimates for $I_1$ and $I_2$, we obtain:
     \begin{align*}
    I_1+I_2&\geq \frac{\tau^2L R^2}{16\|\vec w\|_2^{1/2}} \|\vec w-\wstar\|_{\vec w}\left(1-\sqrt{\eps\|\vec w\|_2}\frac{\boundd 16}{L^{3}R^2\boundc^{3/2}\tau^2}\right)
    \\&\geq \frac{\tau^2L R^2}{32\|\vec w\|_2^{1/2}}\|\vec w-\wstar\|_{\vec w}\;,
\end{align*}
where we used that $\sqrt{\eps\|\vec w\|_2}\boundd/(L^{3}R^2\boundc^{3/2}\tau^2)$ is 
less than a sufficiently small constant from the assumptions of \Cref{prop:gradient-bound-ramp}.  
Note that from the assumptions of \Cref{prop:gradient-bound-ramp}, there exists a constant $C>0$ such that $\| \vec w\|_2\leq C L^6 R^6\boundc^3\tau^4/(\eps\boundd^2)$, and hence it holds
\[
I_1+I_2\geq \sqrt{\eps}\|\vec q\|_{\vec w}\;,
\]
where we used that $L,R,\mu\leq 1$ and $\xi\geq 1$.

Finally, similarly to the previous case, we have that if $\|\vec w\|_2\leq 2/R$, it holds that
\begin{align*}
    I_1+I_2&\geq \frac{\tau^2L R^4}{16} \|\vec w-\wstar\|_2\left(1-\sqrt{\eps}\frac{16\boundd}{L R^4\tau^2}\right) \geq \frac{L R^4\tau^2}{32}\|\vec w-\wstar\|_2 \;,
\end{align*}
where we used that $(1-\sqrt{\eps}\frac{16\boundd}{L R^4\tau^2})\geq 1/2$. 
This completes the proof of \Cref{prop:gradient-bound-ramp}.
\end{proof}

\subsection{Proof of \Cref{clm:constant-contri-reg-1}}
We restate and prove the following claim.
\begin{claim}
If $c'\kappa/(\eps\Lambda_2)\leq\|\vec w\|_2 \leq c'\kappa/\eps$ and $\| \proj_{\vec w^\perp} \wstar \|_2\geq 2K/R$ then 
    \begin{align*}
   \nabla F_\rho(\vec w)\cdot(\vec w-\wstar)\geq (c'/2) \sqrt{\eps} \|\wstar-\vec w\|_{\vec w}
   \,.
\end{align*}

\end{claim}
\begin{proof}
First, we calculate the contribution of the regularizer in the direction $\vec w-\wstar$. This is equal to $2\rho \|\vec w\|_2^2(\|\vec w\|_2-\|\wstar\|_2\cos\theta)$. Note that this is positive when $\|\vec w\|_2-\|\wstar\|_2\cos\theta\geq 0 $ and negative otherwise. Hence, if $\theta\in[\pi/2,\pi)$ the contribution of the regularizer is positive, and therefore it is bounded below by the contribution of the gradient without the regularizer. Moreover, if $\|\vec w\|_2\geq 2\|\wstar\|_2$, then $\|\vec w\|_2-\|\wstar\|_2\cos\theta\geq  \|\vec w\|_2-\|\wstar\|_2\geq 0$, therefore we can again bound from below the contribution of the gradient like before. For the rest of the proof, we consider the case where $\theta\in(0,\pi/2)$ and $\|\vec w\|_2\leq 2\|\wstar\|_2$.

From \Cref{prop:gradient-bound-ramp} and specifically \Cref{eq:lower-bound-large-angle}, we have that as long as
$2/R\leq \| \vec w\|_2\leq c'\kappa/(\eps$ and $\| \proj_{\vec w^\perp} \wstar \|_2\geq 2K/R$ then $ \nabla F(\vec w)\cdot(\vec w-\wstar)\geq c'\frac{\tau^2L R^2}{\|\vec w\|_2^{1/2}}\|\vec w-\wstar\|_{\vec w}$. Therefore, we have that
\begin{align*}
   \nabla F_\rho(\vec w)\cdot(\vec w-\wstar) &\geq c'\tau^2 L R^2 \frac{\|\wstar-\vec w\|_{\vec w}}{\|\vec w\|_2^{1/2}} + \rho \|\vec w\|_2(\|\vec w\|_2-\|\wstar\|_2\cos\theta)
   \\ &\geq c' \tau^2 L R^2 \frac{\|\wstar-\vec w\|_{\vec w}}{\|\vec w\|_2^{1/2}} - \rho \|\vec w\|_2\|\wstar\|_2
   \\&\geq c'\tau^2 LR^2\frac{\|(\vec w-\wstar)^{\|_{\vec w}}\|_2}{\|\vec w\|_2^2} +c' \tau^2 L R^2 \frac{1}{\|\vec w\|_2}\left(\|\proj_{\vec w^\perp} \wstar\|_2-\frac{\rho \|\vec w\|_2^2\|\wstar\|_2}{c'\tau^2 L R^2}\right)
   \,.
\end{align*}
To bound the $\|\proj_{\vec w^\perp} \wstar\|_2-\rho \|\vec w\|_2^2\|\wstar\|_2/(c' \tau^2 L R^2)$, we have that
\begin{align*}
    \frac{\|\proj_{\vec w^\perp} \wstar\|_2 c' \tau^2L R^2}{2\rho \|\vec w\|_2^2\|\wstar\|_2}\geq \frac{K c' \tau^2 L R}{\rho \|\vec w\|_2^2\|\wstar\|_2}\geq 
    \frac{K  F^3(\wstar)L R\tau^2}{\rho c'^2 \kappa^2 U}\geq 1\;,
\end{align*}
where the in the first inequality we used that $\| \proj_{\vec w^\perp} \wstar \|_2\geq 2K/R$; in the second that $\|\vec w\|_2\leq c'\kappa/\eps$ and $\|\wstar\|_2 \eps\leq U$; and in the last inequality that  $\rho\leq  \frac{K  \eps^3L R\tau^2}{ c'^2 \kappa^2 U}$. Therefore, we have that
$\left(\|\proj_{\vec w^\perp} \wstar\|_2-2\rho \|\vec w\|_2^2\|\wstar\|_2/(c' L R^2\tau^2)\right)\geq \|\proj_{\vec w^\perp} \wstar\|_2/2$ and the result follows similar to the proof of \Cref{clm:constant-contri-reg-2}.
\end{proof}

\subsection{Proof of \Cref{lem:bound-difference-bounded}}
We restate and prove the following lemma:
\begin{lemma}[Parameter vs $L_2^2$ Distance]
Let $\D_\x$ be an $(L,R)$ well-behaved distribution.  Let $\sigma$ be a $(\tau,\xi,\mu)$-sigmoidal
activation.  For any vectors $\vec w,\vec v\in \R^d$, we have $\Exx[(\sigma(\vec w\cdot
\x)-\sigma(\vec v\cdot \x))^2] \leq \xi^2 \|\vec w-\vec v\|_2^2$.
Moreover, if $\theta=\theta(\vec w,\vec v) < \pi/4$, 
$\|\vec w\|_2\leq \delta\|\vec v\|_2$ and $\delta\geq 1$, and $\|\vec w\|_2 > 2/R$, 
 it holds
\[
 \Exx[(\sigma(\vec w\cdot \x)-\sigma(\vec v\cdot \x))^2] \lesssim \frac{\boundd^2\delta^3}{L^4\boundc^3} \|\vec w-\vec v\|_{\vec w}^2  \,.
 \]
\end{lemma}

\begin{proof}
First, we consider the case where $\|\vec w\|_2\geq 2/R$ and $\theta\in(0,\pi/4)$. Let $V$ be the subspace spanned by $\vec w,\vec v$ and assume without loss of generality that $\vec w/\|\vec w\|_2=\vec e_2$ and therefore $\vec v=\|\vec v\|_2(\cos\theta \vec e_2 - \sin\theta \vec e_1)$, we abuse the notation of $\D_\x$ to be the distribution projected on $V$. It holds that
\begin{align*}
    \E_{\x\sim \D_\x}[(\sigma(\vec w\cdot \x)-\sigma(\vec v\cdot\x))^2]
    &= \E_{\x\sim \D_\x}\left[\left(\int_{\vec v\cdot\x}^{\vec w\cdot\x}\sigma'(t)\d t\right)^2\right]
    \\  &\leq \E_{\x\sim \D_\x}\left[\left(\int_{\vec v\cdot\x}^{\vec w\cdot\x}\boundd\exp(-\boundc |t|)\d t\right)^2\right]
    \\&\leq  \boundd^2\E_{\x\sim \D_\x}[(\vec w\cdot \x-\vec v\cdot\x)^2 (\exp(-2\boundc |\vec w\cdot \x|)+\exp(-2\boundc |\vec v\cdot \x|))]\;,
\end{align*}
where we used that $\sigma$ is non-decreasing. Let $\vec q=\vec w-\vec v$.
Using  \Cref{clm:bound-square-conce}, it holds that
\begin{align*}
    \E_{\x\sim \D_\x}[(\vec w\cdot \x-\vec v\cdot\x)^2 \exp(-2\boundc |\vec w\cdot \x)|)] &= \E[ ((\vpar{q}{w} \cdot \x)^2 + (\vperp{q}{w}\cdot\x)^2) \exp(-2\vec w\cdot \x \boundc) ]
     \\&\leq     \frac{8}{L^4}( 
     \|\vec q^{\|_{\vec w}}\|_2^2 \frac{1}{\boundc^3\|\vec w\|_2^3}
     + 
     \|\vec q^{\perp_{\vec w}}\|_2^2 \frac{1}{\boundc\|\vec w\|_2})\leq \frac{8\|\vec q\|_{\vec w}^2}{L^4\boundc^3}\;.
\end{align*}
Moreover, for the second term, it holds that 
\begin{align*}
    \E_{\x\sim \D_\x}[(\vec w\cdot \x-\vec v\cdot\x)^2 \exp(-2\boundc |\vec v\cdot \x)|)] &= \Exx[ (\vpar{q}{w} \cdot \x)^2 + (\vperp{q}{w}\cdot\x)^2) \exp(-2\vec v\cdot \x \boundc) ]
    \\ &\leq  \Exx[ (\vpar{q}{w} \cdot \x)^2 + (\vperp{q}{w}\cdot\x)^2) \exp(-2\|\vec v\|_2\cos\theta\x_2\boundc) ]
     \\&\leq     \frac{8}{L^4}( 
     \|\vec q^{\|_{\vec w}}\|_2^2 \frac{1}{\boundc^3\|\vec v\|_2^3\cos\theta^3}
     + 
     \|\vec q^{\perp_{\vec w}}\|_2^2 \frac{1}{\boundc\|\vec v\|_2\cos\theta})\lesssim \frac{\delta^3\|\vec q\|_{\vec w}^2}{L^4\boundc^3}\;,
\end{align*}
where in the last inequality, we used that $\cos\theta\geq 1/2$ and that $\|\vec w\|_2\leq \delta \|\vec v\|_2$.
For the other cases, we use the ``trivial'' upper-bound, i.e.,
\begin{align*}
   \Exx[(\sigma(\vec w\cdot \x)-\sigma(\vec v\cdot \x))^2]&\leq \boundd^2\Exx[(\vec w\cdot \x-\vec v\cdot \x)^2]
   \\&\leq \boundd^2\|\vec w-\vec v\|_2^2\sup_{\|\vec w\|_2=1}\Exx[(\vec w\cdot \x)^2]
   \\&\leq \boundd^2\|\vec w-\vec v\|_2^2\;, 
\end{align*}
where we used that $\sigma$ is $\boundd$-Lipschitz and that the distribution $\D_\x$ is isotropic.
\end{proof}

\section{Omitted Proofs from \Cref{sec:optimization}}
\label{app:optimization}

\subsection{Proof of \Cref{clm:estimation-gradient}}
We restate and prove the following claim.
\begin{claim}
Let $\vec w$ be a vector with $\|\vec w-\wstar\| \geq \sqrt{\eps}$.
Using $N =\wt{O}(d\lambda^4\|\vec w-\wstar\|_2^2W^2\max(\log^2(1/\eps),1)/(L\eps')^2)$ samples, we can compute an empirical 
gradient $\vec g(\vec w)$ such that $\| \vec g(\vec w) - \nabla F(\vec w) \|_2 \leq \eps$ with probability $1-\delta$.
\end{claim}
\begin{proof}
We start by bounding from above the variance in every direction. Let $\Sigma_i=\E[((\sigma(\vec w \cdot \x) -y)\sigma'(\vec w\cdot \x)\x_i)^2]$, we have that
\begin{align*}
\Sigma_i=\E[((&\sigma(\vec w \cdot \x) -y)\sigma'(\vec w\cdot \x)\x_j)^2]
\leq 
\lambda^2  \E[(\sigma(\vec w \cdot \x) -y)^2 \x_j^2]
\\
&\leq 2 \lambda^2\E[(\sigma(\vec w^\ast \cdot \x) - y)^2 \x_j^2] 
 + 2 \lambda^2\E[(\sigma(\vec w^\ast \cdot \x) - \sigma(\vec w \cdot \x) )^2 
 \x_j^2]\;.
\\
&\leq 2 
\lambda^2\E[(\sigma(\vec w^\ast \cdot \x) - y)^2 \x_j^2]
 + 2 \lambda^4
 \|\vec w^\ast - \vec w\|_2^2
 \max_{\|\vec u \|_{2} = 1}
 \E[ (\vec u \cdot \x)^2 \x_j^2]\;.
\end{align*}
To bound the term $\E[(\sigma(\vec w^\ast \cdot \x) - y)^2 \x_j^2]$, we have that
\begin{align*}
    \E[(\sigma(\vec w^\ast \cdot \x) - y)^2 \x_j^2]&=\E[(\sigma(\vec w^\ast \cdot \x) - y)^2 \x_j^2\1\{\x_j\leq M\}]+\E[(\sigma(\vec w^\ast \cdot \x) - y)^2 \x_j^2\1\{\x_j\geq M\}]
    \\&\leq {M}^2\E[(\sigma(\vec w^\ast \cdot \x) - y)^2 ]+4M^2\E[ \x_j^2\1\{\x_j\geq M\}]
    \\&\lesssim {M}^2\eps +4M^2\E[ \x_j^2\1\{\x_j\geq M'\}]\lesssim \eps M^2\;.
\end{align*}
Moreover, note that $\sqrt{\eps}\leq \|\vec w-\wstar\|_2$, hence 
$\Sigma_i\lesssim \lambda^4\|\vec w-\wstar\|_2^2M^2$.
From Markov's inequality, we have that for each $j\leq d$, it holds
\begin{align*}
\pr&\left[\left 
|\sum_{i=1}^N \frac{1}{N}(\sigma(\vec w \cdot \x\ith) -y\ith)\sigma'(\vec w\cdot \x\ith)\x_j\ith 
-\E[(\sigma(\vec w \cdot \x) -y)\sigma'(\vec w\cdot \x)\x_j]\right|
\geq \eps'/\sqrt{d}\right]
\\
&\leq \frac{d}{N \eps'^2}\E[((\sigma(\vec w \cdot \x) -y)\sigma'(\vec w\cdot \x)\x_j)^2]\leq  \frac{d\|\vec w-\wstar\|_2^2\lambda^4M^2}{N \eps'^2}\;. 
\end{align*}
Hence, with $N=O(\frac{d\lambda^4\|\vec w-\wstar\|_2^2M^2}{\eps'^2}) $ samples, 
we can get an $\eps'/\sqrt{d}$-approximation to the $i$-th coordinate of the gradient 
$(\nabla  F(\vec w))_i$ with constant probability, and by using  a standard boosting 
procedure we can boost the probability to $1-\delta$ with a multiplicative overhead of $O(\log(1/\delta))$ samples.
Finally, doing a union bound over all coordinates $j \in \{1,\ldots, d\}$ we obtain
that $N=\wt{O}(d\lambda^4\|\vec w-\wstar\|_2^2W^2\max(\log^2(1/\eps),1)/(L^2\eps'^2))$ samples suffice.
\end{proof}

\subsection{Proof of \Cref{lem:gradient-field-bounded}}
We restate and prove the following lemma.
\begin{lemma}[Gradient Field Distance Reduction]
Let $Z_1 > Z_0 \geq 1$. Let $\vec g: \R^d \mapsto \R^d$ be a vector field 
with $\|\vec g(\vec w)\|_2 \leq B$ for every $\vec w \in \R^d$ with $\|\vec w\|_2\leq 2Z_1$.
We assume that  $\vec g$ satisfies the following properties with respect to some unknown 
target vector $\vec v$:
\begin{enumerate}
\item  
If $\|\vec w\|_2 \leq Z_0$
and $\|\vec w - \vec v\|_2 \geq \alpha_1$
then 
it holds that $\vec g(\vec w) \cdot ( \vec w - \vec v) \geq \alpha_2 \|\vec w - \vec v\|_2$,
for $\alpha_1, \alpha_2 > 0$.
\item 
If $ Z_0 < \|\vec w\|_2 \leq Z_1$ 
and $\|\vec w - \vec v\|_{\vec w} \geq \beta_1$
it holds that $\vec g(\vec w) \cdot ( \vec w - \vec v) \geq \beta_2 \|\vec w - \vec v\|_{\vec w}$, 
for $\beta_1, \beta_2 > 0$.
\item 
For some $\zeta \in (0,1), \gamma>0$, we have that if 
$\|\vec w\|_2 \geq \zeta Z_1 $,
it holds that $\vec g(\vec w) \cdot  \vec w \geq \gamma \|\vec w\|_{\vec w}$.
\end{enumerate}
We consider the update rule $\vec w^{(t+1)} \gets \vec w^{(t)} - \eta \vec g(\vec w^{(t)})$
initialized  with $\vec w^{(0)} = \vec 0$
and step size  \[
\eta = \frac{1}{B^2} \min\left(\alpha_1 \alpha_2, \frac{\beta_1 \beta_2}{Z_1^{3/2}}, \frac{2 \gamma}{Z_1}, (1-\zeta) Z_1 B\right)
\]
Let $T$ be any integer larger than $\Big\lceil\frac{\|\vec v\|^2}{\eta  \min(\alpha_1 \alpha_2, \beta_1 \beta_2/Z_1^{3/2})} \Big\rceil $.
We have that $\|\vec w^{(T)}\|_2 \leq Z_1$.
Moreover, if  $\|\vec w^{(T)} \|_2 \leq Z_0$ it holds that 
$\|\vec w^{(T)} - \vec v \|_{2} \leq \eta B + \max(\alpha_1, (2 Z_0)^{3/2} \beta_1)$
and if $\|\vec w^{(T)}\|_{2} > Z_0$ we have 
$\|\vec w^{(T)} - \vec v \|_{\vec w^{(T)}} 
\leq \sqrt{2} \eta B +  \max(\sqrt{2} \alpha_1, e^{3 Z_1^{3/2} \eta B} \beta_1) \,.
$
\end{lemma}
\begin{proof}
We first show by induction that for every $t\geq 0$ it holds that $\|\vec w^{(t)}\|_2 \leq Z_1$.
For $t = 0$ we have $\vec w^{(0)} = \vec 0$ so the claim holds.  Assume first that 
$\|\vec w^{(t)}\|_2 \leq \zeta Z_1$.  Then
\[
\| \vec w^{(t+1)} \|_2 \leq  \| \vec w^{(t)} \|_2 + \eta B \leq  \zeta R_1 + (1-\zeta) Z_1 = Z_1,
\]
where we used the fact that $\eta B \leq (1- \zeta) Z_1$.
Now, if $\zeta Z_1 \leq \|\vec w^{(t)}\|_2 \leq Z_1$ by assumption 3 of the vector field $\vec g$,
we have that 
\begin{align*}
\| \vec w^{(t+1)} \|_2^2 
&= 
\| \vec w^{(t)} \|_2^2 - 2 \eta \vec g(\vec w^{t} ) \cdot \vec w^{(t)} - \eta^2 \|\vec g( \vec w^{(t)} )\|_2^2 
\\
&\leq 
\| \vec w^{(t)} \|_2^2 - 2 \eta \gamma/\sqrt{Z_1} - \eta^2 B^2
\,.
\end{align*}
Since $\eta \leq 2 \gamma /(B^2 \sqrt{Z_1})$, we have that  
$- 2 \eta \gamma/\sqrt{Z_1} - \eta^2 B^2 \leq 0$, and therefore
$ \| \vec w^{(t+1)} \|_2^2 \leq \|\vec w^{(t)}\|_2^2 \leq Z_1^2$.

Now that we have that $\|\vec w^{(t)}\|_2$ is always smaller than $Z_1$, we know that the vector
field $\vec g(\vec w^{(t)})$ has non-trivial component on the direction $\vec w - \vec v$, i.e.,
either condition 1 or condition 2 is true.
We next show that when condition 1 or 2 of the \Cref{lem:gradient-field-bounded} hold, we can improve the distance 
of $\vec w^{(t)}$ and the target $\vec v$.
We first assume that  $\|\vec w^{(t)}\|_2 \leq Z_0$ and that 
$\|\vec w^{(t)} - \vec v\|_2 \geq \alpha_1$.
Then by assumption 1 of the lemma we have that it holds 
$\vec g(\vec w\tth) \cdot ( \vec w\tth - \vec v) \geq \alpha_2 \|\vec w\tth - \vec v\|_2$.
We have 
\begin{align*}
\|\vec w\tth-\vec v\|_2^2 -\|\vec w^{(t+1)}- \vec v \|_2^2
&=  2\eta \vec g(\vec w\tth)  \cdot(\vec w\tth-\vec v) -\eta^2\|\vec g(\vec w\tth) \|_2^2
\\
&\geq  2\eta \alpha_1 \alpha_2 - \eta^2\|\vec g(\vec w\tth) \|_2^2
\\
&\geq  \eta \alpha_1 \alpha_2 \,,
\end{align*}
where for the last inequality we used the fact that $ \|\vec g(\vec w\tth) \|_2^2 \leq B^2$
and that  $\eta \leq \alpha_1 \alpha_2/B^2$.
On the other hand, if $\|\vec w^{(t)}\|_2 \leq Z_0$ and 
$\|\vec w^{(t)} - \vec v\|_2 \geq \beta_1$, then by assumption 2 of the lemma we have that it holds 
$\vec g(\vec w\tth) \cdot ( \vec w\tth - \vec v) \geq \beta_2 \|\vec w - \vec v\|_{\vec w\tth}$.
Similarly to the previous case, we then have 
\begin{align*}
\|\vec w\tth-\vec v\|_2^2 -\|\vec w^{(t+1)}- \vec v \|_2^2
&\geq 2 \eta \beta_2 \|\vec w\tth - \vec v\|_{\vec w\tth} - \eta^2\|\vec g(\vec w\tth) \|_2^2\;.
\end{align*}
We next bound from below the norm $\|\cdot\|_{\vec w}$ by the $\ell_2$ norm. The following rough estimate
suffices.
\begin{claim}
\label{clm:ell2ellw}
For every $\x \in \R^d$ it holds 
\[
\min\Big(\frac{1}{\|\vec w\|_2^{3/2}}, \frac{1}{\| \vec w \|_2^{1/2}} \Big) 
\|\vec x\|_2  \leq  \|\vec x\|_{\vec w}  \leq  \sqrt{2} 
\max\Big(\frac{1}{\|\vec w\|_2^{3/2}}, \frac{1}{\| \vec w \|_2^{1/2}} \Big)  \|\vec x\|_2 \,.
\]
\end{claim}
\begin{proof}
The claim follows directly from the definition of the norm $\| \cdot \|_{\vec w}$ 
and the inequality $ \sqrt{a^2 + b^2} \leq a + b \leq \sqrt{2} \sqrt{a^2 + b^2}$ that holds for all 
$a,b \geq 0$.
\end{proof}
Using the above bounds and the fact that  $Z_1 \geq 1$, we conclude that  
$\|\vec w\tth-\vec v\|_2^2 -\|\vec w^{(t+1)}- \vec v \|_2^2 \geq \eta \beta_1 \beta_2/Z_1^{3/2} $,
where we used the assumption that $\eta \leq \beta_1 \beta_2/(Z_1^{3/2} B^2)$.
Recall that we have set $\eta = 1/B^2 \min(\alpha_1 \alpha_2, \beta_1 \beta_2/Z_1^{3/2}, 2 \gamma/Z_1, (1-\zeta) Z_1 B)$
which implies that at every iteration where either 
$\|\vec w\tth\|_2 \leq Z_0 $ and 
$\|\vec w^{(t)} - \vec v\|_2 \geq \alpha_1$, or 
$\|\vec w\tth\|_2 > Z_0$ and 
$\|\vec w^{(t)} - \vec v\|_{\vec w\tth} \geq \alpha_1$ we have that 
$\|\vec w\tth-\vec v\|_2^2 -\|\vec w^{(t+1)}- \vec v \|_2^2 \geq \eta \min(\alpha_1 \alpha_2, \beta_1 \beta_2/Z_1^{3/2})$.
Let us denote by $T$ the first iteration such that the distance reduction stops happening.
Since we initialize at $\vec w^{(0)} = \vec 0$ the first time that either of the above condition stops being true, namely, $T$ 
can be at most $\Big\lceil\frac{\|\vec v\|^2}{\eta  \min(\alpha_1 \alpha_2, \beta_1 \beta_2/Z_1^{3/2})} \Big\rceil $.

After either $\|\vec w^{(T)} - \vec v\|_{2} \leq \alpha_1$ or 
$ \|\vec w^{(T)} - \vec v\|_{\vec w^{(T)}} \leq \beta_1$ we stop having the guarantee
that updating the guess with $\vec g(\vec w^{(T)})$ will decrease the distance, and therefore
we may move to the wrong direction.  However, since our step size $\eta$ is small doing 
a small step in the wrong direction cannot increase the distance of $\vec w^{(T+1)}$
and $\vec v$ by a lot.
We next show that even if we continue updating after the iteration $T$, we cannot make 
the distance of $\vec w^{(T)}$ and the target vector $\vec v$ much larger.
We first assume that at iteration $T$ we have $\|\vec w^{(T)} \|_2 \leq Z_0$, and therefore,
it must be the case that $\|\vec w^{(T)} - \vec v\|_2 \leq \alpha_1$.  
Notice that it suffices to show that after one iteration the claim holds, since after
$\|\vec w^{(T+1)}\|_2$ grows larger than $\alpha_1$ the distance of $\vec w $ will start to 
decrease again. For the next iteration, using the bound
$\|\vec g(\vec w^{(T)})\|_2 \leq B$, we obtain that $\|\vec w^{(T+1)} - \vec w^{(T)} \|_2 \leq \eta B$.
Now if $\|\vec w^{(T+1)}\|_2 \leq Z_0$, using the triangle inequality for the $\ell_2$ norm, 
we have that $\|\vec w^{(T+1)} -\vec v \|_2 \leq \alpha_1 + \eta B$.
On the other hand, if  $\|\vec w^{(T+1)}\|_2 > Z_0$, using the triangle inequality of the norm 
$\|\cdot \|_{\vec w}$ we obtain 
\begin{align}
\label{eq:next-step-triangle}
\|\vec w^{(T+1)} - \vec v \|_{\vec w^{(T+1)}}
&\leq 
\|\vec w^{(T+1)} - \vec w^{(T)} \|_{\vec w^{(T+1)}}
+
\|\vec w^{(T)} - \vec v\|_{\vec w^{(T+1)}}
\\
&\leq \sqrt{2} (\eta B + \alpha_1)
\nonumber
\,,
\end{align}
where we used \Cref{clm:ell2ellw}, the fact that $\vec w^{(T+1)} - \vec w^{(T)}
= \eta \vec g(\vec w^{(T)})$ and that $\|\vec w^{(T+1)}\|_2 > Z_0 \geq 1$.
Next we consider the case 
$\|\vec w^{(T)}\|_2 > Z_0$
and $\|\vec w^{(T+1)}\|_2 > Z_0$.  
From \Cref{clm:ell2ellw}, we obtain that $\|\vec w^{(T+1)} - \vec w^{(T)}\|_{\vec w^{(T+1)}} 
\leq \sqrt{2} \eta B$. 
We prove the next lemma that bounds the ratio between two weighted euclidean norms
with different bases.
\begin{lemma} \label{lem:ellw-mutliplicative}
    Let $\vec u, \vec v, \vec x\in \R^d$ be non-zero vectors
    with $\|\vec u\|_2, \|\vec v\|_2 \leq Q$ for some $Q \geq 1$.
    Then it holds that 
    \[
    \frac{\|\vec x\|_{\vec u}}{\| \vec x\|_{\vec v}} \leq 
    \exp \left( 
Q^{3/2}
    \Big(  
    4 \theta(\vec u, \vec v) \Big( \frac{1}{\| \vec v \|^{3/2}} + \frac{1}{\| \vec v \|_2^{1/2}} \Big)
    + 
    \Delta(\vec u, \vec v)
    \Big)
    \right)
    \,,
    \]
    where 
$ \Delta(\vec u, \vec v)
    =
    \Big | \frac1 {\| \vec v \|_2^{3/2}} - \frac 1 {\| \vec u \|_2^{3/2}} \Big|
    + \Big | \frac1 {\| \vec v \|_2^{1/2}} - \frac 1 {\| \vec u \|_2^{1/2}} \Big|
    $.
    \end{lemma}
    \begin{proof}
        We first observe that, $\|\lambda \vec x\|_{\vec u} = |\lambda| \|\vec x\|_{\vec u}$, it
        suffices to consider $\vec x$ with unit norm $\|\vec x\|_2 = 1$.  Moreover, from
        \Cref{clm:ell2ellw} we have that since $\|\vec v\|_2, \|\vec u\|_2 \leq Q$ it holds that both
        $\|\vec x\|_{\vec u}, \|\vec x\|_{\vec v}$ are larger than $1/Q^{3/2}$.  Next, using the fact
        that the function $t \mapsto \log(t)$ is $1/t$-Lipschitz, we obtain that \( \log\left(
        \frac{\|\vec x\|_{\vec u}}{\|\vec x\|_{\vec v} } \right) \leq Q^{3/2} 
        |\|\vec x\|_{\vec u} - \|\vec x\|_{\vec v}| \).  Therefore, it suffices to bound the difference of the two norms.
        We have that 
        \[ 
            |\|\vec x\|_{\vec u} - \|\vec x\|_{\vec v}| 
            \leq 
            \left|
            \frac{\|\proj_{\vec u} \x\|_2}{\|\vec u\|^{3/2} }
            -
            \frac{\|\proj_{\vec v} \x\|_2}{\|\vec v\|^{3/2} }
            \right|
            +
            \left|
            \frac{\|\proj_{\vec u^\perp} \x\|_2}{\|\vec u\|^{1/2} }
            -
            \frac{\|\proj_{\vec v^\perp} \x\|_2}{\|\vec v\|^{1/2} }
            \right| \,.
            \]
            We first bound 
            the term 
            $
            \left|
            \frac{\|\proj_{\vec u} \x\|_2}{\|\vec u\|^{3/2} }
            -
            \frac{\|\proj_{\vec v} \x\|_2}{\|\vec v\|^{3/2} }
            \right|
            \leq 
            \left| \frac{\|\proj_{\vec u} \x\|_2}{\|\vec v\|^{3/2} }
            -
            \frac{\|\proj_{\vec v} \x\|_2}{\|\vec v\|^{3/2} }
            \right|
            + 
            \left| 
            \frac{1}{\| \vec u\|^{3/2}}
            -
            \frac{1}{\| \vec v\|^{3/2}} 
            \right| \,,
            $
            since $\|\proj_{\vec v} \x \|_2 \leq \| \x \|_2 = 1$.  Moreover, by Cauchy-Schwarz
            inequality, we have that $\|\proj_{\vec v} \x - \proj_{\vec u} \x\|_2 \leq 2 \theta(v, u)$.  Similarly, we
            can bound from above the term $\|\proj_{\vec v^{\perp}} \x \- \proj_{\vec u^\perp}\|_2$ by
            $2 \theta(\vec v, u)$.  The bound follows.
    \end{proof}
We will now use \Cref{lem:ellw-mutliplicative}.  We denote $\theta(\vec w^{(T)}, \vec w^{(T+1)})$ the angle between 
$\vec w^{(T)}, \vec w^{(T+1)}$ and by $\Delta(\vec w^{(T)}, \vec w^{(T+1)})$ the corresponding difference defined
in \Cref{lem:ellw-mutliplicative}.
Since $\|\vec w^{(T)}\|_2, \|\vec w^{(T+1)}\|_2 \geq Z_0 \geq 1$,
we have that $\theta(\vec w^{(T)}, \vec w^{(T+1)}) 
\leq \| \vec w^{(T)} -  \vec w^{(T+1)}\|_2 \leq \eta B $.
Moreover, using the fact that the mappings 
$\vec w \mapsto \| \vec w \|_2^{-3/2}$ and $\vec w \mapsto \|\vec w\|_2^{-1/2}$ 
defined for $\|\vec w \|_2 \geq 1$ are $3/2$ and $1/2$-Lipschitz 
respectively, we obtain that $\Delta(\vec w^{(T)}, \vec w^{(T+1)}) 
\leq \|\vec w^{(T)} - \vec w^{(T+1)}\|_2 \leq \eta B$.
Using \Cref{lem:ellw-mutliplicative}, we conclude that 
$ \|\vec w^{(T)} - \vec v\|_{\vec w^{(T+1)}}
\leq  e^{3 Z_1^{3/2} \eta B} \|\vec w^{(T)} - \vec v\|_{\vec w^{(T)} } 
\,.
$
Using the triangle inequality similarly to \Cref{eq:next-step-triangle} 
and the fact that $ \|\vec w^{(T)} - \vec v\|_{\vec w^{(T)} } \leq \beta_1$,
we obtain 
\begin{align*} 
\|\vec w^{(T+1)} - \vec v \|_{\vec w^{(T+1)}} 
\leq 
\sqrt{2} \eta B +  
e^{3 Z_1^{3/2} \eta B} \beta_1 \,.
\end{align*}
Finally, we have to consider the case where 
$\|\vec w^{(T)}\|_2 > Z_0$ and $\|\vec w^{(T+1)}\|_2 \leq Z_0$.
Using once more the triangle inequality and \Cref{clm:ell2ellw}, we obtain
$\|\vec w^{(T+1)} - \vec v \|_2 \leq \eta B +  \|\vec w^{(T)} - \vec v\|_2
\leq \eta B + (Z_0 + \eta B)^{3/2} \|\vec w^{(T)} - \vec v\|_{\vec w^{(T)}} 
\leq \eta B + (2 Z_0)^{3/2} \beta_1$, where we used the fact
that $\eta B \leq 1$ and $Z_0 \geq 1$.
Combining the bounds for the two cases where $\|\vec w^{(T+1)}\|_{2} \leq Z_0$,
we obtain that in this case it holds  $\|\vec w^{(T+1)} - \vec v \|_{2} \leq \eta B + \max(\alpha_1, (2 Z_0)^{3/2} \beta_1)$.
Similarly, when $\|\vec w^{(T+1)}\|_{2} > Z_0$ we have 
$\|\vec w^{(T+1)} - \vec v \|_{\vec w^{(T+1)}} \leq \sqrt{2} \eta B + 
\max(\sqrt{2} \alpha_1, e^{3 Z_1^{3/2} \eta B} \beta_1) \,.
$

\end{proof}

\subsection{Proof of \Cref{clm:estimation-gradient-bounded}}
We restate and prove the following claim.
\begin{claim}[Sample Complexity of Gradient Estimation]
    Fix $B,\eps' > 0$ with $\eps' \leq 1/\sqrt{B}$.  Using $N =  \wt{O}( (d/\eps'^2)
    \poly(\xi/(L \mu)) \log(B/\delta)) $ samples from $\D$, we can compute an empirical gradient
    field $\vec g(\vec w)$ such that for all $\vec w$ with $\|\vec w\| \leq B$, it holds $\| \vec
    g(\vec w) - \nabla F(\vec w) \|_2 \leq \eps'$ and  $\| \vec g(\vec w) - \nabla F(\vec w) \|_{*,
    \vec w} \leq \eps'$ 
with probability $1-\delta$.
\end{claim}
\begin{proof}
The result follows from the fact that the gradient random variable $\nabla F(\vec w)$ is sub-exponential
and therefore, its empirical estimate achieves fast convergence rates.
We draw use $N$ samples $(\x^{(i)}, y^{(i)})$ from $\D$ and form the standard empirical estimate 
\[
\vec g(\vec w) = \frac{1}{N} \sum_{i=1}^m (\sigma(\vec w \cdot \x\ith) -y\ith)\x\ith  \,.
\]
We first prove the estimation error in the dual norm of $\| \cdot \|_{\vec w}$.
Recall that the dual norm is equal to $\| \vec u \|_{*, \vec w} = 
\max( 
\| \proj_{\vec w}\vec u \|_2 \| \vec w \|_2^{3/2}, 
\|\proj_{\vec w^{\perp} } \vec u \|_2  \|\vec w\|_2^{1/2} 
) $.
 Recall that $\sigma'(\vec w \cdot \x) \geq 0 $ and $\sigma'(\vec w \cdot \x) \leq
\xi e^{-\mu |\vec w \cdot \x|}$ for all $\x$.  We first show that the distribution of $\x
(\sigma(\vec w\cdot \x) - y) \sigma'(\vec w \cdot \x)$ is sub-exponential.  
Fix any unit direction $\vec v$, we show an upper bound on the tail probability $|\vec v \cdot \x |
\geq t$.  We first consider the case $\vec v \cdot \vec w = 0$ and without loss of generality we
may assume that $\vec v$ is parallel to $\vec e_1$ and $\vec w$ is parallel to $\vec e_2$.  In the
following calculations we repeatedly use the properties of $(L,R)$-well-behaved distributions
and $(\tau, \mu,\xi)$-sigmoidal activations.
We have
\begin{align*}
    &\E_{(\x,y) \sim \D}[\1\{ |\vec v \cdot \x (\sigma(\vec w \cdot \x) - y) |  \geq t \} \sigma'(\vec w \cdot \x) ]
    \leq 
    \E_{(\x,y) \sim \D}[\1\{ |\vec v \cdot \x |  \geq t (\xi/\mu) \} \sigma'(\vec w \cdot \x) ]
    \\
    &\leq \E_{(\x,y) \sim \D}[\1\{ |\vec v \cdot \x |  \geq t (\xi/\mu) \} \xi e^{-\mu |\vec w \cdot \x|} ]
    \leq \frac{1}{L} \int_{-\infty}^{\infty} \int_{-\infty}^{\infty}
    \1\{ |\vec x_1 | \geq t (\mu/\xi) \} e^{-\mu \|\vec w\|_2 |\vec x_2| }
    e^{- L \sqrt{\vec x_1^2 + \vec x_2^2} } d \vec x_1 d \vec x_2 
    \\
    &\leq \frac{1}{L} \int_{-\infty}^{\infty} \1\{ |\vec x_1 | \geq t (\xi/\mu) \} e^{- L |\vec x_1|/2} d \vec x_1 
    \int_{-\infty}^{\infty} e^{-\mu \|\vec w\|_2 |\vec x_2| } e^{- L |\vec x_2|/2} d \vec x_2 
    = \frac{2}{L}  \frac{e^{-L/2 t \mu /\xi}}{L/2} \frac{1}{ \mu \|\vec w\|_2 + L/2} \\
    &\lesssim \frac{1}{L^2 \mu (\|\vec w\|_2 + 1)}  e^{-L\mu/(2\xi) \ t}  \,,
\end{align*}
where we used again the fact that $|\sigma(\vec w \cdot \x) - y| \leq \xi/\mu$,
the upper bound on the derivative of sigmoidal activations, see \Cref{def:well-behaved-bounded-intro} and
the fact that the density of any $2$ dimensional projection of $\D_\x$ 
is upper bounded by $(1/L)\exp(-L\|\x\|_2)$, see \Cref{def:bounds}.
Moreover, using the same properties as above we have that the variance of 
$\nabla F(\vec w) = \x (y - \sigma(\vec w \cdot \x) \sigma'(\vec w \cdot \x)$ along the direction $\vec v$ is 
at most
\begin{align*}
\var[\nabla F (\vec w) \cdot \vec v] 
&\leq    \Big(\frac{\xi}{\mu} \Big)^2 \Exx[(\sigma'(\vec w\cdot \x) \vec v \cdot \x )^2] 
\\
&\leq    \Big(\frac{\xi}{\mu} \Big)^2 \frac{1}{L} 
\int_{-\infty}^{+\infty} \int_{-\infty}^{+\infty}  
e^{-L \sqrt{\x_1^2+\x_2^2}}
~ \xi e^{- 2 \mu \|\vec w\|_2 |\x_1|} ~ d \vec x_1 d \vec x_2 
\\
&\leq    \frac{\xi^3}{\mu^2 L} 
\int_{-\infty}^{+\infty} \vec x_2^2e^{-L |\x_2|} ~ d \vec x_2  ~  
\int_{-\infty}^{+\infty}  e^{- 2 \mu \|\vec w\|_2 |\x_1|}  ~ d \vec x_1 
\\
&\lesssim   \frac{\xi^3}{\mu^3 L^4} \frac{1}{\|\vec w\|_2}\,.
\end{align*}
Therefore, along any direction $\vec v$ orthogonal to $\vec w$ we have that $\nabla F(\vec w)$ is
$(\sigma_1^2, b_1)$-sub-exponential with variance proxy $\sigma_1^2  = \poly(\xi/(L \mu)) 1/\|\vec w\|_2$
and rate $b_1 = O(\xi/(L \mu))$.  Therefore, Bernstein's inequality (see, e.g.,
\cite{Ver18}) implies that the empirical estimate $g(\vec w)$ with $N$ samples satisfies 
\[ \pr\left[ |g(\vec w) - \nabla F(\vec w)) \cdot \vec v| \geq t \right] 
\leq 2 e^{-\frac{N t^2/2}{ \sigma_1^2 +  b_1 t}} \,.  \]

We next analyze the gradient $\nabla F(\vec w)$ along the direction of $\vec w$.
We denote by $\wh{\vec w}$ the unit vector that is parallel to $\vec w$.
$
\| \proj_{\vec w} g(\vec w) - \proj_{\vec w} \nabla F(\vec w) \|_2
=
\| \wh{\vec w} \cdot \vec g(\vec w) \wh{\vec w} - \wh{\vec w} \cdot \vec g(\vec w) \wh{\vec w} \|_2
= 
|  \wh{\vec w} \cdot \vec g(\vec w)  -  \wh{\vec w} \cdot \vec g(\vec w)| 
$.
We first show the sub-exponential tail bound:
\begin{align*}
    &\E_{(\x,y) \sim \D}[\1\{ |\vec w \cdot \x (\sigma(\vec w \cdot \x) - y) |  \geq t \} \sigma'(\vec w \cdot \x) ]
    \leq 
    \E_{(\x,y) \sim \D}[\1\{ |\vec w \cdot \x |  \geq t (\xi/\mu) \} \sigma'(\vec w \cdot \x) ]
    \\
    &\leq \frac{1}{L} \int_{-\infty}^{\infty} 
    \1\{ |\vec x_2 | \geq t (\mu/\xi) \} e^{-\mu \|\vec w\|_2 |\vec x_2| } e^{- L |\vec x_2|}  d \vec x_2 
    \\
    &= \frac{1}{L(\mu \|\vec w\|_2 + L)} e^{- (\mu \|\vec w\|_2 + L) \mu/\xi t } \,.
\end{align*}
Similarly, we can compute the variance along the direction of $\vec w$.
\begin{align*}
\var[\nabla F(\vec w) \cdot \wh{\vec w}] 
&\leq    \Ey[((\sigma(\vec w \cdot \x) -y)\sigma'(\vec w\cdot \x) \wh{\vec w} \cdot \x )^2] 
\\
&\leq    \Big(\frac{\xi}{\mu} \Big)^2 \Exx[(\sigma'(\vec w\cdot \x) \wh{\vec w} \cdot \x )^2] 
\\
&\leq   \frac{\xi^3}{\mu^2 L} \int_{-\infty}^{+\infty} e^{- 2 \mu \|\vec w\|_2 |t|} t^2 d t
\\
&=  \frac{\xi^3}{2 \mu^5 L} \frac{1}{\|\vec w\|_2^3 } \,,
\end{align*}
where we used the fact that $|\sigma(\vec w \cdot \x) - y| \leq \xi/\mu$,
the upper bound on the derivative of sigmoidal activations, see \Cref{def:well-behaved-bounded-intro} and
the anti-concentration property of $(L, R)$-well-behaved distributions, see \Cref{def:bounds}.

Therefore, along the direction $\vec w$  we have that $\nabla F(\vec w)$ is
$(\sigma_2^2, b_2)$-sub-exponential with variance proxy $\sigma_2^2  = \poly(\xi/(L \mu)) 1/\|\vec w\|_2^3$
and rate $b_1 = O(\xi/ (L \mu^2 \|\vec w\|_2))$.  Therefore, Bernstein's inequality (see, e.g.,
\cite{Ver18}) implies that the empirical estimate $g(\vec w)$ with $N$ samples, satisfies 
\[ \pr\left[ |(g(\vec w) - \nabla F(\vec w)) \cdot \wh{\vec w}| \geq t \right] 
\leq 2 e^{-\frac{N t^2/2}{ \sigma_2^2 +  b_2 t}} \,.  \]

Thus, we can now perform a union bound on every direction $\vec u$ and every $\|\vec w \| \leq B$ by
using a $r = \poly(L \mu)/(\xi B)  \eps')$-net of the unit sphere and the ball of radius $B$ (which
will have size $r^O(d)$, see e.g., \cite{Ver18}), we obtain that along any direction $\vec v$
orthogonal to $\vec w$ it holds that $|\vec v \cdot (\vec g(\vec w) - \nabla F(\vec w))| \leq
\eps/\|\vec w\|_2^{1/2}|$ with probability at least $1 - 2 r^{O(d)} e^{-\frac{N \eps^2/2}{ \|\vec w\|_2
\sigma_1^2 + \|\vec w\|_2^{1/2} b_1 \eps}}$.  Substituting the variance and rate values $\sigma_1^1, b_1$, we
observe that $ \|\vec w\|_2 \sigma_1^2 + \|\vec w\|_2^{1/2} b_1 t \leq \poly(\xi/(L \mu)) $,
where we used our assumption that $\eps' \leq 1/\sqrt{B}$ (and that $\|\vec w\|_2 \leq B$).
Picking $N = \wt{O} (d/\eps'^2 \poly(\xi/(L \mu)) \log(B/\delta))$ we obtain that the above bound holds with
probability at least $1- \delta/2$.  Similarly, performing the same union bound for the direction of
$\vec w$ and using the sub-exponential bound with $\sigma_2^2, b_2$ 
and the fact that $\eps' \leq 1/\sqrt{B}$ we again obtain that $|\vec v
\cdot (\vec g(\vec w) - \nabla F(\vec w))| \leq \eps'/\|\vec w\|_2^{3/2}$ with probability at least
$1-\delta/2$ with $N = \wt{O}(d/\eps'^2 \poly(\xi/(L \mu)) \log(B/\delta))$ samples.
The bound for the $\ell_2$ distance of $\vec g(\vec w)$ and $\nabla F(\vec w)$ follows similarly
from the above concentration bounds.
\end{proof}

\subsection{Proof of \Cref{lem:gradient-field-unbounded}}
We restate and prove the following:
\begin{lemma}[Gradient Field Distance Reduction]
Let $\vec g: \R^d \mapsto \R^d$ be a vector field and fix $W, B \geq 1$.
We assume that $\vec g$ satisfies the following properties with respect to some unknown target vector $\vec v$
with $\|\vec v\|_2 \leq W$. Fix parameters $\alpha_1, \alpha_2 > 0$.
For every $\vec w \in \R^d$ with $\|\vec w\|_2 \leq 2 W$ it holds that $\|g(\vec w)\|_2 \leq B\max(\|\vec w - \vec v\|_2,\alpha_1)$.
Moreover, if $\|\vec w - \vec v\|_2 \geq \alpha_1$ and $\theta(\vec w, \vec v) \in (0, \pi/2)$ then it holds that 
$\vec g(\vec w) \cdot (\vec w - \vec v) \geq \alpha_2 \|\vec w - \vec v\|_2^2 $. 
We consider the update rule $\vec w^{(t+1)} \gets \vec w^{(t)} - \eta \vec g(\vec w^{(t)})$ 
initialized with $\vec w^{(0)} = \vec 0$ and step size  
\(
\eta \leq  \alpha_2/B.
\)
Let $T$ be any integer larger than  $\lceil (W^2+\log(1/\alpha_1))/(\eta \alpha_2) \rceil$, it 
holds that $ \| \vec w^{(T)} - \vec v\|_2  \leq  (1+\eta B)\alpha_1 $.
\end{lemma}
\begin{proof}
We first show that at every round the distance of $\vec w^{(t)}$ and the target $\vec v$ decreases.
Since we initialize at $\vec w = \vec 0$ (notice that in this case it holds trivially $\|\vec w^{(0)}\|_2 \leq W$) after the first update we have 
\begin{align*}
\|\vec w^{(1)} - \vec v \|_2^2 -\|\vec w^{(0)}-\vec v \|_2^2
&=  
-2\eta \vec g(\vec w^{(0)}) \cdot (\vec w^{(0)}- \vec v) +\eta^2\|\vec g(\vec w^{(0)})\|_2^2
\\
&\leq  -2\eta \alpha_2  \|\vec w^{(0)} - \vec v\|_2^2+ \eta^2 B^2 \|\vec w^{(0)} - \vec v\|_2^2
\\
&\leq   -\eta \alpha_2 \|\vec w^{(0)} - \vec v\|_2^2 \,,
\end{align*}
where we used that since 
$\eta \leq \alpha_2 /B$ it holds that 
$\eta\alpha_2-\eta^2B^2\geq 0$.
Moreover, we observe that after the first iteration we are  going to have 
$\vec w^{(1)} \cdot \vec v \geq 0$  (since otherwise the distance to $\vec v$  would increase).  
This implies that $\theta(\vec w^{(1)}, \vec v) \in (0, \pi/2)$.
Observe now that it holds $\|\vec w^{(1)}\|_2 \leq 2 W$ 
since $\vec w^{(1)}$ is closer to $\vec v$ than $\vec w^{(0)}$.
Using induction, similarly to the previous case we can show that 
for all iterations $t$ where $g(\vec w\tth ) \cdot ( \vec w\tth - \vec v) \geq \alpha_2 \|\vec w\tth - \vec v\|_2^2$ 
it holds 
$\|\vec w^{(t+1)}\|_2 \leq 2W $ and $\theta(\vec w^{(t+1)}, \vec v) \in (0, \pi/2)$ and that the distance to the target vector $\vec v$ decreases at 
every iteration:
\[
\|\vec w^{(t+1)} - \vec v \|_2^2 \leq \|\vec w^{(t)}-\vec v \|_2^2(1-\eta\alpha_2)\leq \|\vec w^{(0)}-\vec v \|_2^2(1-\eta\alpha_2)^{t}\;.
\]
Let $T'$ be the first iteration where  the assumption 
$g(\vec w^{(T')} ) \cdot ( \vec w^{(T')} - \vec v) \geq \alpha_2 \|\vec w^{(T')} - \vec v\|_2^2$ 
does not hold.  Since at each round we decrease 
the distance of $\vec w^{(t)}$ and $\vec v$, and we initialize at $\vec 0$ we have that 
 the number of iterations $T' \leq \lceil (\|\vec v\|_2^2+\log(1/\alpha))/(\eta \alpha_2) \rceil$.
Moreover, even if we continue to use the update rule after the iteration $T'$ from the triangle 
inequality we have that 
\[
 \| \vec w^{(T'+1)} - \vec v\|_2  
 \leq  \| \vec w^{(T'+1)} - \vec w^{(T')} \|_2  + \| \vec w^{(T')} - \vec v  \|_2  
 \leq \eta \|\vec g(\vec w^{(T')})\|_2 + \alpha_1\leq (1+\eta B )\alpha_1 \,.
 \]
 Therefore, by induction we get that for all $T \geq T'$ it holds that 
 $ \| \vec w^{(T)} - \vec v\|_2  \leq (1+\eta B)\alpha_1 $.
 \end{proof}

\subsection{Proof of \Cref{clm:truncate-labels-unbounded}}
We restate and prove the following claim.
\begin{claim}
    Let $M=\Theta(W\max(\log(W/\eps),1))$. Denote by $\mathrm{tr}(y)= \sign(y) \min(|y|,M)$. Then, it holds that:
    \[
      \Ey[(\mathrm{tr}(y)-\sigma(\wstar\cdot \x))^2]\leq O(\eps)\;.
    \]
\end{claim}
\begin{proof}
Denote $\mathcal B=\{\x\in\R^d:|\wstar\cdot \x|\leq M\}$, i.e., the points $\x$ such that $|\wstar\cdot \x|$ is at most $M$.
We have that for $\x \in \mathcal{B}$, it holds that 
$ |\tilde y - \sigma(\vec w^\ast \cdot \x)| \leq |y - \sigma(\vec w^\ast \cdot \x)|  $, where $\tilde{y}$ is equal to $y$ 
truncated in $-M\leq y\leq M$. Therefore,
we have that 
\[
 \Ey[(\tilde{y}-\sigma(\wstar\cdot \x))^2\1\{\x\in \mathcal B\}]\leq \Ey[(y-\sigma(\wstar\cdot \x))^2\1\{\x\in \mathcal B\}]\leq \eps\;.
\]
Using the triangle inequality, we have
\begin{align*}
   \Ey[(\tilde{y}-\sigma(\wstar\cdot \x))^2]&= \Ey[(\tilde{y}-\sigma(\wstar\cdot \x))^2\1\{\x\in \mathcal B\}]+
\Ey[(\tilde{y}-\sigma(\wstar\cdot \x))^2\1\{\x\not\in \mathcal B\}]
\\&\leq \eps+ 2M^2\pr_{\x\sim\D_\x}[\x\not\in \mathcal B] +2\Exx[\sigma(\wstar\cdot \x)^2\1\{\x\not\in \mathcal B\}]\;.
\end{align*}
To bound $\Exx[\sigma(\wstar\cdot \x)^2\1\{\x\not\in \mathcal B\}]$, note that $\|\wstar\|_2\leq W$ and that $|\sigma(\wstar\cdot \x)|\leq \lambda |\wstar\cdot\x|$. Assuming without loss of generality that $\wstar$ is parallel to $\vec e_1$ and let $V$ be the subspace spanned by $\wstar$ and any other vector orthogonal to $\wstar$. Using that the distribution $\D_\x$ is $(L,R)$-well-behaved, we have that  it holds that 
\[ \Exx[\sigma(\wstar\cdot \x)^2\1\{\x\not\in \mathcal B\}]\leq \lambda^2W^2\Exx[\x_1^2\1\{\x_1\geq M/W\}]\leq \eps/2\;,\]
which follows from $\lambda^2\Exx[|\x_1|^2\1\{\x_1\geq M/W\}] \leq \lambda^2(M^2/(L^2 W^2))\exp(-L M/W)\leq \eps/2$, its proof is similar with the proof of \Cref{clm:bound-square-conce}.
It remains to bound the $\pr_{\x\sim\D_\x}[\x\not\in \mathcal B]$, we have
\begin{align*}
    \pr_{\x\sim\D_\x}[\x\not\in \mathcal B]=&\int_{|\wstar\cdot\x|\not\in \mathcal{B}}\gamma_{V}(\x)\d\x
    \leq \frac{2}{L}\int_{t\geq  M/\|\wstar\|_2}e^{-L t}\d t=\frac{2}{L^2}e^{-L  M/\|\wstar\|_2}\leq \frac{\eps}{2M^2}\;.
\end{align*}
Hence, $(1/2)\Ey[(\tilde{y}-\sigma(\wstar\cdot \x))^2]\lesssim \eps$.
Therefore, by truncated the $y$ to values $-M\leq y\leq M$, we increase the total error at most $\eps'=O(\eps)$. For the rest of the proof for simplicity we abuse the notation of the symbol $y$ for the $\tilde{y}$ and $\eps$ for the $\eps'$.
\end{proof}

\begin{corollary}
\label{cor:bounded-main-guessing}
Let $\D$ be an $(L,R)$-well-behaved distribution on 
$\R^d \times \R$ and $\sigma(\cdot)$ be a $(\tau, \boundc, \boundd)$-sigmoidal activation.
Define $F(\vec w) =  (1/2) \E_{(\x, y) \sim \D}[ (\sigma(\vec w \cdot \x) - y )^2]$
and $\opt = \inf_{\vec w \in \R^d} F(\vec w) $.  
Set $\kappa = \poly(L R\tau \mu /\xi)$.
There exists an algorithm that given $N = \wt{\Theta}( d/\eps  \log(1/\delta) ) ~ \poly(1/\kappa)$ samples, runs in time $T = \poly(1/(\eps\kappa)) $ and returns a vector $\widehat{\vec w} \in \R^d$ that, 
with probability $1-\delta$, satisfies 
\[ 
F(\widehat{\vec  w} ) \leq \poly(1/\kappa) ~ \opt + \eps\,.
\] 
\end{corollary}
\begin{proof}
The proof of
\Cref{cor:bounded-main-guessing} is similar to the proof of
\Cref{thm:bounded-main-theorem}, the main point which the two proofs diverge is that we  have to guess the value $\rho$ of the regularizer. To do this, we construct a grid of the possible
values of $\opt$, i.e., $\mathcal G=\{\eps,2\eps,\ldots, \Theta(\xi/\mu)\}$. We
choose the upper bound to be of size $\Theta(\xi/\mu)$ because this is the
maximum value that the error can get, see e.g., \Cref{clm:upper-bound-value-bounded}. Therefore, by running
the \Cref{alg:gd-2}, with appropriate parameters as in
\Cref{thm:bounded-main-theorem} but by using for the regularizer, each value
of $\mathcal G$ instead of $\opt$. There exist a $t\in \mathcal G$ such that
$|\opt-t|\leq \eps$, therefore from
\Cref{prop:gradient-bound-regular}, this will converge to a
stationary point with $(1+\eps)\poly(1/k)\opt$, see e.g.,
\Cref{prop:gradient-bound-regular} for $\Lambda=1+\eps$. Hence, by running \Cref{alg:gd-2} for each value in $\mathcal G$, we output a list $U$ that contains $O(\xi/(\mu\eps)) $ different hypothesis. We construct the empirical $\D_N$ of $\D$ by taking $N$ samples. We need enough samples, such that for each $\vec w\in U$, it will hold $\widehat{F}(\vec w)=\E_{(\x,y)\sim \D_N}[(\sigma(\vec w\cdot \x)-y)^2]\leq 2 \E_{(\x,y)\sim \D}[(\sigma(\vec w\cdot \x)-y)^2]$.
From Markov's inequality we have that 
\[
\pr[|\widehat{F}(\vec w)-F(\vec w)|\geq F(\vec w)/2]\leq 4\frac{\E[\widehat{F}(\vec w)^2]}{F(\vec w)^2}\leq \frac{\E[(\sigma(\vec w\cdot \x)-y)^4]}{N F(\vec w)^2}\;.
\]
Note that because $y,\sigma(\vec w\cdot \x)$ is bounded by $\Theta(\xi/\mu)$, we have that
$\E[(\sigma(\vec w\cdot \x)-y)^4]\leq \poly(\xi/\mu)\E[(\sigma(\vec w\cdot \x)-y)^2]$, hence we have that
\[
\pr[|\widehat{F}(\vec w)-F(\vec w)|\geq F(\vec w)/2]\lesssim \poly(\xi/\mu) \frac{1}{NF(\vec w)} \lesssim \poly(\xi/\mu) \frac{1}{N \eps}\;, 
\]
therefore by choosing $N=\poly(\xi/\mu) (1/\eps)$, we have with probability at least $2/3$ that 
$\widehat{F}(\vec w)\leq 2F(\vec w)+\eps$ and by standard boosting procedure we can increase the probability to $1-\delta''$ with $O(\log(1/\delta''))$ samples. Hence, by choosing $\delta''=\delta/|U|$, we have with probability $1-\delta$, that for $\hat{\vec w}=\argmin_{\vec w\in U}\widehat{F}(\vec w)$, it holds that $F(\hat{\vec w})\lesssim \opt +\eps$.
\end{proof} 
\section{Learning Halfspaces using the Ramp Activation}\label{app:halfspaces}
In this section we show that by optimizing as a surrogate function the ramp activation, we can find a $\hat{\vec w}\in \R^d$, that  gets error $\Ey[(\sign(\hat{\vec w}\cdot \x)-y)^2]=O(\eps)$.
Specifically, we show the following corollary of \Cref{cor:bounded-main-guessing}.

\begin{corollary}
\label{cor:halfspaces}
Let $\D$ be an $(L,R)$-well-behaved distribution on 
$\R^d \times \{\pm 1\}$.
Define $F(\vec w) =  (1/2) \E_{(\x, y) \sim \D}[ (\sgn(\vec w \cdot \x) - y )^2]$
and $\opt = \inf_{\vec w \in \R^d} F(\vec w) $.  
Fix $\eps,\delta\in(0,1)$ and set $\kappa = \poly(L R)$.
There exists an algorithm that given $N = \wt{\Theta}( d/\eps  \log(1/\delta) ) ~ \poly(1/\kappa)$ samples, runs in time $T = \poly(1/(\eps\kappa)) $ and returns a vector $\widehat{\vec w} \in \R^d$ that, 
with probability $1-\delta$, satisfies 
\[ 
F(\widehat{\vec  w} ) \leq \poly(1/\kappa) ~ \opt + \eps\,.
\] 
\end{corollary}

\begin{proof}
   We are going to show that by optimizing the ramp activation instead of the $\sign(\cdot)$, we can find a good candidate solution. First note that $r(t)$ is an $(1,e,1)$-sigmoidal activation. The proof relies on the following fact:

  \begin{fact} Let $\vec w\in \R^d$ be a unit vector. Then, for any $\x\in \R^d$, it holds that
   \[\lim_{z\to \infty}r(z\vec w\cdot \x)=\sign(\vec w\cdot \x)\;.\]
\end{fact}
Therefore, for any unit vector $\vec w\in \R^d$, from \Cref{lem:radius-approx} there exists a $\vec w'\in \R^d$ with $\|\vec w'\|_2=O(1/\eps)$ such that $\Exx[(r(\vec w'\cdot \x)-\sign(\vec w\cdot \x))^2]\leq \eps$.
From \Cref{cor:bounded-main-guessing}, we get that there exists an algorithm that with $N = \wt{\Theta}( d/\eps  \log(1/\delta) ) ~ \poly(1/\kappa)$ samples, and runtime $T = \poly(1/(\eps\kappa)) $ returns $\hat{\vec v}\in \R^d$ with $\|\hat{\vec v}\|_2= O(1/\eps)$, such that with probability $1-\delta$, it holds 
\[
(1/2) \E_{(\x, y) \sim \D}[ (r(\hat{\vec v} \cdot \x) - y )^2]\leq \poly(1/(LR))\opt +\eps\;.
\]

The proof follows by noting that because $r(\hat{\vec v})\in (-1,1)$ and $y\in \{\pm 1\}$, it holds that
\[
\E_{(\x,y)\sim \D}[(r(\hat{\vec v}\cdot \x)-y)^2]\geq (1/2)\E_{(\x,y)\sim \D}[(\sign(r(\hat{\vec v}\cdot \x))-y)^2]=(1/2)\E_{(\x,y)\sim \D}[(\sign(\hat{\vec v}\cdot \x)-y)^2]\;.
\]
Hence, $F(\hat{\vec v})\leq \poly(1/(LR))\opt +2\eps$.
\end{proof}

Next, we prove the following sample complexity lower bound for optimizing sigmoidal activations.
\begin{lemma}[Sample Complexity Lower Bound for Sigmoidal Activations]
	\label{lem:sample-lower-bound-sigmoidal}
	Fix any absolute constant $C \geq 1$ and let $\sigma(t) = 1/(1 + e^{-t})$ be the logistic
	activation.  Any $C$-approximate algorithm that learns (with success probability at least $2/3$)
	the logistic activation under any $\eps$-corrupted $\D$ with standard normal $\x$-marginal
	requires $\Omega(d/\eps)$ samples from $\D$.
	\end{lemma}
\begin{proof}
	Assume that $y = \1\{\vec w^\ast \cdot \x \geq 0\}$ for some unit vector $\vec w^\ast$, i.e.,
	$y$ is a noiseless halfspace.  It is easy to see that the corresponding distribution $D$ with
	Gaussian $\x$-marginal
	and $y$ given by $\1\{\vec w\cdot \x \geq 0\}$ is $\eps$-corrupted for any $\eps > 0$.
	Since 
	\(\lim_{z\to \infty} \sigma(z\vec w^\ast \cdot \x)=\1\{\vec w \cdot \x \geq 0 \}\),
    by the dominated convergence theorem we have that 
	\(
		\lim_{z\to \infty}\E_{(\x,y) \sim D} F^{D,\sigma}(z \vec w^\ast) = 0 \,.
	\)
	Therefore, $\inf_{\vec w\in \R^d} F^{D, \sigma}(\vec w) = 0$.
	Let $\cal A$ be any algorithm that given a sigmoidal activation $\sigma$ and $N>0$ samples from $\D$, returns a vector $\vec w$ such that $F^{D,\sigma}(\vec w)\leq \eps$.  
	We will show that $\vec w$ corresponds to a classifier with at most $2 C \eps$ error.  
	By noting that $\sigma(\vec w)\in (0,1)$ and $y\in \{0,1\}$, it holds that
	\begin{align*}
\E_{(\x,y)\sim \D}[(\sigma(\vec w \cdot \x)-y)^2]
&\geq (1/2)\E_{(\x,y)\sim \D}[(\1\{\sigma(\vec w\cdot \x) \geq 1/2 \})-y)^2]
\\
& = (1/2) \pr_{(\x,y) \sim \D }[\1\{\sigma(\vec w\cdot \x) \geq 1/2 \})\neq y]\;.
	\end{align*}
	Thus, we have constructed a binary classifier that achieves disagreement at most $2 C \eps$ with
	$y$.  It is well-known that any algorithm that finds a vector $\vec w \in \R^d$ such that
	$\pr[\sign(\vec w\cdot \x)\neq y]\leq O(\eps)$, needs $\Omega(d/\eps)$ samples (see, e.g., \cite{Long:95}).
Hence, $\cal A$ needs at least $N=\Omega(d/\eps)$ samples.
\end{proof}

 \end{document}